\documentclass[a4paper,12pt,times,numbered,print,index,custommargin]{Classes/PhDThesisPSnPDF}

\usepackage{nomencl} 
\makenomenclature 

\input{Preamble/preamble}

\degreetitle{Computer and Control Engineering}

\cycle{$30^{th}$}

\title{Text-based Sentiment Analysis and Music Emotion Recognition}

\author{Erion \c Cano}
\label{author}

\supervisor{Prof. Maurizio Morisio\\
}

\committee{Prof. Bart Dhoedt, \emph{Referee}, Ghent University, Belgium\\
	Prof. Lo\"{i}c Barrault, \emph{Referee}, Le Mans University, France \\
	Prof. Rosa Meo, \emph{External Member}, University of Turin, Italy  \\
	Prof. Fulvio Corno, \emph{Internal Member}, Politecnico di Torino, Italy \\ 
	Prof. Marco Torchiano, \emph{Internal Member}, Politecnico di Torino, Italy }

\subject{LaTeX} 

\ifdefineAbstract
\fi

\ifdefineChapter
\fi

\begin{document}

\frontmatter

\maketitle


\begin{declaration}
I hereby declare that the contents and organization of this dissertation constitute my own original work and does 
not compromise in any way the rights of third parties, including those relating to the security of personal data.


\end{declaration}


\begin{dedication} 

%
\emph{\Large To my parents for their unconditional love and support} 
%

\end{dedication}

%
\begin{acknowledgements}      
\indent \indent 
This thesis ends a very important journey of my life and is a result of hard work,  
profound commitment as well as support from many people I wish to thank.  
%
My deepest gratitude goes to my parents who have loved and supported me 
unconditionally during all these years of my long education.
%
I am thankful to Politecnico di Torino for believing in me, for their generous support 
and comfortable working environment.  
%
Special thanks go to my supervisor, Prof. Maurizio Morisio for his continuous guidance 
throughout my research work. I had the opportunity to learn a lot from his long experience. 
Furthermore, the freedom he gave me in conducting research strengthened my self-confidence 
and motivation. 
%
\par 
I could not forget to express my gratitude to my SoftEng group colleagues, Prof. Marco 
Torchiano, Luca Ardito, Oscar Rodriguez, Cristhian Figueroa, Iacopo Vagliano,  
Mohammad Rashid, Riccardo Coppola, and Diego Monti, for their valuable suggestions and 
collaboration. I am also highly thankful to 
Ihtesham Ul Islam, Andrea Martina, Amirhosein Toosi, Francesco Strada, Alysson Dos 
Santos as well as the other LAB 1 colleagues and friends for the multicultural and pleasant 
atmosphere they created and shared during these years. 
%
\par 
Sincere thanks go to TIM (formerly Telecom Italia) for
the financial support of my research. I highly appreciate the collaborative and creative 
work I conducted within their JOL MobiLab, in cooperation with wonderful people like 
Marco Marengo, Lucia Longo, Eleonora Gargiulo, Mirko Rinaldini, Luisa Rocca, 
Lucia Laferla and Martina Francesconi.
%
I finally thank the HPC@POLITO team for providing their high-performance computing infrastructure which fulfilled the heavy computational requirements of my experimental activity.
%


\end{acknowledgements}

\begin{abstract}
%
\indent \indent
Nowadays, with the expansion of social media, large amounts of user-generated texts like 
tweets, blog posts or product reviews are shared online. Sentiment polarity analysis of such 
texts has become highly attractive and is utilized in recommender systems, market 
predictions, business intelligence and more. We also witness deep learning techniques 
becoming top performers on those types of tasks. There are however several problems that 
need to be solved for efficient use of deep neural networks on text mining and text polarity 
analysis. 
\par 
First of all, deep neural networks are data hungry. 
They need to be fed with datasets that are big in size, cleaned and preprocessed as well as properly
labeled. Second, the modern natural language processing concept of word embeddings as a dense 
and distributed text feature representation solves sparsity and dimensionality problems of the 
traditional bag-of-words model. Still, there are various uncertainties regarding the use of word 
vectors: should they be generated from the same dataset that is used to train the model or it 
is better to source them from big and popular collections that work as generic text feature 
representations? Third, it is not easy for practitioners to find a simple and highly effective 
deep learning setup for various document lengths and types. Recurrent neural networks are 
weak with longer texts and optimal convolution-pooling combinations are not easily conceived. 
It is thus convenient to have generic neural network architectures that are effective and can 
adapt to various texts, encapsulating much of design complexity.
\par
This thesis addresses the above problems to provide methodological and practical insights for 
utilizing neural networks on sentiment analysis of texts and achieving state of the art results. 
Regarding the first problem, the effectiveness of various crowdsourcing alternatives is explored 
and two medium-sized and emotion-labeled song datasets are created utilizing social tags. 
%
One of the research interests of Telecom Italia was the exploration of relations between 
music emotional stimulation and driving style. Consequently, a context-aware music 
recommender system that aims to enhance driving comfort and safety was also designed. 
To address the second problem, a series of experiments with large text collections of various 
contents and domains were conducted. Word embeddings of different parameters were 
exercised and results revealed that their quality is influenced (mostly but not only) by the 
size of texts they were created from. When working with small text datasets, it is thus 
important to source word features from popular and generic word embedding collections. 
Regarding the third problem, a series of experiments involving convolutional and 
max-pooling neural layers were conducted. Various patterns relating text properties 
and network parameters with optimal classification accuracy were observed. Combining 
convolutions of words, bigrams, and trigrams with regional max-pooling layers in a 
couple of stacks produced the best results. The derived architecture achieves competitive 
performance on sentiment polarity analysis of movie, business and product reviews.  
\par 
Given that labeled data are becoming the bottleneck of the current deep learning systems,
a future research direction could be the exploration of various data programming 
possibilities for constructing even bigger labeled datasets. Investigation of   
feature-level or decision-level ensemble techniques in the context of deep neural networks
could also be fruitful. Different feature types do usually represent complementary 
characteristics of data. Combining word embedding and traditional text features or 
utilizing recurrent networks on document splits and then aggregating the predictions 
could further increase prediction accuracy of such models. 
\end{abstract}


\tableofcontents

\listoffigures

\listoftables


\mainmatter


\chapter{Introduction}  
\label{chapter1}
\ifpdf
    \graphicspath{{Chapter1/Figs/}{Chapter1/Figs/PDF/}{Chapter1/Figs/}}
\else
    \graphicspath{{Chapter1/Figs/Vector/}{Chapter1/Figs/}}
\fi
%
%
%
\indent \indent
%
Social networks and the Internet have increased our communication capabilities to the 
point that everyone with a connected device can be read or heard worldwide. Sharing 
opinions about politics, sport, brands or products in social networks is now a cultural 
trend. Consumers are especially inclined to share or consult online opinions about products
they have bought or are willing to buy. On the other hand, companies are motivated 
to find innovative ways for getting benefit using opinion data that are daily posted 
online. In fact, marketing statistics suggest that 
81\% of shoppers conduct online research before making big purchases and 
92\% of marketers consider social media to be important for their business.\footnote{\url{https://www.hubspot.com/marketing-statistics}} 
\par 
Sentiment Analysis (SA) is considered as the process of computationally identifying and 
categorizing opinions expressed in a piece of text, especially in order to determine 
whether the writer's attitude towards a particular topic or 
product is \emph{positive}, \emph{negative}, or
\emph{neutral}.\footnote{\url{https://en.oxforddictionaries.com/definition/sentiment\_analysis}}
In other words, it is a set of techniques and practices for automatic identification of 
sentiment polarity in texts. 
From the business perspective, analyzing opinions of clients is essential for several 
reasons, such as improve product or service quality, measure and improve the success 
of marketing campaigns, determine or adjust marketing strategy, etc.  
Sentiment analysis results are in fact widely used as a component of recommendation
engines that generate advertisements for online users in many websites.  
\par 
Back in the early 2000s, text polarity was mostly analyzed starting from words or 
phrases and going up to entire document. Creating lexicons of affect terms and 
using them to infer word polarity was a common practice 
\cite{DBLP:conf/lrec/Bestgen08, Kaji+Kitsuregawa:07a}.  
%
Later on, the popularization of the web, social networks, and cloud services lured 
users to give more and more feedback on a daily basis. User opinions about 
different types of items come in various ways. 
Besides text posts or comments, social tags have become very popular as well, 
especially as an instrument for performing sentiment analysis of songs. 
They are basically single word descriptors like \dq{rock}, \dq{sweet} or \dq{awsome} 
that express users' opinion for a certain song or another object type. Tags 
of \emph{Last.fm} (an online radio station and web platform for music listening with 
open API) were frequently used in various Music Emotion Recognition (MER) 
research papers \cite{Lin:2011:EOM:2037676.2037683, Hu2007}. 
%
At the same time, the rising popularity of microblogs that are rich in user
opinions, motivated many researchers to create datasets of emotionally labeled 
texts. As a result, focus gradually shifted from unsupervised to supervised learning 
methods for performing SA. This trend was also propelled by the development of highly 
effective machine or deep learning techniques and the popularization of high-level 
libraries or frameworks for using them. 
\par 
Neural networks offer wide applicability and are able to automate 
feature extraction and selection in many domains and tasks. The release of 
powerful graphics processing units for scientific computing enhanced quick 
training and usability of deep neural networks for heavy tasks such as language
modeling for predicting word combinations. One of the first neural language models 
was proposed by Bengio et \emph{al.} in 2003 \cite{Bengio:2003:NPL:944919.944966}. 
Simpler and easier to train models were proposed in the following years. Huge and 
increasing amounts of text started to be used for training them and generating word 
feature representations also known as word vectors or embeddings. These word 
features offer significant advantages over the traditional bag-of-words text 
representation such as density, reduced dimensionality, preserved semantic relations 
between words, etc. Word embeddings and neural networks started to be used as 
classification features on several text mining tasks. 
\par 
This thesis addresses current problems in sentiment analysis of music and different
types of texts, such as movie reviews, product reviews, etc. A particular
interest of Telecom Italia was the exploration of social tags and other crowdsourcing
alternatives for creating emotionally labeled song datasets of considerable size.
Those datasets form the basis for music emotion recognition, music recommender
systems, and other relevant applications. Quality assessment and effectiveness of
word embeddings on sentiment analysis tasks was another concern. Utilizing different
types of neural network layers for text feature generation, feature selection, and
sentiment prediction is not straightforward. To reach the desired performance, a high
number and type of hyperparameters need to be tuned. The last part of the thesis
addresses the possibility of encapsulating such complexity in a network architecture
that can be used to quickly prototype accurate models for sentiment analysis. In
brief, the research questions we pose are the following:
\begin{description}
%
\item[RQ1] \emph{Are affect tags of songs and other crowdsourcing alternatives 
applicable to music emotion recognition or music emotion 
dataset creation?}
\item[RQ2] \emph{What effect do training method, corpus size, corpus topic, and other 
characteristics have in the performance of generated word vectors used in sentiment 
analysis tasks?}
\item[RQ3] \emph{How could design and hyperparameter tuning complexity
of the neural network models for sentiment analysis be reduced?}
\end{description}
\par 
Regarding RQ1, \emph{Last.fm} user tags of about one million recent songs are 
crawled and analyzed. Our observations, same as those of previous 
related works such as \cite{conf/ismir/LaurierSSH09} show that indeed, social 
tag emotional spaces are in consonance with those of psychologists (e.g., the popular 
model of Russell \cite{russell1980circumplex}). Tags are thus a viable means for 
exploring emotions in music. We also create two relatively big datasets of song emotions
which are publicly released for research use. Other crowdsourcing 
mechanisms such as online games or services and Mechanical Turk  
are explored and found useful for gathering labels about songs or similar types 
of items.   
\par 
Considering RQ2, several experiments with song lyrics and movie reviews were
conducted, using Glove and Skip-Gram methods for generating word vectors of
texts. We noticed that corpus size and its vocabulary richness have a significant impact
on the quality of word vectors that are generated. Word vectors trained on big text
bundles tend to perform better on sentiment analysis of song lyrics and movie reviews.
Thematic relevance between training corpus and task seems to have a lower importance 
on performance and should be further assessed in a more comprehensive experimental 
framework. We also observed that Glove and Skip-Gram training methods have
slightly different performance patterns. The former is a slightly better player on big text
bundles when used to analyze song lyrics whereas the latter gives best results on 
small or average text sources used on movie reviews.
\par 
Responding to RQ3, various experiments with convolution and pooling layers
were conducted. Convolutions of different kernels seem to be highly capable of
capturing $n$-gram word vector combinations of texts. Furthermore, pooling layers
(e.g., max-pooling) are very effective for selecting the most salient word feature maps
for sentiment classification. Combining generic word vectors trained on billion-sized
text bundles with convolution and pooling stacks produced state-of-the-art results in
sentiment analysis of song lyrics, movie reviews or other types of texts. This thesis
proposes a neural network architecture of convolutional and max-pooling neural 
layers that can be used as a template for rapid prototyping of highly efficient sentiment 
analysis models. It adapts to various text lengths and dataset sizes with little need for 
hyper-parameter adaptions.
\par
The thesis is structured as follows. Chapter~\ref{chapter2} introduces different concepts 
about sentiment analysis, summarizes milestone achievements in artificial intelligence
and presents important facts which highlight the current shift towards data-driven
supervised learning techniques. In Chapter~\ref{chapter3} psychological viewpoints about
emotions in music and their representation using simplification models are first
summarized. Later on, crowdsourcing possibilities for harvesting emotional labels
of songs are discussed. Finally, \emph{Last.fm} user tags are utilized for constructing
two public and big datasets of song emotions. Chapter~\ref{chapter4} introduces recommender
systems as important information filtering tools, explores hybrid and context-based
recommenders and presents the design of a recommender system of songs in the
context of car driving. Influence of various text characteristics on the quality of
generated word vectors is explored and presented in Chapter~\ref{chapter5} in details. 
Chapter~\ref{chapter6} presents the results of various experiments with convolution 
and max-pooling neural layers on sentiment analysis of song lyrics, movie reviews and 
other types of texts. NgramCNN neural network architecture variants and their performance 
scores are discussed in Chapter~\ref{chapter7}. Finally, Chapter~\ref{chapter8} presents 
the main contributions, derived conclusions, and possible future research directions. 
\chapter{Background}
\label{chapter2}
\ifpdf
    \graphicspath{{Chapter2/Figs/}{Chapter2/Figs/PDF/}{Chapter2/Figs/}}
\else
    \graphicspath{{Chapter2/Figs/Vector/}{Chapter2/Figs/}}
\fi
%
\begin{flushright}
	\emph{\large \enquote{In god we trust, all others must bring data.}} \\
	-- W. Edwards Deming, statistician 
\end{flushright}
\vskip 0.35in
\indent \indent
In the last fifteen years, enormous amounts of user opinion texts have been posted on 
the Web, especially on social media websites. All these data that keep growing 
exponentially have created incentives for developing automatic methods and tools
that are able to analyze user opinions about different brands, products, services or other entities. 
\emph{Sentiment Analysis} that is also known as \emph{Opinion Mining} is a collection 
of methods, techniques, and practices used for automatic analysis of opinions, sentiments 
or attitudes about those entities. Business intelligence, market predictions, customer care 
and online marketing are some of its most popular and common application realms.
Sentiment analysis is also highly interrelated with other fields such as \emph{Natural Language 
Processing} or \emph{Artificial Intelligence} that are recently enjoying rapid progress as well, 
strongly driven by the online data revolution that is happening. In particular, statistical 
language models, lexicons, and text representation methods, together with machine learning 
and neural networks are highly important for the correct and intelligent interpretation of opinions. 
\par 
In this chapter, we first present some basic concepts and definitions related to sentiment analysis. 
Later in Section~\ref{sec:DataExplosion}, we discuss the current proliferation of data 
in the Web as a result of the advancing digitalization process. Section~\ref{sec:SATechsAndApps} 
provides an overview on some of the most common techniques and successful applications 
of sentiment analysis. In Section~\ref{sec:DLHistory}, important milestone and transformational developments in artificial intelligence, machine learning, and natural language processing are 
outlined. Finally, Section~\ref{sec:LabeledHunger} briefly presents various strategies that can 
be used to generate large quantities of labeled data which are essential for training effective 
prediction models.  
%
%
\section{Sentiment Analysis Concepts and Definitions} 
\label{sec:SADef}
%
Sentiment Analysis (SA) can be considered as the computational examination of 
sentiments, opinions, emotions, and attitude expressed in text units towards an 
entity \cite{medhat2014sentiment}. It is about identifying, extracting and classifying
opinions and attitudes about various issues. SA represents a really wide and  
flourishing research realm today. A good indicator of this fact is obviously the vast 
collection of terms it is referenced by. Various designations like Sentiment Analysis, Opinion 
Extraction, Opinion Mining, Affect Analysis, Emotion Analysis or Subjectivity 
Analysis are interchangeably used in research publications to denote similar tasks. 
%
According to \cite{pang2008opinion}, \emph{Subjectivity Analysis} is the earliest term 
that was first used in late 90s. It insinuates the recognition of opinion-oriented language 
in texts and its separation from the objective language. Later in 2001, we had the first 
research papers using the term \emph{Sentiment Analysis} when referring to the 
automatic sentiment polarity analysis of subjective texts. Shortly after in 2003, the 
term \emph{Opinion Mining} first appeared in a WWW conference publication. 
\par 
Opinion is obviously the central concept in SA. Liu in \cite{liu2012sentiment} provides a 
formalized definition about it. According to him, an opinion is a quintuple 
\emph{(e\textsubscript{i}, a\textsubscript{ij}, s\textsubscript{ijkl}, h\textsubscript{k}, t\textsubscript{l})}.
Here \emph{e\textsubscript{i}} is the name of an entity and \emph{a\textsubscript{ij}} is an aspect
(or component) of \emph{e\textsubscript{i}}. Together \emph{e\textsubscript{i}} and 
\emph{a\textsubscript{ij}} represent the target (\emph{g}) of the opinion. Furthermore, 
\emph{s\textsubscript{ijkl}} is the sentiment on aspect \emph{a\textsubscript{ij}}, 
\emph{h\textsubscript{k}} is the opinion holder and \emph{t\textsubscript{l}} is the time when 
the opinion was expressed. Opinion sentiment \emph{s\textsubscript{ijkl}} is 
\emph{positive}, \emph{negative}, or \emph{neutral}. It might also be a numeric rating 
that expresses the intensity of the sentiment 
(e.g., 1 -- 5 stars rating). \emph{Sentiment classification} task
is about determining what polarity does opinion sentiment \emph{s\textsubscript{ijkl}} 
have on aspect \emph{a\textsubscript{ij}} (or on target \emph{g}).  
%
Another important concept in SA is that of \emph{opinion words} that are terms 
commonly used to express \emph{positive} (e.g., \dq{excellent}, \dq{wonderful}, \dq{great}) 
or \emph{negative} (e.g., \dq{poor}, \dq{terrible}, \dq{awful}) opinions. Lists of 
opinion words are usually created as a means for solving sentiment classification 
tasks and form a \emph{sentiment lexicon}. 
\par 
In the case when a text document contains opinions of only one opinion holder about a single 
entity and the different aspects of that entity are irrelevant, the document is analyzed 
entirely. In this case the task is known as \emph{document-level sentiment classification} \cite{liu2012survey}. This task represents the main focus of this thesis. 
Analysis of user reviews about online products typically falls into this category. When sentiment
classification is applied to single subjective sentences the task is usually called 
\emph{sentence-level sentiment classification}. Here we assume that the sentence holds only 
one opinion from a single opinion holder. In other words, the sentence must be simple like 
\enquote{Battery life of this camera is great}. This task is also highly interrelated with 
\emph{subjectivity classification} which determines if a sentence is subjective or 
objective. Document-level and sentence-level sentiment classifications do not exactly reveal
what one likes or not about an entity (e.g., battery life). In complex sentences like 
\enquote{This camera is great and I like the picture quality}, the user provides opinions about 
different entity aspects. Analyzing such sentences by first identifying each aspect or target  
is called \emph{aspect-level sentiment classification}.  
%
Finally, it is important to mention that SA is an NLP problem that is restricted only to some 
aspects of word, sentence or document semantics. Nevertheless, a holistic understanding
of the application domain and problems is usually an essential prerequisite. As we will see 
in the following chapters, text preprocessing and cleaning part is an important step in 
every SA solution. 
%
\section{Exponential Online Data Explosion}    
\label{sec:DataExplosion} 
%
Technological progress keeps making computing gadgets faster, lighter and cheaper at the 
point that more and more people can afford them. One of the biggest consequences of this 
is the digitalization of our lives; today we use digital mail, books, photos, music, documents, 
maps, etc. Everyone of us relies on online learning, news or advertisements. We also order
products, tickets or food and pays bills using digital devices connected to the Internet.   
%
It is interesting to note that most of the free content we daily consume online is also 
freely created and made available by us. The paradigm of today is \dq{put it online, 
get it online.} The consequences of this trend have been pointed out paradoxically in the 
appealing quote of Tom Goodwin: 
%
\emph{\dq{Uber, the world's largest taxi company owns no vehicles, Facebook the world’s 
most popular media owner creates no content, Alibaba, the most valuable retailer has no 
inventory and Airbnb the world's largest accommodation provider owns no real estate. 
Something interesting is happening}}.\footnote{Tom Goodwin in \url{http://www.techcrunch.com}, 
March 2015}  
\par
\begin{figure}
	\centering
	\includegraphics[width=0.72\textwidth]{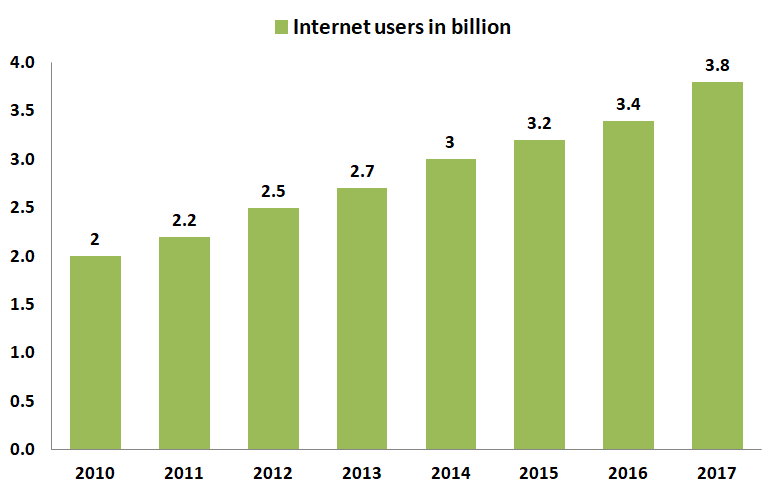}
	\caption{Tendency of Internet users in the last years}
	\label{fig:UsersTrend}   
\end{figure}
\begin{figure}
	\centering
	\includegraphics[width=0.77\textwidth]{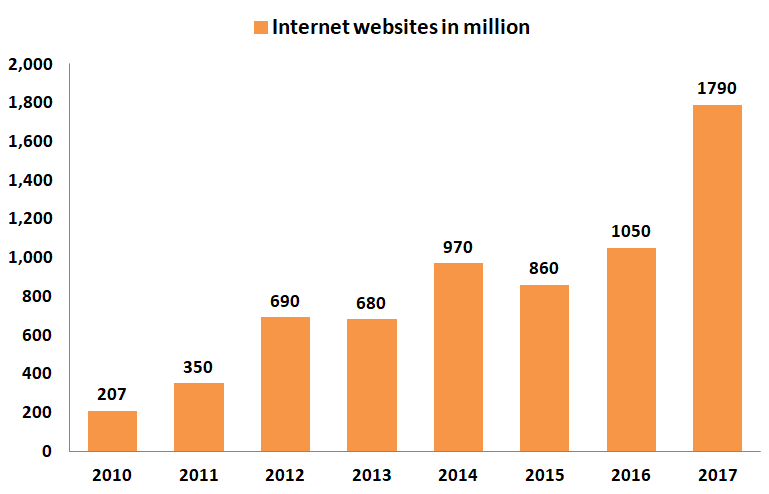}
	\caption{Tendency of Internet websites in the last years}
	\label{fig:WebsitesTrend}   
\end{figure}
\begin{figure}
	\centering
	\includegraphics[width=0.77\textwidth]{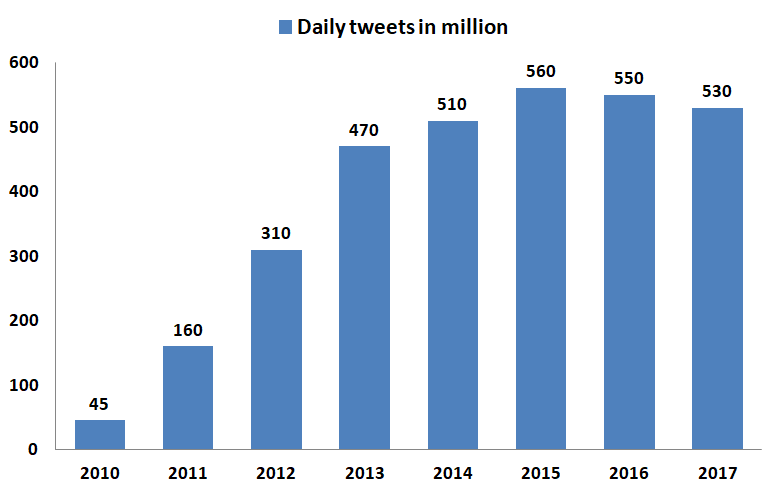}
	\caption{Tendency of daily tweets in the last years}
	\label{fig:TweetsTrend}   
\end{figure}

In fact, Internet statistics reveal us that user-generated content in sites like Wikipedia, Twitter, 
Facebook, Youtube or Instagram has indeed been growing exponentially in the last 
years.\footnote{\url{http://www.internetlivestats.com}}
According to those statistics, as of December 2017, there are roughly 3.8 billion 
Internet users (the trend in Figure~\ref{fig:UsersTrend}), almost 1.8 
billion websites (the trend in Figure~\ref{fig:WebsitesTrend}) and 43 million Wikipedia 
pages. Furthermore, every hour we have 182,000 TB of Internet traffic, 9.5 billion emails sent 
and 232 million Google searches served. There are also 28 million tweets sent (the trend in 
Figure~\ref{fig:TweetsTrend}), three million photos uploaded on Instagram, 200 million 
posts shared on Facebook and 30 thousand hours of video playback uploaded on Youtube. 
\par
Users also contribute with valuable content in the form of reviews on sites like Amazon, 
TripAdvisor, and Yelp. That content has become highly important in the decision making 
of consumers. People rely on informal reviews to decide where to eat, 
what gadgets to buy, where to sleep etc. According to 
statistics,\footnote{\url{https://www.vendasta.com/blog/50-stats-you-need-to-know-about-online-reviews
}} 92\% of consumers today read online reviews and only 12\% are prepared to read more than ten 
of them. Regarding persuasion effect of reviews, 40\% of the subjects form an opinion by 
reading just one to three reviews and 73\% of them form an opinion by reading up to six 
reviews. Also, one up to three bad online reviews would be enough to deter most of (67\%)
shoppers from buying a product or service. There are of course credibility problems 
created by fake reviews (e.g., from AI trained bots) that come out once in a while. In fact, 
statistics reveal that 95\% of consumers doubt faked reviews when they do not see bad scores, 
or that 30\% of them assume online reviews are tricked if there is no negative feedback at all. 
Despite the credibility issues, all that free online content that is growing exponentially 
has greatly motivated the invention and utilization of novel SA solutions. In the following 
section, basic methods for solving the sentiment classification problem are discussed.   
\section{Sentiment Analysis Techniques and Applications} 
\label{sec:SATechsAndApps}
%
As mentioned above, sentiment classification problem is central to SA. This problem 
has been addressed using a high variety of techniques. Survey works like 
\cite{medhat2014sentiment} or \cite{ravi2015survey} categorize those techniques 
as \emph{machine learning}, \emph{lexicon-based} or \emph{hybrid}.  
%
The first to consider sentiment polarity analysis as a type of document classification were 
Pang and Lee in \cite{Pang:2002:TUS:1118693.1118704}. They exercised na\"ive Bayes,
Maximum Entropy and Support Vector Machine algorithms on movie reviews they had 
collected and prepared. 
%
Since that time, researchers have extensively explored all kinds of supervised learning 
algorithms like Bayesian networks or $k$-nearest neighbors, linear classifiers like logistic 
regression, decision trees, random forests, rule-based classifiers or even the most 
recent deep neural networks. In the early 2000s, there was a tendency of trying unsupervised 
or hybrid approaches to overcome the need for labeled data. For example, authors of \cite{Ko:2000:ATC:990820.990886} implemented a hybrid approach by first 
dividing each document into sentences which are classified as \emph{positive} 
or \emph{negative} using keyword lists of each category. Labeled sentences are then 
used as data for training traditional supervised algorithms and making the overall prediction. 
\par
A popular unsupervised method for review polarity classification was proposed by Turney 
in \cite{turney2002thumbs}. First, a POS tag is assigned to each word of the reviews to 
recognize adverb or adjective phrases. Then semantic orientation of words or phrases is computed 
using PMI-IR algorithm. Pointwise Mutual Information (PMI) between two 
words $w_1$ and $w_2$ is given by Equation~\ref{PMIw1w2}:     
\begin{equation}
\label{PMIw1w2}
PMI(w_1, w_2) = log_2 (p(w_1 \, and \, w_2) / p(w_1) p(w_2))
\end{equation}
Here $p(w_1 \, and \, w_2)$ is the probability of the two words appearing near each other. 
The Semantic Orientation (SO) of each word or phrase is thus computed using 
Equation~\ref{SOw1w2} below:
\begin{equation}
\label{SOw1w2}
SO(phrase) = PMI(phrase, ``excellent") - PMI(phrase, ``poor")
\end{equation}
This way average semantic orientation is computed for every review. Values greater than zero 
are associated to \emph{positive} reviews whereas negative semantic orientation values are 
associated to \emph{negative} reviews. Experimental results of Turney revealed accuracy 
scores ranging from 65\% to 84\%, depending on the types of items analyzed. 
In the following years, unsupervised sentiment analysis has been little explored, in contrast
with supervised techniques that gained exceptional popularity motivated by the rapid 
growth of online user reviews and posts.  
%
Lexicon-based approaches extensively utilize 
opinion words or phrases found in texts that express \emph{positive} or \emph{negative} 
polarity. The first step here is filling the list of words of each polarity with a couple of seed 
words. Next, the lists are extended by searching for synonyms or antonyms in dictionaries. 
Another common practice is to search on large text corpora for co-occurrence of seed words 
with other opinion words. Similar approaches analyze texts based on word statistics or 
formal ontologies that capture semantic associations between concepts. 
\par
%
Regarding applications of SA, there seems to be a well-established tradition in 
macroeconomic predictions about markets, businesses, brands or products 
\cite{ravi2015survey}. There are many studies that search for correlations between
positivity of texts in social media and financial data records, 
using the former to model and predict the latter.  
%
Companies are also highly motivated to investigate the reasons why customers accept or reject 
a new product. In this context, understanding customer needs and preferences through 
opinion mining is highly important and most big companies incorporate it as 
part of their mission. Online product reviews are becoming important to the point that 
various websites are soliciting user feedback about restaurants, hotels or other commodities.  
%
A common but less obvious application of SA is its use as a subcomponent in other 
quickly growing technologies like recommender systems. One can easily imagine 
utilizing positivity of user reviews about certain items to create clusters of users
that might become potential clients. The same information can be utilized to augment 
and improve item profiling as well. 
%
Other potential applications of SA include question answering systems, stock market 
forecasting, political surveys, criminology etc.  
%
%
%
\section{Machine and Deep Learning Expansion}  
\label{sec:DLHistory}
%
%
\subsection{Early Mileston Developments}
The recent success of machine learning and especially deep neural networks 
in various industries is producing fascinating results. Andrew Ng who is considered as 
one of the pioneers in the field has predicted a revolutionary effect of deep 
learning in today's society, similar to that of electricity about 100 years 
ago.\footnote{\url{http://fortune.com/2016/10/05/ai-artificial-intelligence-deep-learning-employers/}}
%
The first milestone was probably the neuron prototype designed back in 1943 by 
McCulloch and Pitts \cite{mcculloch43a}. They combined a weighted linear function 
with what they called \dq{threshold logic} to model the working of the human brain. 
%
Inside the philosophical debate of whether a machine can think or not, Alan Turing 
evaded the question by proposing \dq{The Imitation Game} (known as Turing Test), 
a series of criteria for assessing if a machine could be enough intelligent to fool a person 
into thinking it is actually a human \cite{10.2307/2251299}. While the debate still 
goes on, the pragmatic need to sort out humans from computers in the Web brought out 
automatic and public versions of Turing Test commonly known 
as CAPTCHA.\footnote{Completely Automated Public Turing test to tell Computers 
and Humans Apart}
\par
The progress in AI would not become visible to the general public until Frank
Rosenblatt invented the first artificial neural network called \dq{Perceptron}   
back in 1957 \cite{rosenblatt1958perceptron}. Inspired by the human visual system 
and designed for image recognition, this newborn machine created a lot of enthusiasm
and resurrected the old debate about the limits of AI. 
%
Technically a \dq{Perceptron} is a linear function of input values, limited in 
representing linear decision boundaries for learning operations like AND, OR and NOT,
but not XOR. The first to make this observation were Minski and Papert in 1969 
who also proved that is was theoretically impossible for the \dq{Perceptron} to 
learn XOR function \cite{minsky69perceptrons}. Consequently, the so-called first 
AI winter came on and progress was sluggish until the beginning of the 1980s.   
%
The most important development that followed was a paper from David Rumelhart, 
Geoff Hinton and Ronald Williams, entitled \dq{Learning representations by
back-propagation errors} \cite{rumelhart1985learning}. In that paper, they showed the 
simple procedure (Backprop algorithm) that could be used to train neural networks with 
many hidden layers which unlike the \dq{Perceptron} could learn nonlinear functions.  
Almost at the same time, Yann LeCun made the first practical demonstration of a
convolutional neural network trained with back-propagation for recognizing
handwritten digits.  
%
Unfortunately, neural networks could still not scale to larger problems and the broken 
expectations together with the collapse of Lisp machine 
market\footnote{\url{https://danluu.com/symbolics-lisp-machines/}} 
resulted in a second winter (during the 1990s).  
\subsection{From Logic to Data}
%
While neural networks were left untouched, there was still time for other developments.
Ensemble learning came out as a novel branch of machine learning techniques that utilize multiple 
learners to solve the same problem with higher predictive accuracy. The various techniques
usually differ in how they select the training data and the way they aggregate the 
decisions of each classifier. Bootstrap aggregating known as \emph{Bagging} trains
ensemble models on randomly picked subsets of the training set and makes them vote 
with equal weight. Random forest, for example, is a combination of decision trees 
with bagging that provides a robust classification on problems with huge numbers of features. 
\emph{Boosting} on the other hand, incrementally learns various classifiers adding them 
to a final model. When each classifier is added, the data are reweighted giving importance 
to training instances that previous learners misclassified. In works like
\cite{freund1996experiments}, Adaboost which is the most common boosting
implementation is reported to yield better improvements than bagging.  
\par
Roughly at the same time, Support Vector Machine (SVM) classifier was invented and 
proven to be highly effective \cite{cortes1995support}. It maps the input vectors into a 
high dimensional feature space and constructs a linear decision surface to ensure high 
generalization ability and global optimality. Furthermore, the notion of soft margins extends 
its applicability to non-separable training data. A few years later, the kernel trick generalized 
usability of SVMs beyond linear functions by transforming the input space into a feature 
space \cite{Scholkopf:2001:LKS:559923}.    
%
All these advances in machine learning helped to shift the focus from logic-driven,
deductive solutions to data-driven, statistical ones. The power of computers 
started to be used for analyzing big quantities of data and inferring conclusions from the
obtained results. 
%
%
Moreover, popularization of the Web helped to create bigger experimental
datasets and boosted research in disciplines like \emph{information retrieval}, 
\emph{text analysis}, \emph{text mining}, etc. Bag-Of-Words (BOW) representation 
was commonly used for representing texts in many studies. It was typically combined 
with various classification algorithms like $k$-nearest neighbors, na\"ive Bayes, decision 
trees or SVM. In \cite{joachims1998text} we find one of the first empirical 
comparisons of such algorithms in text categorization tasks. According to its experimental 
results, SVM appears the dominant classification algorithm in text categorization tasks.  
That finding emphasizes the fact that strong points of SVM match text document characteristics 
such as high dimensionality of input space (big vocabulary size), the few irrelevant 
features (words) and the sparsity of document-term matrix.
\subsection{Current Hype of Deep Learning}
\label{sec:DeepLearningHype}
%
The exponential increase of user-generated content in the emerging Web 2.0 
social media of the early 2000s contributed to the proliferation of even bigger datasets 
of labeled images or texts.
ImageNet,\footnote{\url{http://image-net.org}} a huge dataset of millions of 
labeled images from different categories was created and made available to support 
research in computer vision and image recognition. 
%
At the same time, faster and cheaper hardware (especially GPUs for generic computations) 
became available and could be quickly and economically deployed in the form of cloud services.
This extra supply in computing infrastructure stimulated proliferation of startups that
created intelligent and innovative services.   
%
Interest in neural network research resurrected and was rebranded as Deep Learning (DL). 
Groundbreaking network designs like Long Short-Term Memory (LSTM) which was proposed in 
1997 and improved in 2000 came out. Generative neural networks that can work in both 
supervised and unsupervised contexts like \emph{deep belief networks} 
or \emph{deep Boltzmann machines} followed in 2006 and 2009. 
More recent designs like \emph{generative adversarial networks} presented in 
2014, significantly reduced computational costs. 
%
Furthermore, novel regularization techniques like Dropout or Batch Normalization 
helped for the mitigation of problems like overfitting and accelerated training by reducing
covariate shift. That kind of problems had hindered usability of deep neural architectures 
for many years. Consequently, improved and deeper architectures of convolutional neural networks 
like \emph{VGG-19},\footnote{\url{http://www.robots.ox.ac.uk/~vgg/research/very_deep/}}
\emph{AlexNet},\footnote{\url{https://en.wikipedia.org/wiki/AlexNet}} or
\emph{Inception}\footnote{\url{https://github.com/google/inception}} 
were developed and made available for public use. Those big models 
have won many international image recognition competitions like ILSVRC.\footnote{\url{http://www.image-net.org/challenges/LSVRC}}  
Together with open source software platforms like \emph{Torch} 
(2002), \emph{Theano} (2008) or \emph{Tensorflow} (2015), they have democratized AI 
creating the hype of today.    
\subsection{From n-grams to Neural Language Models}  
%
An important contribution of neural networks in NLP has to do with the transition 
from the traditional statistical language models to the more advanced neural ones. 
A language model is a probability function that is able to describe important 
statistical characteristics of word sequence distributions in a natural language. It typically 
enables us to make predictions about the next (or previous) word appearing in a sequence 
of given words. This capability is successfully implemented in various applications of 
natural language processing like speech recognition, automatic language translation, 
spell checking, etc. 
In a $n$-gram model for example, the probability $P(w_1, ~ \dots ~ , w_m)$ of word 
sequence $w_1, ~ \dots ~ , w_m$ is computed as:
\begin{equation}
\label{ngramModProb}
P(w_1, ~ \dots ~ , w_m) = \prod_{i=1}^{m}P(w_i | w_{1}, ~ \dots ~ , w_{i-1}) \approx \prod_{i=1}^{m}P(w_i |  w_{i-(n-1)}, ~ \dots ~ , w_{i-1}) 
\end{equation}
As we can see, the above approximation assumes that the probability of observing the $i^{th}$
word $w_i$ in context of the proceeding $i - 1$ words can be calculated as the probability of 
observing it in the shorter context history of the proceeding $n - 1$ words. This property 
(known as \emph{Markov} or \emph{memoryless} property) is a characteristic of those 
stochastic processes in which the conditional probability distribution of future states 
depends upon the present state only. The most common $n$-gram models are 
\emph{unigram} ($n = 1$), \emph{bigram} ($n = 2$) and \emph{trigram} ($n = 3$). 
The conditional probability of those models is obtained from the frequency of $n$-gram
counts\footnote{In practice, smoothing algorithms are used to give some probability
weight to unseen n-grams.} as:
\begin{equation}
\label{ngramModFreqCounts}
P(w_i | w_{i-(n-1)}, ~ \dots ~ , w_{i-1}) = \frac{count(w_{i-(n-1)}, ~ \dots ~ , w_{i-1}, w_i)}{count(w_{i-(n-1)}, ~ \dots ~ , w_{i-1})}
\end{equation}
\par 
In contrast, neural language models utilize continuous space word embeddings (also known 
as distributed word feature vectors) to make language predictions. One of the motivations for 
developing such models was the need to fight the \emph{course of dimensionality} 
problem that results when models are trained on large text collections with a big vocabulary 
(number of unique words) size $V$. The total number of possible word sequences increases 
exponentially with $V$ causing severe data sparsity.  
Neural language models avoid the \emph{course of dimensionality} problem by representing 
text words in a distributed way. 
%
They are conceived as probabilistic classifiers that learn a probability 
distribution over vocabulary $V$ given some linguistic context that is typically a 
fixed-size window of previous $k$ words. Neural language models are thus trained to predict:
%
\begin{equation*}
\label{NeuralModelProb}
P(w_m | w_{m-k}, ~ \dots ~ , w_{m-1}) ~~ \forall m ~ \in ~ V 
\end{equation*}
%
Bengio \emph{et al.} in \cite{Bengio:2003:NPL:944919.944966} presented one of the first 
attempts to train and evaluate a distributed representation of words. Authors proposed to 
learn the probability function $P(w_m | w_1, ~ \dots ~ , w_{m-1})$ by decomposing it in 
the following two parts:
\begin{enumerate}
	\item A mapping $C$ from any element $i$ of $V$ to a real vector $C(i) \in \mathbb{R}^{d}$.
	This basically represents the distributed feature vectors (word embeddings) of $d$ dimensions
	associated with each word of the vocabulary.
	\item A function $g$ that maps an input sequence of distributed context word vectors 
	$C(w_{t-k+1}, ~ \dots ~ , C(w_{m-1}))$ to a probability over words in $V$ for 
	next world $w_m$. Function $g$ creates as output a vector whose $i$-th element estimates 
	the probability $P(w_m = i | w_1, ~ \dots ~ , w_{m-1})$.
\end{enumerate}
This way, according to their approach:
\begin{equation}
\label{Bengio2003Decomposition}
P(w_m = i | w_1, ~ \dots ~ , ~ w_{m-1}) = g(i, C(w_m-1),~\dots ~ , ~ C(w_m-n+1)) 
\end{equation}
As a result, the number of free parameters scales only linearly (not exponentially as in 
$n$-gram models) with vocabulary size $V$. Authors implement their solution by means of a 
Multi-Layer Perceptron (MLP) network and compare against a state-of-the-art trigram 
model. They report 10 -- 24 percent higher performance at the cost of a significantly longer
training time. Yet computation requirements scale linearly with the number of conditioning
variables. 
%
Their work paved a path of significant improvements in language models,
transiting from discrete and sparse representations ($n$-grams) to continuous and dense 
feature vectors that are highly compliant with the emerging neural network technologies.
In the epoch of faster GPUs and continuously improving neural architectures, many 
other researchers followed the same path, proposing neural models with better generalization 
abilities or improved computational efficiency.
\subsection{Word Embeddings for Text Representation}
\label{sec:WordEmbIntro}
%
One of the first popular methods for generating word embeddings, known as C\&W (for 
Collobert and Weston) was presented in \cite{Collobert:2008:UAN:1390156.1390177}. 
Authors make use of a CNN architecture to generate word embeddings from a Wikipedia 
text corpus. They report significant improvements in various NLP tasks such as part-of-speech 
tagging, named entity recognition, semantic role-labeling etc. A few years later, 
Continuous Bag-of-Words (CBOW), and Skip-Gram were presented in 
\cite{DBLP:journals/corr/abs-1301-3781}. Authors utilize shallow networks for 
easier training of models that predict a word based on the context (CBOW) or predict 
the context words of a given word (Skip-Gram). Both methods are considerably improved 
(in terms of training time) in \cite{DBLP:conf/nips/MikolovSCCD13} where 
negative sampling and subsampling of frequent words are also presented. 
At this point, interesting properties of word embeddings 
like the exhibition of syntactic and semantic regularities between words became more 
evident. For example, word analogies (in this case regarding gender) appear in algebraic
regularities of the form:
\begin{equation*}
\label{WordAnalogyEq}
v(king) - v(male) \approx v(queen) - v(female)
\end{equation*}
where $v$ represents the learned $d$-dimensional feature vector of the corresponding word. 
Datasets with word analogy questions were created and started to be used for evaluating 
the quality of word embeddings. 
\par
Glove introduced in \cite{conf/emnlp/PenningtonSM14} represents another recent and 
popular method for generating word embeddings. Authors train texts using word-word 
co-occurrence counts and global corpus statistics. At the same time, they preserve the linear 
structure of CBOW and Skip-Gram. According to authors' results, Glove scales very well 
on huge text collections and outruns similar methods on many tasks, including word analogies. 
In Section \ref{sec:QualWordEmb} we present and discuss our own experiments, comparing 
Glove and Skip-Gram on different SA tasks. 
%
%
Regarding applicability in NLP or SA, BOW representation 
remains popular despite the data sparsity problem it suffers from. This popularity is 
mainly because of its simplicity and performance, especially when used in combination 
with SVM classifiers on texts of relatively small vocabularies. Word embeddings on the 
other hand, while still being more computation intensive have better generalization 
abilities and are immune to high dimensionality problem even on texts of large vocabularies. 
The tendency towards bigger training datasets and continuous speed-ups of neural networks 
is expected to favor distributed text representations via word embeddings.
%
%
\subsection{Deep Learning Applications and Achievements}
%
The growing applicability of deep neural networks, besides shadowing earlier machine 
learning techniques in traditional tasks, also opened up new perspectives in more 
difficult scenarios. 
%
Speech recognition and automatic machine translation are two application domains where 
LSTMs are excelling. 
%
Healthcare is probably the most important domain for humanity. Deep learning image 
recognition systems are already analyzing X-rays better than radiologists do. Drug discovery, 
melanoma screening or brain cancer detection are appealing applications as well. 
%
Genomics or gene editing is also a highly complex task that is getting advantage from 
deep neural networks. 
%
In the realm of cybersecurity, pattern recognition of newer viruses or network threats is 
being trusted to neural network models. 
%
Other applications include optimizing space mission efforts where an Italian team of 
scientists is currently applying neural networks \cite{pub2657678}. 
\par 
At the same time, we have witnessed spectacular achievements involving humans competing 
against machines. In 2011 for example, Watson,\footnote{\url{https://www.ibm.com/watson/}} 
the supercomputer of IBM defeated the two most reputable 
champions of Jeopardy, a quiz show where players are introduced with knowledge indications 
and are expected to respond with a question. That game involves complex skills like 
memorization, intuition, agility, etc. Furthermore, AlphaGo program of Google beat in 2016 a 
professional human Go player. The program renamed AlphaZero was later generalized to 
play chess and other mind games as well. 
\par
The natural question that comes to our minds is what makes deep neural networks 
better players than traditional machine learning approaches (e.g., SVM) in so many
tasks. In fact, an obvious advantage of today's neural network architectures is their ability to 
automatically extract and select features. This eliminates the need for 
hand-crafting features out of data, speeding up model creation and deployment.
Convolution layers, for example, are famous for their ability to find the most representative
features in images (e.g., ears, noses or eyes in face recognition). 
%
When different convolution layers are stacked one after the other, they transform 
features received from previous the layers into more complex and detailed features
that are eventually passed to the classifier (usually the last few layers). This very successful 
paradigm of using deep feature extraction and selection layers combined with a simple 
classifier has been the basis of the image award-winning architectures mentioned 
in Section~\ref{sec:DeepLearningHype}. 
%
\section{Hunger for Labeled Data} 	
\label{sec:LabeledHunger}
%
%
\emph{\enquote{Data is the new oil.}} We have been reading or hearing this intriguing 
phrase by so many information technology leaders, public speakers or even politicians 
in the last decade. Clive Humbly, a mathematician and data scientist is usually credited 
as the first one to have coined it.\footnote{\url{http://ana.blogs.com/maestros/2006/11/data_is_the_new.html}} 
%
The basic idea is to emphasize the essential role that data and information have in 
society. Actually, data has been around for hundreds of years. The difference 
today is in the quantity of data that users provide daily by utilizing free online services 
like Google searches, Twitter or Facebook posts and comments, etc. The other important 
difference is our ability to process it and extract great value from it using advanced AI
techniques and technologies. This ability is giving such an immense advantage to 
technology giants, making economy experts discuss antitrust regulations, similar to those 
imposed to oil companies at the beginning of $20^{th}$ 
century.\footnote{\url{http://www.cbc.ca/news/technology/data-is-the-new-oil-1.4259677}}   
\par 
Same as internal combustion engines invented about 150 years ago, AI techniques can 
be considered as \dq{prediction engines} of today. AI is \dq{fueled} from 
labeled training data in the same way as combustion engines are fueled by refined oil. 
However, there is one point where this interesting analogy breaks. It is usually harder 
to find crude oil than to refine it into gas. The same thing is not true about data. It is 
so easy today to find it in immense quantities. Yet it is harder and expensive to label 
or \dq{refine} it, producing the necessary fuel for our prediction engines. 
%
Deep neural networks of today are incredibly complex and utilize millions of parameters to 
generate their \dq{magical} predictions. They are thus data hungry, requiring in many 
cases hundred thousands or millions of labeled training samples. Asking Subject Matter 
Experts (SME) to hand-label the required data is not feasible anymore as it would take 
many months and a lot of money. Labeled data have consequently become the development 
bottleneck for real-world AI applications. There are still some indirect or partial solutions to 
this problem:
\begin{description}
%
%
\item \textbf{Crowdsourcing services} Platforms like Amazon Mechanical 
Turk\footnote{\url{https://www.mturk.com/}} enable 
hiring of non-expert workers that can fulfill massive tasks. They also provide high-quality
tools for a precise and efficient labeling process. Although crowd workers will certainly not 
perform same as SMEs, labeling quality might be acceptable for many applications.    
\item \textbf{Online applications} Fancy and entertaining games or other applications usually 
lure many users in short online interactions. One possibility would be to obtain classification 
or labeling of the data by users that randomly interact with such applications.  
\item \textbf{Public datasets} Obtaining labeled data from existing free datasets could be 
the fastest and most economical way for solving the problem. Crawling tools or abilities 
can also help for harvesting large quantities of data from websites. However, copyright issues 
may need to be carefully considered as many public data are distributed with non-commercial
licenses. 
\item \textbf{Model tunning} If there are enough labeled data from public datasets but
their quality is suspicious, one possibility could be to tune the model on progressively 
better data. This way the model is first trained on the bigger quantity of public data. 
Next, it is retrained (and thus tuned) on smaller, professionally labeled data.
\item \textbf{Transfer learning} Sometimes there might be tons of professionally labeled 
data for a similar but not identical task that can be utilized. There are also tools that could 
be used to automatically determine which elements of the source data can be repurposed  
for the new task.
\end{description}
\emph{Data programming} proposed in \cite{ratner2016data} is a novel and more systematic 
paradigm for obtaining large quantities of labeled data efficiently. It is based on higher-level 
supervision over unlabeled data from SMEs (e.g., in form of heuristic rules) for generating 
noisy training samples programmatically via labeling functions. Authors propose learning 
the accuracies of the labeling functions and their correlation structure in the form of a 
generative model for automatically denoising the generated data. They test their solution on 
\emph{relation mention extraction} task applied in texts of news, genomics, pharmacogenomics and 
diseases, reporting very promising results. Data programming might thus become an important 
specialization of data science in the near future.    
\chapter{Emotions in Music} 
\label{chapter3}
\ifpdf
    \graphicspath{{Chapter3/Figs/}{Chapter3/Figs/PDF/}{Chapter3/Figs/}}
\else
    \graphicspath{{Chapter3/Figs/Vector/}{Chapter3/Figs/}}
\fi
%
\begin{flushright}
	\emph{\dq{Some sort of emotional experience is probably the main \\ 
			reason behind most people’s engagement with music.}} \\
		\vskip 0.07in
	\footnotesize{-- P. Juslin and J. Sloboda, \emph{Handbook of Music and Emotion}, 2001}
\end{flushright}
\vskip 0.35in
\indent \indent
Music listening is a universal experience that has been highly influenced by technology.
The transition from personal music player devices to online streaming platforms changed
music industry significantly in the last fifteen years. These platforms boosted personalization
of music listening experience by recommending music to users based on their listening 
history and profile. Retrieval and recommendations of music are based on title, genre, artist
or other metadata, as well as on mood, an important attribute of music. These advances 
promoted research in areas like \emph{Music Information Retrieval} or \emph{Music 
Emotion Recognition}. The later is a form of sentiment analysis in music domain where the 
text is only a partial source of attributes (features). It uses analogous supervised or unsupervised
methods and same constructs like lexicons. 
\par 
We start this chapter by presenting some fundamental concepts that concern music and 
emotions. In Section~\ref{sec:PopularEmoModels} we introduce some popular music emotion 
models proposed by psychologist. Then in Section~\ref{sec:SongFeatureTypes}, various
feature types used for emotion identification in songs are described. 
Section~\ref{sec:CrowdEmotions} presents some crowdsourcing methods for massive collection 
of song emotion labels and discusses their applicability. In Section~\ref{sec:TagDatasets}
we describe the steps and methodology we followed for creating MoodyLyrics4Q and 
MoodyLyricsPN, two relatively big datasets of song emotions. In the end, 
Section~\ref{sec:LableGenerationLexicons} explores the possibility of using a lexicon-based
sentiment analysis method as a generative function of song emotion labels.  
\section{Music and Emotions}  
\label{sec:MusicConcepts}
\subsection{Music Emotions: Concepts and Definitions}
%
In the epoch of online networking and forums, music listening and appreciation 
has become social and collective. Personal CD players and collections of album disks 
kept in drawers are part of history. Subscriptions in streaming Web and mobile platforms 
like \emph{SoundCloud}, \emph{Pandora Radio}, \emph{AllMusic}, 
\emph{Last.fm} or \emph{Spotify} took their place.  
%
The first big shift came in when Apple introduced iTunes and iPod in 2001. They enabled 
users to purchase, download and store digital music files in the fancy and easy to use 
iPod, instead of going to the store for buying the preferred album CDs. The eminent 
transition from local to cloud storage and the proliferation of smart mobile devices created the 
right conditions for transiting to the above-mentioned subscription platforms that offer 
listeners streaming of the favorite tracks instead of buying whole albums. The immediate 
consequence was a steady decrease in album sales. 
Data (songs) are now stored in the cloud and users can create playlists or music preference 
profiles, find songs, like and share them or even suggest them to other peers. 
They also receive music recommendations based on their profiles or previous preferences.
Search and retrieval of songs is based on typical metadata like title, artist or genre, as well 
as on emotions they convey. 
\par 
All these developments were made possible by (and also promoted) significant advances 
in Music Information Retrieval (MIR) and Music Emotion Classification (MEC) research. 
Wheres MIR is the application of information retrieval to find music or songs, MEC can 
be seen as sentiment analysis applied in music domain. Both are based on data mining 
and AI (machine learning and/or deep learning) techniques, utilizing various types of 
music metadata and features. The problem here is that we are dealing with \emph{music}
and \emph{emotions}, two abstract and complex concepts that are study subjects of 
other disciplines like psychology, philosophy or sociology. There are certainly many 
definitions of music and emotions from the perspective of such disciplines. In general, 
music can be described as \emph{an art of sound in time that expresses ideas and emotions 
in significant forms through the elements of rhythm, melody, harmony, and 
color}.\footnote{\url{http://www.dictionary.com/browse/music}} 
%
From our perspective, music or emotion definitions are not that important. 
A more intriguing question is \enquote{Why do people listen to music?}. Authors in \cite{schafer2013psychological} find (by means of statistical evidence) and report that 
the most important motivations for daily listening to music are: 
%
\begin{itemize}
	\item \textbf{Self-awareness} -- \emph{Music listening is highly associated with self-related 
	thinking, emotions and sentiments, absorption and escapism, etc.} 
	\item \textbf{Social relatedness} -- \emph{Music involves people in socializing and affiliation, 
	helping certain individuals to reveal that they belong to particular social groups, connect to 
	their peers, etc.} 
	\item \textbf{Mood regulation} -- \emph{Music can help to get in a positive 
	mood, to stimulate the physiological arousal, to amuse or entertain, etc.}
\end{itemize}
%
Their findings show a strong correlation between music listening and self-regulation of 
mood. People tend to listen to specific kinds of music when they are in a particular
emotional state.  
Music expresses and induces \emph{emotions} and influences \emph{mood}, two 
concepts that are even more abstract and difficult to stipulate.    
%
%
In psychological studies like \cite{tenenbaummeasurement} we find the following 
definitions: 
\begin{itemize}
	\item \textbf{Affect} -- \emph{A neurophysiology state consciously accessible as a 
	primitive, non-reflective feeling, most evident in mood and emotion but always available 
	to consciousness.} 
	\item \textbf{Emotional episode} -- \emph{A complex set of interrelated sub-events related 
	with a specific object.}
	\item \textbf{Mood} -- \emph{The appropriate designation for affective states that are about 
	nothing specific or about everything, about the world in general.}	
\end{itemize}
Something we can spot from these confusing definitions is the fact that moods usually 
last longer in time than emotions. This has also been stressed in the literature. 
Nevertheless, from our perspective, the two terms are quite similar. As a result, throughout 
this thesis, they are used interchangeably. 
%
%
%
%
%
%
\subsection{Models of Music Emotions}
\label{sec:PopularEmoModels}
%
%
\begin{figure}
	\centering
	\includegraphics[width=0.6\textwidth]{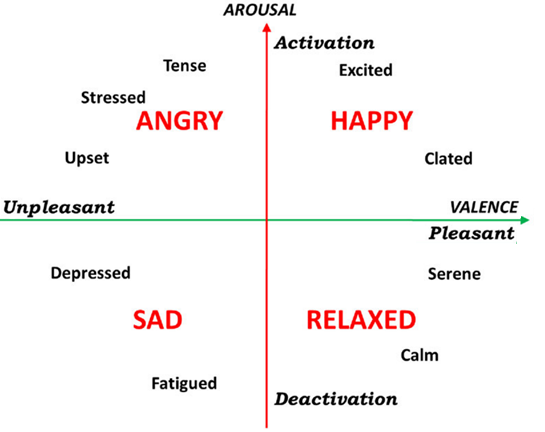}
	\caption{Model of Russell}
	\label{RussellModelFig}
\end{figure} 
\begin{figure}
	\centering
	\includegraphics[width=0.5\textwidth]{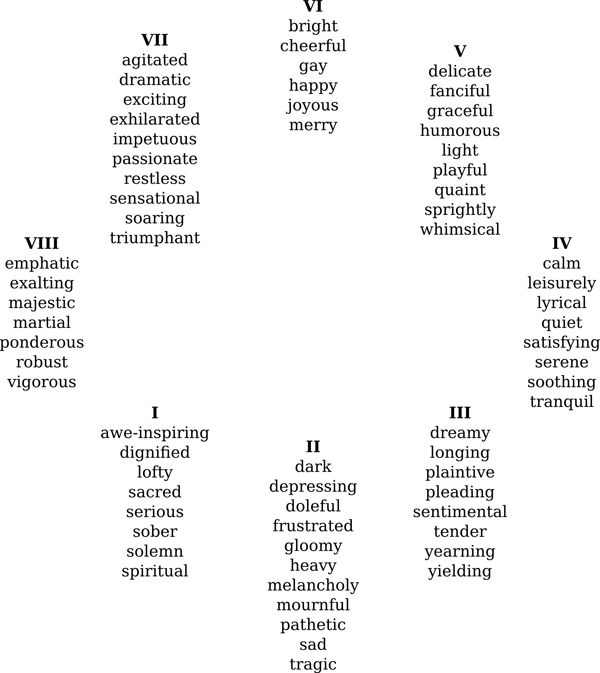}
	\caption{Model of Hevner}
	\label{HevnerModelFig}
\end{figure} 
\begin{table}
	\caption{Mirex emotion clusters}
	\centering
	\begin{tabular}{l | l}   
		\toprule	
		\multicolumn{1}{c}{Cluster} & \multicolumn{1}{| c}{Emotion Adjectives} \\
		\midrule
		Cluster 1 & passionate, rousing, confident, boisterous, rowdy \\
		
		Cluster 2 & rollicking, cheerful, fun, sweet, amiable/good natured \\
		
		Cluster 3 & literate, poignant, wistful, bittersweet, autumnal, brooding \\
		
		Cluster 4 & humorous, silly, campy, quirky, whimsical, witty, wry \\
		
		Cluster 5 & aggressive, fiery, tense/anxious, intense, volatile, visceral \\
		\bottomrule
	\end{tabular}
	\label{5ClassModelTable}
\end{table}
%
%
It is not easy to elaborate or describe music emotions. Actually, there are big variations in 
taxonomies and the terminology that is used. For example, when searching for music by mood
in \emph{AllMusic} website, the user is confronted with about 570 
descriptors.\footnote{\url{https://www.allmusic.com/moods}} Reducing this excessive complexity 
in a manageable set of categories is an essential prerequisite. Psychologists have developed 
models or taxonomies of music emotions that are very important for alleviating the problem.
%
Two types of music emotion models can be found in the literature: categorical and dimensional. 
Categorical taxonomies describe music emotions using short text labels or descriptors. Those 
labels that are enough close semantically (synonyms) are clustered together to form a music 
emotion category. Contrary, labels representing contrasting emotions should appear in different
and distant categories. Dimensional models, on the other hand, represent emotions using numeric 
values of certain parameters (dimensions) in a continuous space. Valence and arousal are two 
typical dimensions that are commonly used in different dimensional models. No common 
agreement exists about what type of models are best in various situations. There are 
however some of them that have gained wide popularity in the research literature.  
\par 
One of the earliest studies in this field was conducted by Hevner in 1936 \cite{Hevner1936}. 
She presented a categorical model of 66 emotion labels clustered in eight categories as shown
in Figure~\ref{HevnerModelFig}. Obviously, there is high synonymy between labels of same 
class and dissimilarity (sometimes antonymy) between labels of different classes. 
The model of Hevner has not been used widely in its original form. Nevertheless, it is considered 
as an important starting point for constructing other music emotion representation models. 
%
A more recent categorical model proposed in \cite{hu2007exploring} organizes music 
emotion labels in five classes as shown in Table~\ref{5ClassModelTable}. 
%
Despite the semantic overlapping between clusters 2 and 4 reported in \cite{laurier2007audio},
that model has been widely used in several MEC tasks. 
%
One of the most popular dimensional models was proposed by Russell 
in 1980 \cite{russell1980circumplex}. It represents music emotions as points in a 
two-dimensional plane of valence and arousal dimensions as shown in 
Figure~\ref{RussellModelFig}. \emph{Valence} (pleasant - unpleasant) represents 
how much an emotion is perceived as \emph{positive} or \emph{negative} whereas 
\emph{Arousal} (sleepy - aroused) indicates 
how strongly the emotion is felt. Russell's model is simple, intuitive and highly used in 
research literature. There are also other models that add more dimensions to represent
music emotions. 
\subsection{Sentiment Analysis of Songs}
\label{sec:SongFeatureTypes}
%
Sentiment analysis of songs is about utilizing data mining or AI techniques in combination 
with different features to correctly classify songs in mood categories based on the most typical 
emotion types they express. Identifying emotions in songs is however complex and difficult. 
The main trouble is the subjective nature of music perception. Songs are perceived in various 
ways from different subjects. Another difficulty comes from the heterogeneous form of features 
that are found in songs. Besides audio which is the most important component, there is also 
text (lyrics) and other types of attributes like instrumentation, genre, epoch, etc. The different 
approaches are usually identified by these features they mostly process. In the literature, the 
most popular feature types are: 
%
%
%
%
\begin{itemize}
	\item \textbf{Tags} -- Social tags such as \enquote{mellow}, \enquote{bittersweet}, 
	\enquote{cool}, \enquote{90s} and many more, may be useful for mood, genre or 
	instrument recognition tasks.
	\item \textbf{Lyrics} -- Sentiment analysis of song lyrics can also be highly convenient 
	for emotion identification in musical tracks.
	\item \textbf{Audio} -- Techniques that work with sound were the earliest explored and 
	are based on acoustic features like cepstral coefficients, energy distribution, etc.  
	\item \textbf{Hybrid} -- For higher accuracy some researchers mix various features 
	in certain ways and construct multimodal solutions.
\end{itemize}
%
In the context of music listening, a tag is just a free text descriptor provided by any user 
for describing a musical object such as a song, album or artist. Some typical examples 
are \dq{rock}, \dq{melodic}, \dq{bass}, etc. Today's music portals collect 
many of them for free on a daily basis. The power of these tags lies in the detailed and 
descriptive information they carry. Tags are semantically rich and easy to work with. In 
the literature, social tags are frequently analyzed to build and test folksonomies of music 
emotions. 
%
Authors of \cite{saari2014semantic} for example, propose a method they call Affective 
Circumplex Transformation (ACT) for transforming the planar model of Russell into 
a space of \emph{valence}, \emph{arousal} and \emph{tension}. This new space enables 
a better representation for emotionality of music tags and tracks.   
%
Also in \cite{hu2007creating}, authors exploit \emph{Last.fm} tags of emotions
for creating a simple representation space of three clusters only. Their emotion space 
seems oversimplified and has not gained popularity. Nevertheless, their approach can be 
considered as a valuable guideline for similar works. 
Despite their advantages, social tags have drawbacks as well. The absence of a common 
vocabulary of tags creates ambiguity and problems such as polysemy. Furthermore, 
user tags do frequently contain irrelevant feedback (noise) and thus require careful 
preprocessing. 
\par 
Same as tags, song lyrics represent another source of high-level features. They are 
easily processed and usually found for free (contrary to audio that is mostly copyrighted). 
Studies utilizing song lyrics usually fall into two categories: lexicon-based and corpus-based.
The former try to predict emotionality by mapping lyrics words with their synonyms in 
affect lexicons. 
%
ANEW (Affective Norms for English Words) presented in \cite{bradley1999affective} 
is one of such affect lexicons that has been widely used. It provides a set of normalized emotional 
scores for the 1040 English words it contains. Those words were rated in terms of 
\emph{valence}, \emph{arousal} and \emph{dominance} from many human subjects
who participated in the psycholinguistic experiments. Dominance is an emotion dimension 
that represents the scale of ascendancy (vs. submissiveness) a word induces. 
%
Authors in \cite{oh2013music} make use of ANEW terms to find the emotional category 
of the intro and refrain parts of song lyrics. They assume that intro and refrain are the most 
emotionally significant parts of song lyrics. 
%
Furthermore, in \cite{Hu09lyric-basedsong} authors create and use ANCW, the Chinese
version of ANEW. They compute valence and arousal of each word appearing in song lyrics 
and afterwards the aggregate values of each sentence. For integrating values of all sentences 
and deriving the emotion category of entire song, fuzzy clustering is utilized.  
\par 
Corpus-based approaches utilize collections of mood-annotated lyrics to train models. 
In this respect they represent pure supervised learning solutions. One such study is 
\cite{van2010automatic} where authors show that word oriented metrics like term frequency
or tf-idf provide important insights for automatic mood classification of lyrics. Feeding such
features in traditional classifiers they train a model that is able to predict emotionality  
of unlabeled songs. Despite being free and easy to work with, lyrics are not available for  
some types of music like instrumental, classical etc.   
%
In contrast with that, audio is the universal component in all types of music. It is also 
very rich in various feature types. 
%
Authors in \cite{conf/ismir/HanRDH09} experiment with loudness, pitch, tempo, 
tonality, harmonics, key, and rhythm. Support vector regression is used to map 
features with emotion categories and 94.55\% accuracy is reported.  
There are still difficulties when working with audio. Expertise in musical concepts 
and signal processing is required at some scale. There are also performance limitations 
related to the low-level audio features. Their semantic gap with the high-level user 
perception reduces accuracy and applicability.   
%
To mitigate the cons of the above feature types, some studies like \cite{hu2010improving}
or \cite{Yang2008} employ feature aggregation strategies. It is typical for example to 
combine tags with audio or lyrics. Sometimes other types of metadata like artist, genre 
or instrumentation are mixed in.   
\section{Crowdsourcing Song Emotions}   
\label{sec:CrowdEmotions}
The recent tendency towards intelligent models that learn from data does not exclude
music domain. As explained in the previous section, many researchers prefer supervised learning 
strategies for emotion recognition in songs. However, one difficulty they commonly face
(and we faced) is the lack of song datasets with emotion labels correctly assigned. 
Manual annotation of emotions to musical tracks is time-consuming, costly and labor intensive. 
Crowdsourcing is a recent network-powered work paradigm that may solve this problem.
%
As a term, it was first coined by Jeff Howe in \cite{howe2006rise}. He describes cases
in which this distributed labor approach has proven highly successful not only for lowering 
production costs but also for finding innovative R\&D solutions. The fundamental elements 
that make crowdsourcing work well are independence and variety of opinions, work 
decentralization, and opinion aggregation \cite{surowiecki2004wisdom}. The interested parties (employers, organizations, institutions, researchers, etc.) publish job requests or unsolved 
problems in online platforms that serve as marketplace networks. 
\par 
Some of these platforms (e.g., InnoCentive, NineSigma, iStockphoto or YourEncore) target 
specialized or talented subjects, especially for addressing R\&D creative problems in specific areas. 
%
%
Other platforms like Amazon Mechanical Turk address simple and repetitive tasks that any subject 
with Internet access can elaborate. In MTurk participants are paid by the publisher of those tasks after 
successfully completing them. MTurk is thus highly suitable for activities which require massive social 
involvement like emotion annotation of a high number of tracks. Other forms of crowdsourcing
campaigns are conceived as challenges with bountiful rewards like Netflix \$1M Prize. Back in 
2006, Netflix requested an algorithm for movie recommendations with error rate 10\% lower 
than state-of-the-art and offered the money to the team that would propose it first.  
That challenge had a positive public impact, boosting research in the field of recommender systems 
which are now part of almost every commercial or advertising web platform. 
The suitability of crowdsourcing alternatives for experimental microtasks such as statistical 
surveys or data collection has attracted interest from researchers, including those that work in 
music emotion recognition. Some of the most explored alternatives for collecting emotion
labels of songs are:   
\begin{description}
\item[MTurk] As mentioned above, this marketplace is probably the most popular for tasks 
of massive involvement. Many researchers use it for gathering feedback about emotionality of 
songs. Authors of \cite{mandel10b} for example, involve MTurk workers for collecting
tags of various types like mood, instrument, genre, etc. After analyzing the data,
they conclude that different intervals of the same song are usually described differently from 
users. There are also authors that use MTurk not only for creating datasets, but also 
compare labeling quality with that of other methods to assess the viability of MTurk. 
In \cite{speck2011comparative} for example, they compare annotations crowdsourced from 
MTurk with those obtained from MoodSwings, a collaborative game they developed. Both 
annotations follow an emotional model of four categories derived from the planar 
model of Russell. Based on their results, authors report accordance between MoodSwings
and MTurk data, concluding that the later is an applicable method for song annotation.  
Furthermore in \cite{Lee:2012:GGT:2232817.2232842}, MTurk labels are contrasted 
with the ones collected from MIREX\footnote{\url{http://www.music-ir.org/mirex/wiki/MIREX_HOME}}
campaign. Authors conclude that agreement rates are satisfactory and MTurk 
crowdsourcing can serve as a practical and inexpensive alternative for creating 
music mood ground truth datasets.    
\item[Online games] Fancy and amusing online applications like games may represent
an interesting option for crowdsourcing opinions. Online users are certainly more inclined 
to play games than answer survey questions.
MajorMiner\footnote{\url{http://majorminer.org/info/intro}} described in \cite{mandel2008web} 
was one of the first online games specifically designed to gather opinions about emotions 
of certain songs. Users first listen to ten-second clips and then select the most representative 
tags from a predefined list. Authors compare tags collected from their game with those 
obtained from \emph{Last.fm} music portal. They conclude that MajorMiner tags are less 
noisy and can serve for creating emotion datasets of songs.      
TagATune is another collaborative game developed to crowdsource music clip labels
\cite{law2003tagatune}. Here players are involved in a rich audio experience and 
coupled with a partner for tagging tunes agreeably. Authors pretend that TagATune
is more effective than MajorMiner because of its entertaining features.   
MoodSwings mentioned above was specifically developed to collect mood tags of songs.
It is more effective in tag collection as it makes users provide ratings on a per-second basis. 
User ratings are then converted in labels using the valence-arousal planar model of emotions. 
\item[Social tags] From the different music listening platforms, \emph{Last.fm} is 
probably the most popular among academics. This is because of the open API it provides 
for collecting and analyzing tags, metadata and musical preferences of its users. 
Consequently, \emph{Last.fm} user tags appear as an important research resource in 
many academic works. In general, users tend to provide tags about songs and other online 
objects for several reasons. Creation of assistance for future searches, expression of opinions 
and social exposure are some of the most important \cite{Ames:2007:WWT:1240624.1240772}.
%
One of the first studies that examined type distribution of song tags is
\cite{doi:10.1080/09298210802479284}. According to this study, 68\% of tags are related to 
song genre, 12\% to locale, 5\% are mood tags, 4\% of them are about instrumentation tags and 4\% express opinion.  
In \cite{Hu2007} on the other hand, we find one of the first studies that specifically 
examined mood social tags. Authors report an unbalanced distribution of emotion tag 
vocabulary. They also infer that many labels are interrelated or reveal different views 
of a common emotion category. 
In \cite{7536113} authors utilized \emph{AllMusic} tags to create a ground truth dataset of 
song emotions. They first used tags and their norms in ANEW to categorize each song in one 
of the four valence-arousal quadrants of Russell's model. Afterwards, three persons  
validated and improved annotation quality manually. The resulting dataset has 771 songs and 
their corresponding emotion category. 
\item[Other] Research papers explore many other strategies for collecting opinions about songs
like online web services, traditional surveys etc. An example is Songle: a web service with 
music visualizations presented in \cite{DBLP:conf/ismir/GotoYFMN11}. It enables users 
to play music and associate each played track with corresponding visualizations of beat 
structure, vocal melody, chords and other characteristics that are sometimes erroneous. 
Users that spot errors in visualizations can provide corrections which are shared for 
improving the experience of future users. 
\end{description}
The high number of works in the literature that are exploring crowdsourcing options emphasizes 
the growing importance of the network-powered crowd intelligence. A more detailed 
discussion about the crowdsourcing alternatives discussed in this section can be 
found in \cite{pub2677905}. 
\section{Creating Song Datasets from Social Tags}   
\label{sec:TagDatasets}
In the previous section, we discussed different alternatives for collecting emotion labels
about songs. Here we present the creation steps of two relatively big song emotion
datasets, MoodyLyrics4Q and MoodyLyricsPN. Firstly, various existing  
datasets together with their limitations are described. Afterwards, we illustrate the 
systematic process for dataset creation. 
\subsection{Existing Song Emotion Datasets}
%
The need to experiment with emotionally annotated songs has motivated researchers 
to utilize various strategies for creating music ground truth datasets. Such datasets should 
possess the following characteristics:
%
\begin{enumerate}[label=\textbf{\arabic*}.]    
	\item \emph{Contain as many songs as possible (e.g., more than 1000)}
	\item \emph{Annotated following a well-known model of emotions}
	\item \emph{Have polarized annotations to be usable as ground truth}
	\item \emph{Publicly released for cross-interpretation of experiments}
\end{enumerate}
%
The above characteristics are conflicting and hardly achieved together. 
Due to the subjective nature of music appraisal, complete cross-agreement between 
different subjects involved as annotators is hardly achieved. Hiring several music 
professionals for manual annotation would certainly produce high-quality data. 
However, it would also be time-consuming and probably very costly.  
Actually, organizations like \emph{Pandora} employ music 
experts for annotating tracks with relevant emotion words. Nevertheless, they are not 
willing to share their datasets for public use.
Many researchers have explored crowdsourcing alternatives discussed in the previous section 
for the sole purpose of constructing labeled song datasets. Authors in \cite{hu2009lyric}
crawl \emph{Last.fm} tags and use them for constructing a big dataset of 5296 songs 
dispersed in 18 emotion categories (synonymous tags). Annotation is automatically 
performed using a binary method (tagged, not tagged) for each song and emotional 
category. This dataset could be convenient for various types of music emotion recognition
experiments. Nevertheless, it has not been released to the public. 
%
Also in \cite{7536113}, authors created a public and polarized dataset of songs
using \emph{AllMusic} tags and following the popular valence-arousal model of Russell. 
However, that dataset is relatively small, consisting of 771 tracks only. 
\par 
Another music dataset is described in \cite{soleymani20131000}. 
Authors involved many MTurk workers to annotate songs according to 
valence-arousal planar model. They claim that each clip has been labeled by
at least ten workers which guarantees high polarity and data annotation quality. 
The final dataset contains 744 entries and is publicly released as a whole 
(clips and features). 
Also in \cite{4432652}, a traditional survey based on questions to paid participants is 
used to obtain emotional labels. The resulting dataset is public and consists of 500 
(small) western songs. 
A multimodal dataset containing text and audio features of 100 songs (very small) is presented 
in \cite{Mihalcea:2012:LME:2390948.2391015}. Authors collect emotion labels from 
MTurk workers and provide the dataset upon request for academic use only. 
Several other research works create close datasets that are used to evaluate methods or 
algorithms they propose.  
%
Though hard to believe, we could not find any experimentation dataset that fulfills all 
four requisites listed above. In the following sections, we describe the steps we followed 
for creating two such datasets ourselves. MoodyLyrics4Q is a dataset of 2,000 songs, fully 
compliant with the four requisites listed in the previous section. We also created 
MoodyLyricsPN, a bigger collection of 5000 songs labeled as \emph{positive} or 
\emph{negative} only (violation of second requisite). 
%
%
\subsection{Folksonomy of Emotion Tags}
\label{sec:TagFolksonomy}
%
%
\begin{figure}
	\centering
	\includegraphics[width=0.65\textwidth]{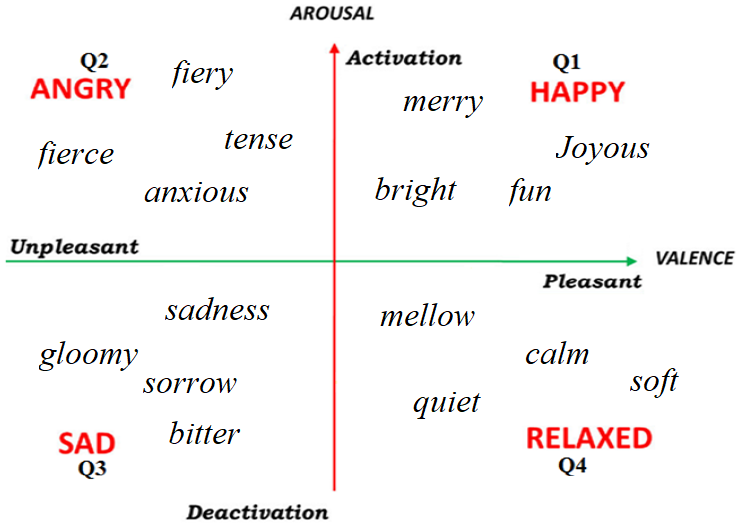}
	\caption{Adopted model of tag emotion categories}
	\label{fig:Russell4Q}
\end{figure} 
%
For the annotation of the songs, we decided to crawl social tags from \emph{Last.fm}.
To comply with the second requisite of the previous section, we had to find a commonly
used emotional model. Apart from the few popular models of psychologists discussed in 
Section~\ref{sec:PopularEmoModels}, there are also folksonomies of music emotions built in 
research works starting from social tags about songs. 
In \cite{conf/ismir/BischoffFPNLS09} for example,
they perform clustering on \emph{AllMusic} mood tags and aggregate them in four 
classes that are analogous to the four quadrants of Russell. A similar work utilized 
\emph{Last.fm} emotion tags \cite{conf/ismir/LaurierSSH09}. Authors apply 
unsupervised clustering and Expected Maximization algorithm to document-tag
matrix. They report four as the optimal number of emotion tag clusters. Moreover, 
their four clusters (\emph{happy}, \emph{angry}, \emph{sad}, \emph{relaxed}) 
are again analogous to the four quadrants of Russell's model. 
%
All these research results confirm that categorical models of song emotions that are 
derived from social tags do comply with the theoretical models of psychologists and 
are applicable for sentiment analysis of songs. They also convinced us that among 
the various models of emotions, the categorical version of Russell's model with one 
emotional category for each quadrant (\emph{happy} for Q1, \emph{angry} for Q2,
\emph{sad} for Q3 and \emph{relaxed} for Q4) is the most simple, intuitive, widely 
recognized and practical for our purpose. A graphical illustration of the model is shown 
in Figure~\ref{fig:Russell4Q}. 
\begin{table}
	\caption{Four clusters of tag terms}
	\centering
	\begin{tabular}{l l l l}   
		\toprule	
		\multicolumn{1}{l}{\textbf{Q1-Happy}} & \multicolumn{1}{l}{\textbf{Q2-Angry}} &
		\multicolumn{1}{l}{\textbf{Q3-Sad}} & \multicolumn{1}{l}{\textbf{Q4-Relaxed}}  \\
		\midrule
		happy & angry & sad & relaxed \\
		happiness & aggressive & bittersweet & tender\\
		bright & fierce & bitter & soothing \\
		joyous & outrageous & sadness & mellow \\	
		cheerful & rebellious & depressing & gentle \\	
		fun & anxious & tragic & peaceful \\
		humorous & fiery & gloomy & soft \\
		merry & tense & miserable & calm  \\
		exciting & anger & funeral & quiet \\
		silly & hostile & sorrow & delicate \\
		\bottomrule
	\end{tabular}
	\label{table:ClassModel}
\end{table}
\par
\begin{figure}
	\centering
	\includegraphics[width=0.55\textwidth]{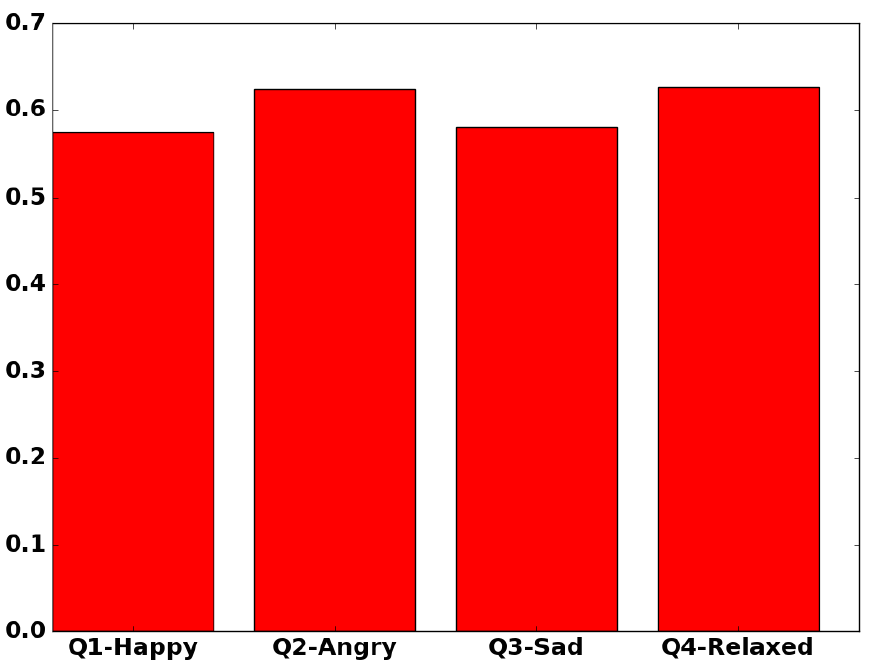}
	\caption{Intra-cluster similarity of tags}
	\label{fig:2Moods}
\end{figure}

\begin{figure}
	\centering
	\includegraphics[width=0.6\textwidth]{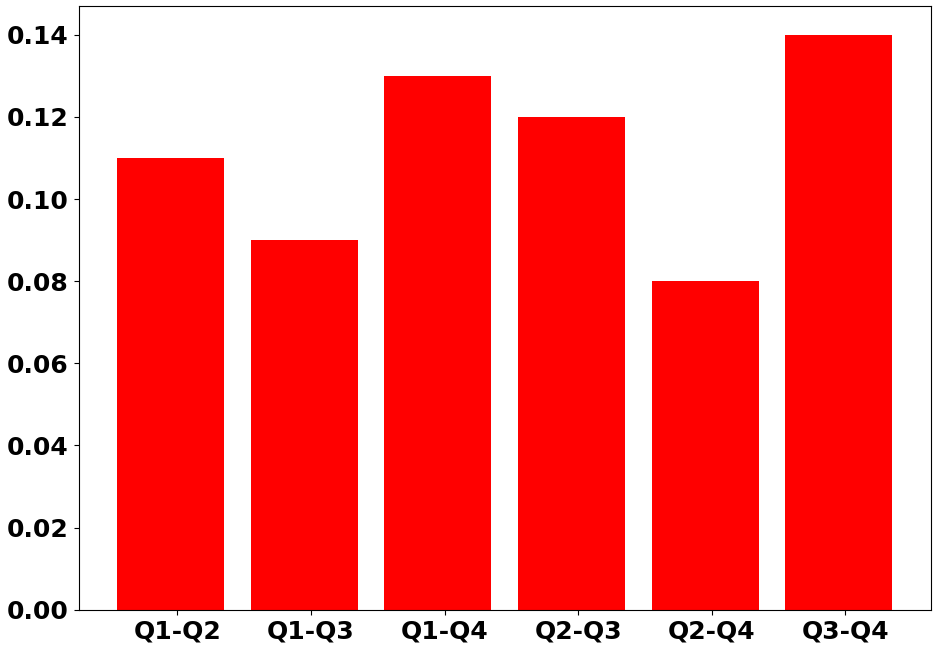}
	\caption{Inter-cluster similarity of tags}
	\label{fig:ClusterDissimilarity}
\end{figure}

 
For organizing tags, we constructed a folksonomy of terms that is very similar to the 
one of \cite{conf/ismir/LaurierSSH09}. First, we retrieved 150 mood terms 
from relevant research papers and the mood terms from 
\emph{AllMusic}\footnote{\url{https://www.allmusic.com/moods}} portal. A preliminary
selection process was conducted manually to keep in only terms that clearly fall in one 
of the four categories of our model and filter 
out ambiguous ones. We consulted ANEW valence and arousal norms of each word 
for objectivity in selection. The quality of a folksonomy of terms can be measured by 
the average intra-cluster (as high as possible) and inter-cluster (as high as possible) similarities. 
For this reason, we utilized word embeddings trained from a corpus of 2 billion tweets with Glove 
method.\footnote{\url{http://nlp.stanford.edu/data/glove.twitter.27B.zip}}  
Word embeddings are popular for their ability to capture semantic similarities 
between words (see Section \ref{sec:WordEmbIntro}). The average intra-cluster similarities 
were optimized by trying a high number of tag combinations inside each of the four 
clusters. The optimal configuration resulted the one shown in Table~\ref{table:ClassModel}
which comprises the ten most suitable mood tags in each cluster. Intra-cluster
similarities are presented in Figure~\ref{fig:2Moods}. Inter-cluster similarities, on the 
other hand, are shown in Figure~\ref{fig:ClusterDissimilarity}. As we can see from that 
figure, the less similar (or more dissimilar) clusters are Q1-Q3 and Q2-Q4 (the diagonals
in the plane) which differ in both valence and arousal. 
%
\subsection{Data Processing and Annotation}
%
To produce a large final set of labeled songs (first requisite) we imported all tracks of
\emph{Million Song Dataset} presented in \cite{Bertin-mahieux11themillion}. It 
is one of the biggest song collections, created to test the scalability of algorithms to commercial
sizes. We mixed in the records of Playlist dataset as well. This is a smaller collection 
(75,262 tracks) of more recent songs \cite{Chen:2012:PPV:2339530.2339643}. At this point,
a total of 1018596 tracks was reached. Data processing went on removing duplicate tracks. 
Afterwards, we crawled all tags of each track utilising 
\emph{Last.fm API}.\footnote{\url{https://www.last.fm/api/show/track.getTags}} 
Songs with no tags were removed and statistical analysis of tags was performed. The most 
frequent tag was \emph{rock} appearing 139295 times, followed by \emph{pop} with 
79083 occurrences. We also analyzed tag type frequencies. Genre tags were the most
common with 36\% of the total, followed by opinion (16.2\%) and mood (14.4\%) tags. 
\par
\begin{figure}
	\centering
	\includegraphics[width=0.65\textwidth]{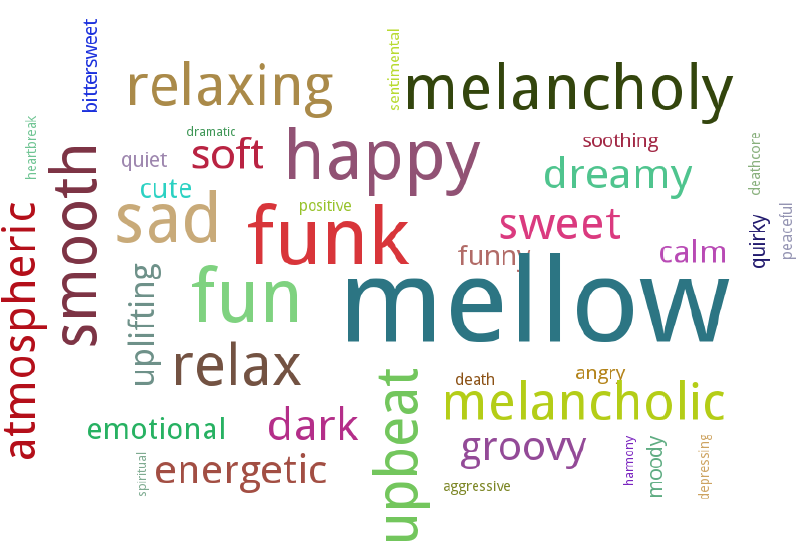}
	\caption{Word frequency cloud of mood tags}
	\label{fig:TagCloud}
\end{figure}

  
Among mood tags, \emph{mellow} was the most frequent with 26,890 occurrences,
followed by \emph{funk} (16324) and \emph{fun} (14777). The word cloud of 
mood tags is shown in Figure~\ref{fig:TagCloud}. There was an obvious bias
towards positive emotion tags. This is probably because people are more inclined to 
give feedback when they listen to positive songs. Popularity bias may be another reason. 
After concluding the analysis of tag statistics, we moved on removing every tag that was 
not about mood or other tags that were ambiguous (e.g., we could not know if tag 
\emph{love} means the user loves that song or he/she thinks it is about love). 
At the end of this phase, we reached to 288708 tracks. Further details about data processing 
steps and tag statistics can be found in \cite{pub2669975}.     
%
Next, we identified and counted emotion tags of each cluster appearing in the 
remaining tracks. Four counters (one per emotion cluster) were obtained for every track.
To reach to a polarized collection of songs (third requisite) we used a tight annotation scheme.
A track is set to quadrant $Qx$ if it fulfills one of the following conditions: 
\begin{itemize}
	\item has 4 or more tags of $Qx$ and no tags of any other quadrant
	\item has 6 up to 8 tags of $Qx$ and at most 1 tag of any other quadrant
	\item has 9 up to 13 tags of $Qx$ and at most 2 tags of any other quadrant
	\item has 14 or more tags of $Qx$ and at most 3 tags of any other quadrant
\end{itemize}
Songs with fewer than four tags or those not fulfilling any of the above conditions were discarded.
%
This scheme guarantees that even in the worst case scenario (song tag distribution), any song 
set to Qx quadrant has more than 75\% of all its received tags being part of that quadrant. 
What remained was a collection of 1986 happy or $Q1$, 574 angry or $Q2$, 783 sad or $Q3$ and 
1732 relaxed or $Q4$ songs for a total of 5075 (2,000 after balancing). 
\par 
Datasets with \emph{Positive vs. Negative} representation are clearly oversimplified and 
do not reveal much about song emotionality. However such kind of datasets could be used for 
various experimental purposes. We merged $Q1$ with $Q4$ (\emph{happy} with \emph{relaxed}) 
considering them as \emph{positive}, and $Q2$ with $Q3$ (\emph{angry} with \emph{sad}) 
for the \emph{negative} category. The corresponding tags of each cluster were 
recombined as well. As binary discrimination is easier, an even tighter annotation scheme
was enforced. A track is considered to belong to Qx (\emph{positive} or \emph{negative}) 
only if:  
\begin{itemize}
	\item it has 5 or more tags of $Qx$ and no tags of the other category
	\item it has 8 up to 11 tags of $Qx$ and at most 1 tag of the other category
	\item has 12 up to 16 tags of $Qx$ and at most 2 tags of the other category
	\item has 16 or more tags of $Qx$ and at most 3 tags of the other category
\end{itemize}
%
This scheme guarantees that even in the worst case scenario (song tag distribution), 
any song labeled as \emph{positive} or \emph{negative} has more than 85\% of all its 
received tags being part of that category. 
We got a collection  
of 2589 \emph{negative} and 5940 \emph{positive} songs, for a total of 8529 (5,000 
after balancing). Apparently, the resulting datasets are imbalanced towards positive songs, 
same as the corresponding emotion tags they were derived from.
%
%
To have an idea about the quality of the first labeling scheme that was used, we compared 
our labels of the first dataset (ML4Q) with those of another one considered as ground-truth. 
The most appropriate for our purpose was the dataset (here A771) described in \cite{7536113}. 
It consists of 771 songs labbeled according to the planar 
model of Russell, same as we did. Authors used \emph{AllMusic} tags for the process 
and involved three persons to validate the annotation quality. The problem is 
however the size of this dataset. From the 771 songs it contains, only 117 were part 
of our initial collection of 5075 labeled tracks.  
\begin{table}
	\caption{Confusion matrix between A771 and ML4Q datasets}
	\centering
	\begin{tabular}{c | c c c c}   
		\toprule	
		\multicolumn{1}{c |}{A771~\textbackslash~ML4Q} & \multicolumn{1}{c}{Happy} &
		\multicolumn{1}{c}{Angry} & \multicolumn{1}{c}{Sad} &  
		\multicolumn{1}{c}{Relaxed} \\ [0.2ex] 
		\midrule
		Happy & \textbf{97.43} & 0.85 & 0 & 1.7 \\ [0.5ex] 
		Angry & 0.85 & \textbf{98.29} & 0.85 & 0 \\ [0.5ex] 
		Sad & 0 & 0.85 & \textbf{97.43} & 1.7 \\ [0.5ex] 
		Relaxed & 1.7 & 0 & 1.7 & \textbf{96.58} \\	
		\bottomrule
	\end{tabular}
	\label{table:ML4QconfusionMatrix}
\end{table}
\par 
In Table~\ref{table:ML4QconfusionMatrix} we show the confusion matrix between 
labels of our dataset and those of A771 for each category. As we can see, the overall 
agreement between the two datasets is 97.28\%. Despite the fact that this result is based
on a small portion of the records, it seems to be high enough to confirm the validity of 
our method.   
%
Both datasets presented here can be freely downloaded from our group 
website.\footnote{\url{http://softeng.polito.it/erion/}}
Lyrics or metadata of their songs can be easily retrieved from online music websites. 
Audio is usually copyrighted and hard to find. Researchers who have access to audio of 
songs can experiment with sound features as well.    
\section{Music Data Programming via Lexicons} 
\label{sec:LableGenerationLexicons}
\begin{figure}
	\centering
	\includegraphics[width=0.70\textwidth]{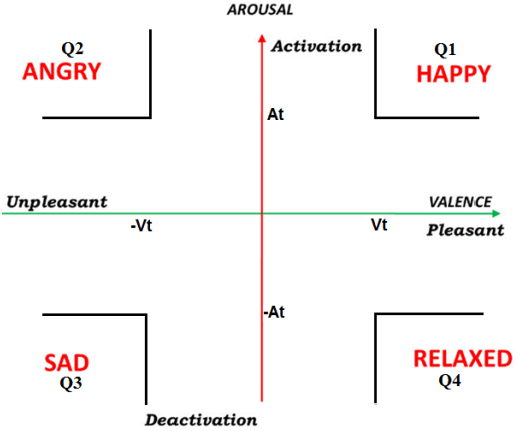}
	\caption{Planar model for emotion categories of texts}
	\label{fig:VAnorms4Q}
\end{figure}

%
The two datasets we created may be big enough to feed traditional machine learning
algorithms. However, they are still small for deep neural networks. Actually, it is 
difficult to collect data in music domain as songs are usually copyrighted. 
Alternative implementations of data programming introduced in Section~\ref{sec:LabeledHunger} 
might be good options for constructing bigger datasets of emotional categories. In this 
section we present the results of some experiments we conducted with a text emotion 
identification method that was used as a generative function of mood labels.   
We also observed the quality of generated labels by comparing them with a benchmark
dataset.  
%
%
The basic method for text sentiment identification is described in \cite{Dodds2010} 
where authors illustrate its use for computing overall positivity of large-scale texts such 
as song lyrics, blog posts, etc. It is based on utilization of valence norms for each word found in 
ANEW lexicon. The norm of each word appearing in the text under analysis is summed 
and then the total is divided by the total number of content words to get the average. Authors 
utilize this simple and fast technique for estimating overall positivity in song lyrics of different 
epochs. To construct a dataset of four emotion categories we can use both valence 
and arousal norms of ANEW and compute their totals for each song text with the following
equations:
\begin{equation}
v_{lyric} = \sum\limits_{i = 1}^{n} v_i f_i ~/~ \sum\limits_{i = 1}^{n} f_i 
~~~~~~~~~~
a_{lyric} = \sum\limits_{i = 1}^{n} a_i f_i ~/~ \sum\limits_{i = 1}^{n} f_i 
\label{equ:VAtotals}
\end{equation}
Here $v_{lyric}$ and $a_{lyric}$ represent valence and arousal of all words in the text 
that also appear in ANEW. Also, $f_i$ is their frequency in the text wheres $v_i$ and 
$a_i$ represent ANEW valence and arousal norms of each text word. Because the values 
are real numbers from 1 to 9, we adapted valence-arousal planar model as shown in 
Figure~\ref{fig:VAnorms4Q}.  
\par
\begin{figure}
	\centering
	\includegraphics[width=0.70\textwidth]{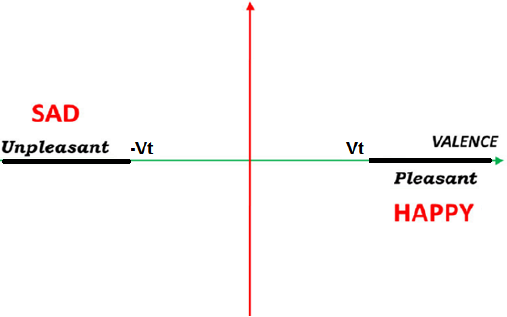}
	\caption{Planar model for emotion polarity of texts} 
	\label{fig:VAnorms2Q}
\end{figure}

 
Obviously, the 1 -- 9 interval was transformed to get zero-centered values ranging
from -4 to 4 and then use the sign value as the discriminator. Aggregate $v_{lyric}$ 
and $a_{lyric}$ values of a text computed with Equation~\ref{equ:VAtotals} provide a point in 
the plane that falls in one of the quadrants. The corresponding emotion category is assigned to 
the text. To avoid misclassification of the points appearing near the origin (values close to 0), 
threshold valence and arousal ($Vt$ and $At$) values are used. The rectangular zone 
$[(Vt,At), (-Vt,At), (-Vt,-At), (Vt,-At)]$ is considered as \dq{unknown}. One of the four 
labels is assigned to each text only if its point falls in one of the quadrants indicated in
Figure~\ref{fig:VAnorms4Q}. For example, a text is labeled as \dq{sad} only if  
$v_{lyric} < -Vt $ and $a_{lyric} < -At$. If we are interested in text positivity only,
we can use just the first formula of Equation~\ref{equ:VAtotals} to compute $v_{lyric}$. 
This value represents a point which falls somewhere in valence axis of Figure~\ref{fig:VAnorms2Q}.
If it is enough displaced (black bars in the figure) the text takes the corresponding polarity.  
%
A problem with the above method is that it does not work well for texts containing
just a few words that are part of ANEW. Given that song lyrics usually contain slang or 
rare words, we faced the need to extend ANEW. A much bigger and generic English 
lexicon is WordNet which contains at least 166,000 \emph{(word, sense)} pairs \cite{Miller:1995:WLD:219717.219748}. Words in WordNet
have synonymy relations which other and word senses are sets of synonyms called 
synsets. WordNet-Affect, on the other hand, is a highly reduced subproduct of Wordnet 
that contains emotion terms \cite{Strapparava2004}. To overcome the size problem 
of ANEW, we decided to combine the three lexicons in the following way. 
For each ANEW word, we checked WordNet synsets that include that word and extended
it with the resulting synonyms. All imported words from WordNet took valence 
and arousal values of the ANEW source word. Afterwards, we kept only those words
that belong to synsets of WordNet-Affect labeled as \emph{Mood}, \emph{Sensation} 
or \emph{Emotion}. All added words of other synsets were removed. This way we reached 
a set of 2162 words which is more than double size of ANEW. In \cite{Hu:2010:IMC:1816123.1816146} authors extend ANEW in a similar way to 
experiment with heterogeneous text features. 
\par 
We evaluated labeling quality of the method applying it to an existing dataset of 
emotionally labeled songs. 
%
Once again, we used the dataset described in \cite{7536113} as external ground-truth.   
We applied the method described above on each song text generating 
the new labels which were compared with those of the dataset. First, $Vt=0.25$ and 
$At=0.25$ were used and the agreement was low. Increasing $Vt$ and $At$ increments 
polarization of generated labels but also reduces their quantity. This is because more lyrics 
start to fall inside the \dq{unknown} zone of the plane. At this point, the goal was to 
explore many $Vt$ and $At$ combinations for maximixing accuracy of the generated
labels with respect to those of the ground-truth dataset. $Vt$ and $At$ were increased by 0.01 
on each comparison. After many trials, we stopped at $At=0.34$ and $Vt=0.34$, reaching 
a maximal conformity of 74.1\% from 220 labeled lyrics. Further increases of $Vt$ and 
$At$ values significantly reduced number of comparable lyrics and computed accuracy 
started to go down.  
More details and statistics about dataset lyrics and the 
automatic annotations we obtained can be found in \cite{ismsi17}. 
%
%
%
\begin{table}
	\caption{Confusion matrix of lexicon-generated song labels}
	\centering
	\begin{tabular}{c | c c c c}   
		\toprule	
		\multicolumn{1}{c |}{True~\textbackslash~Pred} & \multicolumn{1}{c}{Happy} &
		\multicolumn{1}{c}{Angry} & \multicolumn{1}{c}{Sad} &  
		\multicolumn{1}{c}{Relaxed} \\ [0.2ex] 
		\midrule
		Happy & \textbf{68.68} & 3.63 & 2.72 & 25 \\ [0.5ex] 
		Angry & 5.9 & \textbf{80.45} & 13.63 & 0 \\ [0.5ex] 
		Sad & 7.27 & 15.9 & \textbf{74.54} & 2.27 \\ [0.5ex] 
		Relaxed & 18.18 & 0 & 9.09 & \textbf{72.72} \\	
		\bottomrule
	\end{tabular}
	\label{table:MLconfusionMatrix}
\end{table}
\par 
Table~\ref{table:MLconfusionMatrix} presents confusion matrix between the generative 
method we used and the ground-truth dataset. 
The overall accuracy of 74.1\% is probably not high enough 
for considering the method as applicable. Various reasons could be the cause of this. 
First of all, the method itself is \dq{crude}. It simply sums valence and arousal norms. 
Some terms (e.g., verbs) could be emotionally more important than other terms. 
Furthermore, meaning and emotionality of words are highly dependent on the context in which 
they are used. Unfortunately, ANEW norms of words are static numbers that do not count for 
that context. Another problem could be the way we extended ANEW. In conclusion, we do 
not consider this method as an applicable generative function for emotion labels of texts. 
Nevertheless, it might be useful if combined with other high-level heuristics devised 
from music and emotion experts.   

\chapter{Mood-Aware Music Recommenders}
\label{chapter4}
\ifpdf 
    \graphicspath{{Chapter4/Figs/}{Chapter4/Figs/PDF/}{Chapter4/Figs/}}
\else
    \graphicspath{{Chapter4/Figs/Vector/}{Chapter4/Figs/}}
\fi
%
\begin{flushright}
	\emph{\dq{Information overload is a symptom of our desire \\
			to not focus on what's important. It is a choice.}} \\
			\vskip 0.07in
		\footnotesize{-- Brian Solis, digital analyst}
\end{flushright}
\vskip 0.35in
\indent \indent
In the era of pervasive computing and \dq{everything online} culture, people have an  
essential need for automatic filtering tools to alleviate the information overload 
problem and the distraction it induces which is stressing. Search engines and recommender 
systems are such tools that have become very popular. The latter were particularly 
promoted by the increasing Internet commerce. They utilize several filtering strategies
as well as various types of data to predict items that should be the most useful for 
the users. Hybrid recommender systems try to make better user preference predictions by 
combining two or more basic filtering techniques whereas context-aware recommenders
utilize contextual data for achieving the same goal. Songs are one type of items 
that most people consume consistently on a daily basis, especially in  
the context of car driving.
\par 
This chapter presents survey results about recent research trends in hybrid and 
context-aware recommender systems. Furthermore, we describe a contextual 
mood-based recommendation system for music suggestions to car drivers that aims 
to enhance their driving experience, comfort, and cautiousness. Section~\ref{sec:IntroToRSs}  
introduces the basic recommendation techniques and describes some public
datasets that can be utilized for experimentations. Section~\ref{sec:HybridAndContext} 
provides an even more detailed discussion about hybrid and context-aware recommender 
systems as well as their applicability. Finally, in Section~\ref{sec:MoodCarRS}, design 
steps and module details of the contextual music recommender are presented. 
\section{Recommender Systems}
\label{sec:IntroToRSs}
\subsection{Introduction and Early History}
\label{sec:IntroAndHistory}
%
People have historically counted on their peers or experts for suggestions or 
recommendations about what products to buy, what places to visit, what 
songs to listen, etc. Things have changed in the last two decades with the 
proliferation of the Internet and Web access. Most people today use search engines and 
information found on websites for such suggestions. The huge and increasing amount 
of data available on the Web, combined with the massive daily-generated user content
have created the problematic phenomenon of information overload. Regardless of the quote 
at the head of this chapter, \emph{information overload} is more formally defined as 
\emph{\dq{a situation in which you receive too much information at one time and cannot think 
about it in a clear 
way.}}\footnote{\url{https://dictionary.cambridge.org/dictionary/english/information-overload}}
This problem induces stress and restricts our capability to review specifications of the 
objects for choosing the most convenient one from the many alternatives. 
To address the problem, computer science and technology have reacted appropriately
and automatic information filtering tools have been developed. Recommender Systems (RS)
represent a category of such tools invented in the 90s to provide recommendations or 
suggestions of interesting items to users \cite{Ricci2011}. Nowadays, RSs appear 
everywhere in the Web for assisting users in finding different items or services. 
They are also an important instrument for businesses, advertising products and 
increasing sales.  
\par 
In their dawn (early 90s), RSs were mostly studies of research disciplines like human-computer
interaction or information retrieval. One of the earliest that appeared was Tapestry, a manual 
Collaboration Filtering (CF) mail system \cite{Goldberg:1992:UCF:138859.138867}. 
The first computerized versions (GroupLens, Bellcore, and Ringo) of the mid-90s also implemented collaborative filtering strategy 
\cite{Resnick:1994:GOA:192844.192905,Hill:1995:REC:223904.223929}.
GroupLens was a CF engine designed to find and suggest news. 
Bellcore presented in \cite{Hill:1995:REC:223904.223929} was a video recommendation 
algorithm embedded in the Mosaic\footnote{A popular Web browser of the 1990s, discontinued in  1997.} browser interface. Ringo, on the other hand, utilized preference similarities of users to suggest
them personalized music. There were also other implementations such as NewsFeeder or InfoFinder 
for news and documents. They used Content-Based Filtering (CBF) and item features to  
generate their recommendations \cite{lang95newsweeder,krulwich1996learning}. 
Knowledge-Based Filtering (KBF) or hybrid (combining different strategies) recommenders 
followed shortly, completing the recommender system mosaic of today.  
\subsection{Basic Recommendation Techniques}
%
Technically, RSs are information filtering engines that try to predict the 
rating or the preference value that users would give to certain items and then
suggest them the best one (or top $n$). Suppose we have a set of users $U$
and a set of items (e.g., movies) $I$. If the cardinality of $I$ is high, we normally 
expect each user $u \in U$ to have watched only certain movies $i \in I$ and 
given ratings (e.g., 1 to 5) $r_{ui}$ to few of them. In this scenario, 
we have a $U \times I$ matrix that is mostly sparse, with only a subset of 
$r_{ui}$ ratings available. The job of the recommender is thus to predict 
the unknown ratings and fill the matrix by clustering together similar users
and items. This simple recommendation approach is known as collaborative 
filtering. In the case of movies, the recommender predicts what rating would each 
user give to movies he/she has not watched yet. Afterwards, the movie with the highest 
predicted rating (or top $n$ after ranking) is recommended to each user. 
The  CF recommender described above uses only three elements or data types: 
users, items, and ratings. In different types of RSs, other data types are involved or 
more knowledge is required. The various recommendation strategies that have been 
proposed differ in the data (or knowledge) and filtering algorithms they combine. 
In this context, four main RS categories are usually identified: collaborative,
content-based, knowledge-based and hybrid filtering \cite{Ricci2011}. A brief 
description of each category is presented below:
\begin{description} 
\item [Collaborative filtering] CF recommenders assume that users with similar
preferences in the past will keep having similar preferences in the future as well.
As briefly mentioned above, ratings or other forms of user feedback are used to 
identify and cluster common tastes among user groups and then provide suggestions
based on intra-user similarities \cite{Ekstrand:2011:CFR:2185827.2185828}.
In this way, users \dq{collaborate} with each other by \dq{exchanging} their
item preferences. A common problem of CF is data sparsity that happens when 
very few ratings are available and $U \times I$ matrix is extremely sparse. Another 
common problem is cold-start, a situation with new users or items that have 
no $r_{ui}$ ratings at all.  
\item [Content-based filtering] CBF is an approach that usually requires more data 
(especially about items) than CF. Here, item features are analyzed to identify 
and cluster together items with similar characteristics. CBF assumes that users 
who liked items with certain attributes in the past will prefer items with same 
attributes in the future as well. This type of recommender is highly dependent on 
(and limited by) the extracted features of recommended items. CBF suffers from 
cold-start problem, same as CF. 
\item [Knowledge-based filtering] In KBFs, knowledge about user requirements and 
item characteristics is utilized to infer the type of items that match user preferences and 
suggest accordingly \cite{Burke00knowledge-basedrecommender}. They are more 
appropriate in scenarios when little or no interaction between users and the system 
exists. In these cases, users have not provided ratings about items. As a result, CF or CBF 
cannot be used. For example, when people buy houses no ratings about their previous
house preferences are available. In this cases, users enter their item (house) requirements
in the knowledge base and the system confronts them with item characteristics 
to find the best matches. The most important weakness of KBFs is the difficulty 
to maintain and update the knowledge base.  
\item [Hybrid filtering] This is a more complex approach that mixes together two or 
more of the above techniques to alleviate their weaknesses. The most commonly 
adapted hybrid strategy is the combination of CF with CBF to combat 
data sparsity problems, increase recommendation accuracy, etc.  
\end{description}
Context-Aware Recommender Systems (CARS) represent another complex and 
advanced filtering strategy. They exploit contextual information (e.g., time, location, etc.) to 
generate adequate and useful suggestions. Sometimes these RSs are considered 
as a distinctive category and in other cases, they are described as a special type 
of hybrid recommender. Section~\ref{sec:HybridAndContext} discusses both 
context-aware and hybrid RS types in more details. 
\subsection{Experimentation Datasets} 
\label{sec:RsEvalDatasets} 
%
%
The growing popularity of recommender systems built to direct and assist users online 
poses the need for systematic and rigorous evaluation of their characteristics to assure 
user satisfaction. Accuracy, diversity, and novelty are among the most common quality 
criteria of recommenders that are assessed. One of the most viable and yet effective methods
for RS experimentation and evaluation is based on utilizing datasets with feedback data 
from real users to compare newly developed algorithms or methods with existing ones in 
the given settings. To help researchers experiments with RSs, in \cite{7325106} and \cite{cloudrs2015} we describe properties of the most popular datasets available, the repositories 
they can be retrieved from and various cloud-based recommender systems.
%
%
These datasets usually contain user feedback about amusing items like movies, books, music, etc. 
Most of the datasets were built after 2004. The oldest we found was Chicago 
Entree,\footnote{\url{http://archive.ics.uci.edu/ml/datasets/Entree+Chicago+Recommendation+Data}}  
a collection of restaurant preferences that dates back to 1996. MovieTwittings\footnote{\url{https://github.com/sidooms/MovieTweetings}} is the newest 
(2013 and on), containing movie preferences expressed in tweets. 
\par 
We observed that most of the datasets 
are made up of explicit item ratings provided by users. They are thus highly suitable for 
evaluating CF recommenders or user similarity measures. Nevertheless, there are still collections 
of book or music features that are highly appropriate for assessing content-based RSs. Few datasets
we found contain subjective user reviews in form of comments. They are thus better suited 
for sentiment analysis experiments. 
%
Regarding access and availability, most of the datasets can be freely retrieved and 
utilized for non-commercial purposes. Some of them can be obtained upon request to 
the owner/publisher. Few datasets are closed, restricted or retired from public access. 
Regarding the format of the data, in most of the cases, they come as simple texts. In some cases 
though, the data are organized in .csv, .sql or .mdb formats. 
%
Prior to using the datasets, researchers are encouraged to make data quality controls 
to ensure their research requirements are met. A more comprehensive discussion about RS 
evaluation techniques and practices can be found at \cite{pu2012evaluating}. 
\section{Hybrid and Context-Aware Recommender Systems}
\label{sec:HybridAndContext}
%
The debut of Amazon in online commerce late in the 90s boosted interest and 
academic research in RSs. During that period, hybrid and other complex RS types 
came out. The very first hybrid recommender was probably Fab, a filtering system 
that was used to suggest websites \cite{Balabanovic:1997:FCC:245108.245124}. 
Fab combined CF for finding similar users with CBF to gather websites of similar content. 
Many other hybrid RSs like \cite{Sarwar:1998:UFA:289444.289509} that followed
explored other combinations. In Sections~\ref{sec:SLRsubsec1} and \ref{sec:SLRsubsec2} we 
present the results of a systematic literature review we conducted on hybrid RSs.
Also, in Section~\ref{sec:ContextRSs} we describe context-aware recommendation strategy. 
This later was used in Section~\ref{sec:MoodCarRS} to create a music recommender in the context 
of car driving. 
%
\subsection{Hybrid Recommenders: Review Methodology}  
\label{sec:SLRsubsec1}
%
For the survey on hybrid RSs, we followed the guidelines for systematic literature reviews 
defined by Kitchenham and Charters in \cite{guidelines-2007}.  
%
The following research questions were addressed: 
\begin{description}
	\item[RQ1] \emph{What studies addressing hybrid 
	recommender systems are the most relevant?}
	\item[RQ2] \emph{What problems and challenges are faced by the researchers 
	in this field?}
	\item[RQ3] \emph{What technique combinations are explored and implemented in hybrid RSs?}
	\item[RQ4] \emph{What hybridization classes are used, based on the taxonomy of Burke?}	 
	\item[RQ5] \emph{In what domains are hybrid recommenders applied?}
	\item[RQ6] \emph{What methodologies are used for the evaluation and which metrics they utilize?}	
	\item[RQ7] \emph{Which directions are most promising for future research?}
\end{description}
%
In RQ1 we observe the relevant studies and try to see any pattern with respect to publication
type (e.g., journal vs. conference), publication date, etc. RQ2 identifies the most common 
problems that hybrid RSs address. RQ3 examines popular technique combinations and 
associated problems each of them tries to solve. 
%
In RQ4 we examine the possible ways in which different techniques can be combined 
with respect to the systematic taxonomy proposed by Burke 
\cite{Burke:2002:HRS:586321.586352}. The author examined a plethora of 
existing hybrid RSs and created a complete taxonomy of seven hybrid 
recommender classes we briefly describe below:
\begin{description}
\item[Weighted] This type of hybrid RSs are very simple and intuitive. They  
calculate utility scores of recommended items by aggregating output scores of different 
recommendation strategies utilizing weighted linear functions. 
\item[Feature combination] In this class of hybrid RSs, the output of one recommender
is considered as additional feature data that enters as input to the other (main) recommender.
This latter generates the final item suggestions. 
\item[Cascade] These hybrid RSs that combine two or more strategies makes up 
another hybridization class. The first recommender in the cascade generates a coarse 
ranking list and the second strategy refines that list. Cascades are order-sensitive, which 
means that a CF-CBF cascade is different from a CBF-CF one. It is certainly possible to have 
cascades of three or even more basic modules chained together. 
\item[Switching] These recommenders switch between the composing 
techniques in accordance with certain criteria. As a simple example, we can consider a CF-CBF 
that mostly uses CF and occasionally switches to CBF (using item features) 
when CF does not have enough data about users. 
\item[Feature augmentation] This hybrid RS type uses one technique to produce item 
predictions or listings that are further processed by the second technique. As an example, we 
can consider an association rule engine generating item similarities that are fed as augmented 
features in a second recommender. 
\item[Meta-level] These hybrid RSs utilize the entire model produced by a first 
technique as input for the second one. A content-based recommender, for example, may be used 
to create item representation models that may be entered to a second CF for better item 
similarity matching.   
\item[Mixed] These hybrids use different RSs in parallel and select the best predictions 
of each of them to create the final recommendation list. They represent the simplest form of 
hybridization and are suitable in cases when it is possible to use a high number (e.g., more than 
three) of RSs independently. 
\end{description}
Experimentation and application domains are examined in RQ5. 
RQ6 addresses metrics and methodologies that are used for evaluating hybrid RSs and 
finally, RQ7 summarizes promising research directions.  
%
As primary sources for scientific studies, we picked five scientific digital libraries
shown in Appendix~\ref{sec:Appendix1}, Table~\ref{table:DigitalLibraries}. 
Meanwhile, a set of keywords including basic terms like \dq{Hybrid}, 
\dq{Recommender} and \dq{Systems} was defined. Later, we added synonyms 
and organized terms to form the search string shown in Listing~\ref{listing:HRS}. 
The whole string was applied in the search engines of the digital libraries and 
9673 preliminary research papers were retrieved. The set of inclusion / exclusion 
criteria listed in Table~\ref{table:InclusionExclusion} of Appendix~\ref{sec:Appendix1} 
were defined for an objective selection of the final papers from the preliminary ones.  
%
%
\vskip 0.2in
\begin{lstlisting}[caption={The search string for finding studies in digital libraries}, 
label={listing:HRS}]
("Hybrid" OR "Hybridization" OR "Mixed") AND 
("Recommender" OR "Recommendation") AND ("System" OR 
"Software" OR "Technique" OR "Technology" OR "Engine" 
 OR "Approach")
\end{lstlisting}
%
Utilizing inclusion/exclusion criteria and a coarse inspection based on abstract and 
metadata, we reached to a set of 240 most relevant papers. Next, we carried out an even 
more detailed analysis, examining besides abstract, content parts of each paper as well. 
In the end, we reached to the final set of 76 included papers that are listed in 
Table~\ref{table:FinalIncluded} of Appendix~\ref{sec:Appendix1}. 
%
We also performed a quality assessment of the final included papers. For a systematic 
evaluation, we defined six quality questions listed in Table~\ref{table:PaperQuality} of 
Appendix~\ref{sec:Appendix1}. Each of them was given a certain weight for highlighting the 
importance of that question in the overall paper assessment. Quality evaluation process consisted
in responding with \dq{yes}, \dq{partly} or \dq{no} to each of the six questions. Finally, the overall quality score of each study was computed using the following formula:
\begin{equation}
score = \sum_{i=1}^{6} w_{i} * v_{i} / 6
\end{equation}
\vskip 0.05in
\noindent $w_{i}$ is the weight of question $i$ \textit{(0.5, 1, 1.5)} 		\\
$v_{i}~$ is the vote for question $i$ \textit{(0, 0.5, 1)} 			\\	\\	
%
As part of data extraction phase, both paper attributes (e.g., title, authors, year, etc.) and 
content data were collected. The data extraction form we used is shown in 
in Appendix~\ref{sec:Appendix1}, Table~\ref{table:DataExtraction}. All extracted 
information was stored in Nvivo,\footnote{\url{http://www.qsrinternational.com/products.aspx}} 
a data analysis software that automates identification and labeling of initial text 
segments from the selected studies.
%
For the thematic synthesis, Cruzes and Dyba methodology was followed \cite{6092576}. 
To organize and aggregate extracted information, that methodology utilized the concept of 
codes which are labeled segments of text. Codes were later merged into themes for grouping 
the selected papers. Each research question was mapped with the corresponding
themes and extracted data were summarized in categories that were reported as results of 
the survey. More details and statistics about each step we followed can be found in 
\cite{pub2659069}.
\subsection{Hybrid Recommenders: Review Results} 
\label{sec:SLRsubsec2}
%
We discuss in this section the obtained results of the systematic literature review on 
hybrid recommenders and answer each research question listed in the previous section.  
Regarding quality of the selected studies, we observed that journal papers tend 
to have a slightly higher quality score. Regarding publication year of studies, 
more than 76\% were published after 2010. This is an indication of a high and 
increasing interest in RS research, same as reported in similar surveys like  
\cite{Park:2012:LRC:2181339.2181690} or \cite{Bobadilla:2013:RSS:2483330.2483573}.
Regarding research problems (RQ2), the most frequently addressed was cold-start 
which mainly affects CF recommenders. It was followed by data sparsity that can
affect any kind of RS. Both problems are attacked with a variety of data mining or
matrix manipulation technique combinations or aggregation of extra user data and 
item features. Increasing accuracy or scalability and providing higher diversity in 
recommended items were other typical addressed problems.  
\par 
A high variety of basic data mining or machine learning techniques and 
algorithms are serving as centric parts of hybrid recommenders. K-nearest neighbors 
is the most popular, especially as part of collaborative filtering implementations. Clustering 
algorithms with \emph{K-means} the most popular are also highly utilized, 
especially in the preliminary phases when similar items or users are identified.  
In most of the cases, two recommendation strategies are mixed together, 
with CF-CBF being the predominant combination. The goal is to alleviate problems
like data sparsity and cold-start from which both CF and CBF suffer a lot. 
With respect to hybridization classes (RQ4), \emph{weighted} hybrids are the most popular,
followed by \emph{feature combination} and \emph{cascade} RSs. In many cases CF and 
CBF are put together through a weighting function. \emph{Mixed} hybrids are the least 
studied and implemented.
\par 
Regarding RQ5 and application domains, movies recommenders are still the most common. 
This is partly because of the many experimentation movie datasets (e.g., those described 
in Section~\ref{sec:RsEvalDatasets}) that are publicly available. Moreover, Netflix \$1M 
prize did certainly promote research and implementation of movie RSs in 
some way. Education and especially e-learning represent another common and interesting 
application domain. MOOCs (Massive Open Online Course) are gaining a lot of popularity. 
Also, education materials on the Web have been increasing dramatically in the last decade. 
Evaluation of RSs is not an easy task \cite{Herlocker:2004:ECF:963770.963772}. 
According to our findings, most of the studies perform evaluations by comparing 
their hybrid RSs with similar baseline recommenders. Datasets and various metrics
such as \emph{mean average error} and \emph{root mean square error}, or information 
retrieval metrics like \emph{precision}, \emph{recall} and \emph{F1} score are used for 
this process. Accuracy is still the most commonly assessed characteristic, followed by 
diversity of the recommendations.   
A highly desired characteristic that is reported as future direction (RQ7) is to have 
hybrid RSs that suggest items of different and changing domains (cross-domain 
recommenders). Another possibility is the increase of data utilization by parallelizing 
algorithms with \emph{MapReduce} model as suggested in \cite{barragans2010hybrid}.
Other common future works that are reported include increasing personalization of 
recommendations, reducing their computational cost, etc.  
\subsection{Context-Aware Recommenders}
\label{sec:ContextRSs}
Context-aware recommenders represent a family of RSs that are based on the notion 
of context. Recommendations and decision making, in general, are inherently 
related to contextual data. For example, current activity as the context has 
a significant influence on one's musical choices. Nobody expects the same type 
of music recommendations in different situations like working out, 
car driving, studying or going to sleep. Similarly, a restaurant menu
recommended for a quick lonely lunch should differ from that 
of a business dinner with colleagues. 
Context as a concept has been defined in various forms, especially characterized
by location and nearby people or objects. In \cite{Dey:2001:UUC:593570.593572} 
we find a complete and formal definition from Anid K. Dey:  
%
\emph{\dq{Context is any information that can be used to characterize the
situation of an entity. An entity is a person, place, or object 
considered relevant to the interaction between a user and an
application, including users and applications themselves.}}
%
%
The information mentioned in the above definition is usually about (or comes from) 
various contextual factors like time, location, actual activity or occasion (above examples). 
They are aggregated with user and item information by means of the recommendation 
function. More formally, a CARS can be modeled as a function
$f: U \times I \times C  \rightarrow \mathbb{R}$, where the real-value item ratings 
are generated by aggregating context factors $C$ with users $U$ and items $I$ 
\cite{Adomavicius2011}.
%
From this formalization, we can distinguish three components: the input data 
($U$, $I$, and $C$), the recommendation algorithm or function $f$ and the 
output (recommendation list). Contextual factors as part of input data are 
sometimes not easily obtained and aggregated. This is especially true when they 
continuously change in time. 
\par 
In fact, a highly desired characteristic from RSs is 
exactly the ability to adapt to quickly shifting user interests as a result of context change. 
The pervasive utilization of mobile devices has simplified the ability to obtain 
certain factors like time of day or location. However, other context objects such 
as people, occasion or goal remain difficult to interpret. 
%
Contextual data can be applied to the recommendation process in different phases
and components. We can thus identify three types of CARS:
\begin{description}
\item[Contextual pre-filtering] This is the term used to describe CARS in which 
contextual factors are used in data input phase. They guide input data selection 
or processing. This approach is usually simple and can be applied to other 
recommendation strategies to improve their performance. 
\item[Contextual modeling] In this case, context is applied in the recommendation 
function as part of the rating prediction model. As a result, the function becomes
highly complex. For this reason, this method cannot be applied to existing 
recommendation strategies.  
\item[Contextual post-filtering] In this case, contextual factors are applied to
output data (recommendation list). They are ignored in input data selection 
and ratings are initially predicted via traditional approaches. Afterwards, 
the initial set of ratings is corrected using the contextual factors. The correction 
may be performed through a filter or a rearrangement.   
\end{description} 
Performance of these contextualization approaches is highly dependent on the 
application. Usually prefiltering is simpler than the other two, and thus more suitable
for non-critical performance requirements. 
\section{Mood-based On-Car Music Recommendations}
\label{sec:MoodCarRS} 
%
Since the first in-car radios were introduced back in the 
1930s,\footnote{\url{https://www.caranddriver.com/features/the-history-of-car-radios}}
music listening has been the favorite activity for most people while driving their cars. 
According to \cite{dibben2007exploratory}, roughly 70\% of car drivers 
do it habitually.  
%
Various psychological studies report relations between background music 
and concentration, comfort or driving performance, providing evidence
that music behaves as a stimulator that can have both positive and negative 
effects on mood and driving \cite{Nesbit, valenza2014revealing}.  
They opened up several research possibilities that attempt to create  
relaxing car conditions for optimal car driving by means of proper music 
recommendations. 
%
In this section, we present the design of a mood-based recommender in the 
context of car driving. It tunes song recommendations using different sources 
of contextual data like driver's heart rate dynamics, his/her musical preferences, 
driving style obtained from telemetry data as well as location and time.
%
The ultimate goal is to enhance driving comfort and safety by means of 
a proper music induction.    
\subsection{System Prerequisites}
%
Important correlations between driver's mood and his/her driving patterns 
are highlighted in different studies. In \cite{garrity} for example, authors 
reveal influence that negative moods like depression or anger have in driving
cautiousness. Furthermore, in \cite{Nesbit}, they show correlations 
between an angry emotional state and aggressive driving style. 
Regulative efforts by music stimulation such as those of \cite{VanDerZwaag} 
have resulted somehow effective for a gradual shift in the mood of the driver. 
Several prerequisites are essential for building a car-based music mood 
recommender.   
%
Same as in Section~\ref{sec:TagDatasets}, we had to pick up a model for  
representing emotion categories of songs. The only external feedback about 
song emotions that we could access was that of social tags. For this reason,  
we once again utilized the tag folksonomy of Table~\ref{table:ClassModel}
and the emotion model of Figure~\ref{fig:Russell4Q}. They are both practically 
convenient and highly compatible with the psychological model of Russell  
(Section~\ref{sec:TagFolksonomy}).   
\par 
Impact of music in one's mood is assessed by psychologists using a standard 
method called Musical Mood Induction Process (MMIP) which consists in 
replaying mood-eliciting tracks to participants. Authors in \cite{vastfjall2002emotion} 
describe various MMIP methods such as behavioral measures, self-reports
or physiological measures. For this project, we trusted on physiological data
such as heart rate dynamics. Similarly, in \cite{Peter2005}, authors recognize emotions 
based on skin conductivity, skin temperature, and heart rate.
Other contextual factors are driver's mood state and his/her current driving style.  
One approach is to include all contextual factors in the recommendation function 
(contextual modeling), obtaining a multidimensional model from the Cartesian  
product of the most relevant attributes \cite{Adomavicius:2005:ICI:1055709.1055714}. 
In the case of our project the dimensions could be: 
%
\begin{center}
\textit{\textbf{Users} \,\, $\subseteq$ UserName $\times$ Age $\times$ Gender $\times$ Profession} \\ 
\textit{\textbf{Items} \,\, $\subseteq$ SongTitle $\times$ Artist $\times$ Genre $\times$ MoodLabel} \\ 
\textit{\textbf{Contx} \, $\subseteq$ HeartRate $\times$ \,Place $\times$ \,Time $\times$ \,DriveStyle}  
\end{center}
Contextual parameters can be retrieved in different ways. Location and time, for example,
are easily retrieved from the GPS and car dashboard respectively. For the heart rate, there 
are different cheap sensors like \emph{Empatica} that can be used. Driving style can be
inferred from car telemetric data by means of OBD-II technology which provides   
diagnostic information about the car. 
Nowadays, a plethora of OBD-II adapters are made available in the market. 
They include APIs to mobile applications and are easily integrated with 
the quickly growing on-car infotainment dashboards which are connected to the 
Internet \cite{viereckl}. 
%
%
\subsection{System Architecture}
\begin{figure}[h!]
	\centering
	\includegraphics[width=0.87\textwidth]{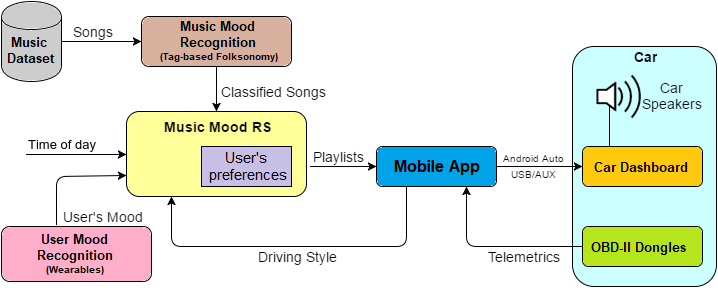}
	\caption{Holistic view of the system}  
	\label{fig:systemscheme}  
\end{figure}

%
Figure~\ref{fig:systemscheme} shows the entire parts of the system connected together.    
The central module is the music recommender which is interconnected with all other 
parts. Its role is to generate the appropriate playlist of tracks that will be suggested 
to the driver. The recommender takes in various types of data such as contextual 
factors (e.g., time or location), driving style patterns, driver's mood state, and 
emotionally labeled songs. 
%
User mood recognition module obtains driver's mood 
data (heart rate dynamics) from wearable sensors (Empatica).   
To recognize mood state of the driver, the system presented in \cite{valenza2014revealing} 
is implemented. That system is based on cardiovascular dynamics (heart rate variability)
observations on short-time emotional stimuli. Authors have used as emotional model
the Circumplex Model of Affect which is very similar to the one we adopted. 
Two levels (high and low) of \emph{arousal} and \emph{valence} that correspond to
to one of the four emotional categories are recognized and transmitted to the 
main module (the recommender). 
\par 
OBD-II module generates car telemetry data from which driving style patterns are 
extracted. 
Driver's aggressive patterns are identified by computing the jerk
(the first-order derivative of acceleration) 
and considering car acceleration profile. To discriminate between calm and 
aggressive driving styles, a heuristic threshold of jerk is utilized. We considered 
as jerk threshold the one provided by \cite{Murphey09} that was derived as average
driving jerk value on a number of driving cycles of typical scenarios. As a result 
we obtain \emph{aggressive} style when the actual jerk is greater than the 
threshold. Driving style goes as a flag to the mobile application and is used inside the 
recommender to affirm or dissent the mood state of the wearable sensor. 
Usually, and aggressive driving is associated with high levels of arousal 
and/or an angry emotional state. 
%
In the top left part of the scheme, we also see the music mood recognition 
module. It is responsible for the emotional annotation of songs retrieved from 
public datasets such as Million Song 
Dataset.\footnote{\url{https://labrosa.ee.columbia.edu/millionsong/}}
Social tags collected from \emph{Last.fm} as well as the folksonomy of 
Table~\ref{table:ClassModel} are used for this process. Each song receives a label 
that may be \emph{happy}, \emph{angry}, \emph{tender} or \emph{sad} based on 
the planar model shown in Figure~\ref{fig:Russell4Q}. 
%
More details about the different modules of the system can be found in 
\cite{pub2650985}.        
\subsection{Recommender and Mobile Application}
\begin{figure}[H]   
	\centering
	\includegraphics[width=0.74\textwidth]{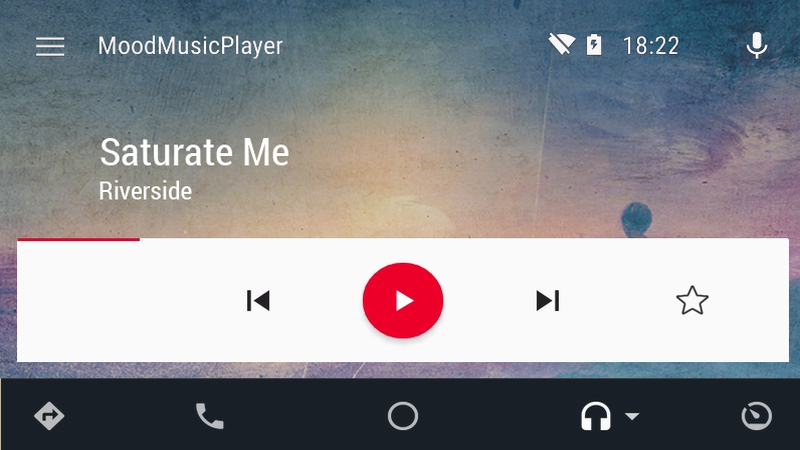}
	\caption{Interface of mobile application}
	\label{fig:mockup2}
\end{figure}

%
After obtaining all contextual data, the recommender has to generate the 
appropriate playlist for the driver. Besides the modules (and data) described above, 
there is also another feature: time of day. We assume that there is no need for extra arousal 
during the day and consider \dq{tender} (relaxing) as the default target 
mood category. Contrary, during a night drive it is better to avoid sleepy state 
of the driver by recommending \emph{happy} (more aroused) music. 
The goal of the system is to maintain a relaxed state of the driver with 
song recommendations of his/her taste. The recommender takes in also the 
driving style which can be \emph{aggressive} or \emph{normal}. 
When the user is already in relaxed mood and driving style is normal, 
priority is given to past musical preferences. Otherwise, relaxing music 
is displayed. The recommendation list goes to the mobile application. 
\begin{figure}[ht]
	\centering
	\includegraphics[width=0.74\textwidth]{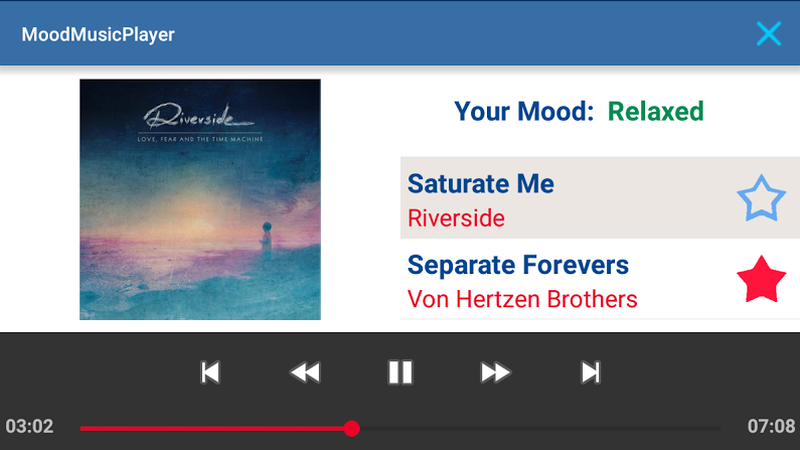}
	\caption{Iterface of song recommendations} 
	\label{fig:mockup1}
\end{figure} 
%
This is an Android mobile application that provides a simple user interface and 
enables media playback. Since it is supposed to be used in the car environment,
it was programmed to be compatible with the Android Auto platform for 
the in-car streaming. The phone needs to be connected to the car dashboard 
via AUX/USB cable for music playback through car speakers. The user interface of 
the application is designed not to cause distractions 
to the driver. There is also a button that enables the user to express appreciation for the 
recommended songs of the list. Application interface is presented in 
Figure~\ref{fig:mockup2}. The user is free to select which song to play from the 
recommended list (Figure~\ref{fig:mockup1}). If no selection is performed 
the top-ranked song starts automatically.  

\chapter{Distributed Word Representations}
\label{chapter5}
\ifpdf
    \graphicspath{{Chapter5/Figs/}{Chapter5/Figs/PDF/}{Chapter5/Figs/}}
\else
    \graphicspath{{Chapter5/Figs/Vector/}{Chapter5/Figs/}}
\fi
%
\begin{flushright}
	\emph{\large \dq{The beginning of wisdom is the definition of terms.}} \\
	-- Socrates
\end{flushright}   
\vskip 0.35in
\indent \indent
Distributed word feature representations known as word embeddings are generated 
training shallow neural architectures with huge text bundles for learning word 
relation predictions. Several neural architectures have been designed for that purpose.
Some of them are quite efficient and produce vectors that are able to retain semantic
and syntactic similarities between words, making them applicable in tasks like 
topic modeling or sentiment analysis. Performance of those 
feature vectors depends on various factors like training method and parameters,
size and vocabulary of source texts, the thematic relevance of application domain
with that of source texts, etc. Experimental observations reveal interesting relations 
between the influencing factors and performance of word embeddings on each task. 
For example, sentiment analysis of song lyrics and movie reviews seems to be more 
sensitive to corpus size than to other factors like the thematic relevance of texts.  
\par 
This chapter presents obtained results from various comparative experiments on sentiment 
analysis tasks with word embeddings as dense text feature representations. 
Section~\ref{sec:WordRepModels} introduces local (traditional) and distributed text representation 
models, highlighting their advantages and disadvantages. Details of the most popular word 
embedding generation neural architectures are described in Section~\ref{sec:PopularEmbMethods}. 
Finally, Section~\ref{sec:QualWordEmb} presents and further discusses the  
empirical results that were reached.
\section{Word Representation Models}   
\label{sec:WordRepModels}
%
Distributed word representations generated from neural language models are 
replacing Bag-Of-Words (BOW) representation in various text analysis applications. 
BOW has been traditionally recognized for its simplicity
and efficiency. It was first proposed in \cite{harris1954distributional} where 
the authors discuss the possibility of describing languages via distributional 
structures (e.g., in terms of co-occurring parts). Sometimes Set-Of-Words (SOW)
representation is used, where each word of vocabulary $V$ is counted only 
once and its presence or absence is encoded and used as a feature. 
%
Both BOW and SOW are considered as discrete representations where each word is 
encoded with a binary or frequency number. Other vectorization and scoring methods 
like \emph{term frequency-inverse document 
frequency} (\emph{tf-idf}) are also popular. In fact, BOW combined with 
\emph{tf-idf} have been successfully applied in many text classification studies, especially 
in combination with support vector machine used as classifier \cite{joachims1998text}.
SOW is also an example of localist representations, in the sense that it allocates a unit
of memorization for every word in all documents that word appears in.
\par 
The main problem with BOW and SOW is their poor scalability with respect
to vocabulary size $V$. Every word is encoded in a sparse vector of a 
$V$-dimensional space. As vocabulary $V$ can grow to hundred thousands of words, 
data sparsity becomes a serious issue. Another problem is the very high 
feature dimensionality (again with respect to large $V$) that results, leading to 
overfitting (the infamous \emph{curse of dimensionality} problem).    
%
Furthermore, BOW representation is not able to conserve order of text words. For 
example, the phrase \dq{excellent and not expensive service} has same 
representation with the phrase \dq{expensive and not excellent service}.
The former expresses a \emph{positive} opinion whereas the latter a \emph{negative} one. 
This problem causes performance degradation on sentiment polarity 
analysis tasks. 
%
From the linguistic point of view, BOW is unable to retain semantic relations
of words. For example, words \dq{boy} and \dq{girl} are semantically related
(gender, human beings) wheres their corresponding vectors are orthogonal.  
\par 
Word embeddings trained from neural networks on large text corpora were 
invented to solve the above problems. They are examples of continuous space
representations where every word is encoded to a $D$-dimensional (typically 
100 -- 300) vector of real (continuous) values. They are also called distributed 
in the sense that every word vector is stored only once and shared in all documents
containing that word. The main difference with BOW is the fact that $D$ is fixed
and independent of $V$. As a result, word embeddings offer dense data 
representations of reduced dimensionality even when vocabulary size is very big. 
%
Moreover, studies like \cite{DBLP:journals/corr/abs-1301-3781} 
or \cite{conf/emnlp/PenningtonSM14} that present Skip-Gram and Glove
methods, also confirm that word embeddings trained from large text 
corpora are able to preserve syntactic and semantic word relations. They test this 
property by means of word analogy tasks and report very good results. 
%
It is, however, important to note that word feature quality depends on training data, 
number of word samples and size of vectors. It takes a lot of computation time to obtain 
high-quality representations. The following sections describe in details some of the most 
popular training methods that are available today.   
%
%
\section{Popular Word Vector Generation Methods}
\label{sec:PopularEmbMethods}
\subsection{Continuous Bag of Words}
\begin{figure}
	\centering
	\includegraphics[width=0.75\textwidth]{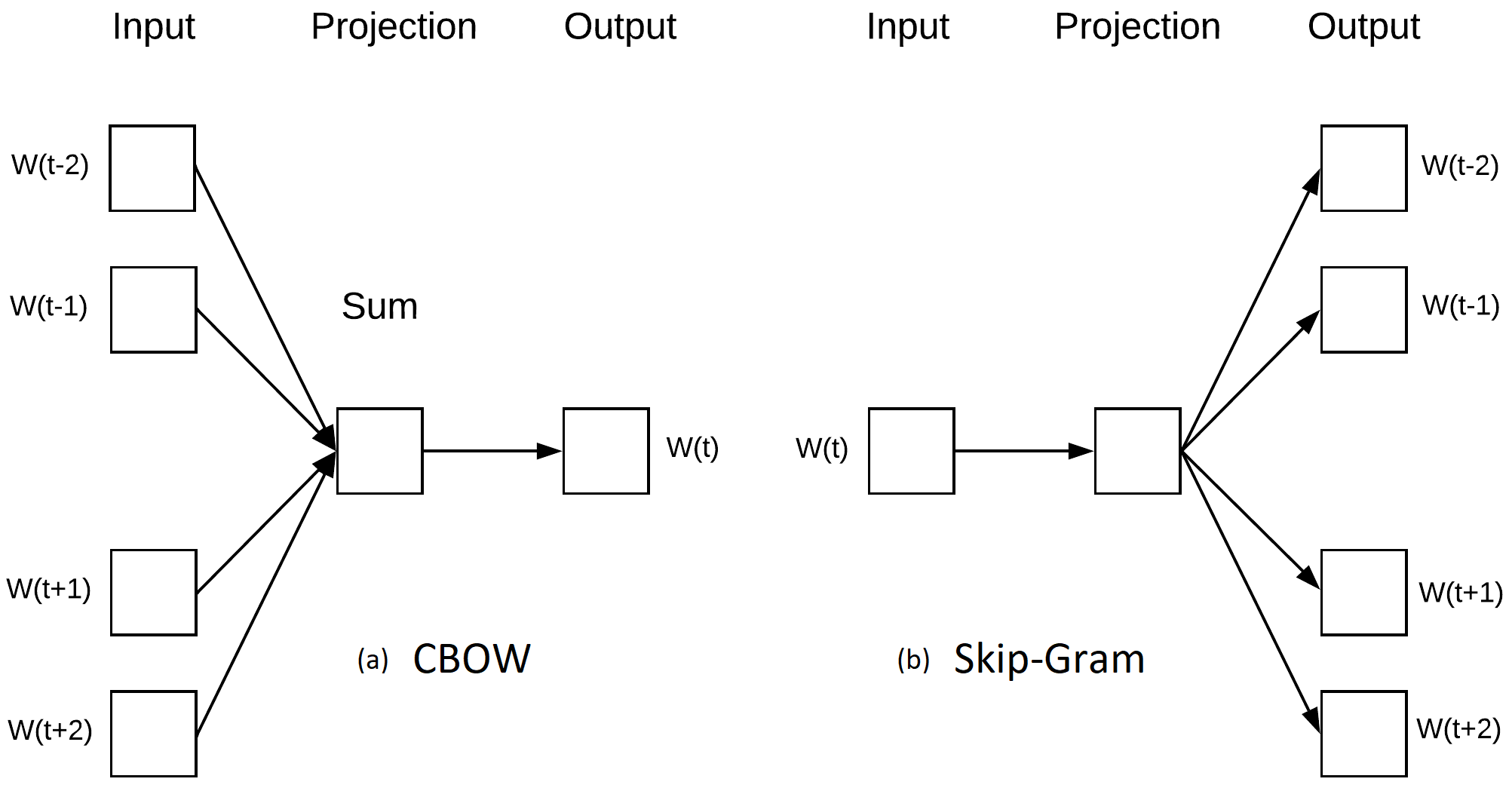}
	\caption{CBOW and Skip-Gram neural architectures}
	\label{fig:cbowSkipGram}
\end{figure}

CBOW architecture proposed in \cite{DBLP:journals/corr/abs-1301-3781}
is a simplification upon the feed-forward neural model of \cite{Bengio:2003:NPL:944919.944966}.
The hidden layer that introduces non-linearity is removed and the input window of 
$Q$ words is projected into a $P$-sized projection layer. $Q$ future words are used 
as well and the objective is to correctly predict the middle word. They use a log-linear 
classifier with a binary tree representation for the vocabulary. This way the number 
of units in the output layer drops from $V$ to $log_{2}(V)$. As a result, the total 
training complexity of CBOW becomes: 
\begin{equation}
C = E \times T \times (Q \times P ~ + ~ P \times log_{2}(V))
\label{equation:cbowComplex}
\end{equation}
where $E$ is the number of training epochs and $T$ is the total number of tokens
appearing in the training text bundle. Throughout this thesis, the term \dq{token}
is used to indicate raw words that do usually repeat themselves within a text document 
or text bundle. Unique words that form the vocabulary of that text structure are called 
\dq{vocabulary words} or simply \dq{words}.  
The architecture of CBOW is schematically presented in Figure~\ref{fig:cbowSkipGram} 
(a) and shows the use of distributed context words to predict the current word.  
\subsection{Skip-Gram}
Skip-Gram architecture shown in Figure~\ref{fig:cbowSkipGram} (b) is similar to 
that of CBOW. However, it starts from the current word and predicts context words 
appearing near it \cite{DBLP:journals/corr/abs-1301-3781}. A log-liner classifier with
projection layer takes the current word as input and predicts nearby words that appear 
before and after (inside a window) the current word. The training complexity of 
this architecture is 
\begin{equation}
C = E \times T \times (W \times (P ~ + ~ P \times log_{2}(V)))
\label{equation:skipGramComplex}
\end{equation}
where W is the word window size. Authors report that enlarging the size of the window 
enhances the quality of generated vectors but also increases computation cost as suggested
from Equation~\ref{equation:skipGramComplex}. 
%
For benchmarking vector quality of different architectures, 
they create the analogical reasoning task that consists of analogies 
like \emph{Italy} : \emph{Rome} ~~==~~ \emph{France} : \_\_\_. These tasks 
are solved by finding the vector of a word $x$ (in this case $x$ is \emph{Paris}) such that 
$vec(x)$ is closest in cosine distance to $vec(Italy) - vec(Rome) + vec(France)$.
Besides semantic analogies, the task dataset also contains syntactic analogies as well
(e.g., \emph{walk} : \emph{walking} ~~==~~ \emph{run} : \emph{running}) and is available 
online.\footnote{\url{https://code.google.com/archive/p/word2vec/source/default/source}}
Authors generated a collection of word vectors trained on a huge Google News corpus of 100 
billion tokens and released them for public 
use.\footnote{\url{https://code.google.com/archive/p/word2vec/}} 
\subsection{Glove}
Glove (GLObal VEctors) described in \cite{conf/emnlp/PenningtonSM14} is a log-bilinear
regression model that generates word vectors based on global co-occurrences of words. 
Authors argue that probability ratios of word co-occurrences can be used to unveil
aspects of word meanings. For example, they observe that $P(solid | ice) / P(solid | steam)$
is about 8.9 whereas $P(gas | ice) / P(gas | steam)$ is only 0.085. This is something we 
logically expect, since \dq{solid} is semantically more related with \dq{ice} than it is 
with \dq{steam}. Similarly, \dq{gas} is closer to \dq{steam} than it is to
\dq{ice}. Based on these premises, they build a weighted least square regression model that
learns word vectors by means of word-word co-occurrence statistics. The
calculations on real text corpora show a complexity of $O(|T|^{0.8})$ for the model, 
where $T$ is again the total number of tokens appearing in the train texts.
%
To evaluate the quality of word vectors, authors create various big datasets of varying 
contents.\footnote{\url{https://nlp.stanford.edu/projects/glove/}}
They also utilize the word analogy task described above and report that Glove performs 
slightly better than other baselines such as CBOW or Skip-Gram, especially with 
larger training corpora. Moreover, its scalability provides substantial improvements from 
further increase of training corpus size. 
\subsection{Paragraph Vectors}
%
A common limitation of all above word vector generation methods 
in text mining applications is the fact that they produce fixed-length vectors for
words but not for variable-length texts like sentences or paragraphs. 
Many text datasets contain documents that have different lengths. As a result,  
they must be preliminarily clipped and/or padded to a fixed length.
To overcome this limitation, authors in \cite{Le:2014:DRS:3044805.3045025}
propose \emph{Paragraph Vector}, a method for learning fixed-length
continuously distributed representations of variable-length text excerpts such 
as sentences, paragraphs or entire documents. Same as in CBOW, paragraph vectors
contribute to the prediction of the next word (following few words in a fixed window)
using the context words sampled from the paragraph. Paragraph and word vectors 
inside each paragraph are concatenated to predict the next word. As soon as the word and 
paragraph vectors are obtained from training texts (training phase), vector prediction
for unseen paragraphs is performed (inference phase). Authors report significant 
improvements when comparing with BOW representation on supervised 
sentiment analysis of short texts like sentences. For longer documents like movie 
reviews, accuracy gains of paragraph vectors are lower. 
\section{Performance of Word Embeddings on Sentiment Analysis Tasks}  
\label{sec:QualWordEmb} 
The purpose of the conducted experiments with word embeddings was to observe
the role that various factors like training method, size of training corpus and thematic
relevance between training texts and analyzed documents might have on sentiment 
analysis prediction accuracy. To this end, three research questions were posed:
\begin{description}
	%
	\item[RQ1] \emph{Is there any observable difference in performance between 
	Skip-Gram and Glove word embeddings?}
	\item[RQ2] \emph{What is the role of training corpus size in the performance of the 
		generated word embeddings on sentiment analysis tasks?}
	\item[RQ3] \emph{How does thematic relevance of training texts influence behavior 
	of word embeddings on sentiment analysis of different text types?}
\end{description}
The following sections present the experimental resulst and the concluding remarks that 
answer these questions. 
\subsection{Word embedding models and corpora}
%
%
In \cite{pub2668229} we contrasted word embedding models trained with text corpora 
of various attributes. Some of those models are publicly available and two of them 
were trained by us (Text8Corpus and MoodyCorpus). The full list with some basic 
attributes is presented in Table~\ref{table:embedCorp}. 
%
\begin{table}[ht] 
	\caption{List of word embedding corpora}  
	\small 
	\centering      
	\setlength\tabcolsep{4.7pt}  
	\begin{tabular}
		{l c c c c c}  
		\topline
		\headcol \textbf{Corpus Name} & \textbf{Training} & \textbf{Dim} & \textbf{Size} & \textbf{Voc} & \textbf{URL}  	\\ [0.5ex] 
		\midline   
		Wiki Gigaword 300 & Glove & 300 & 6B & 400K & \href{http://nlp.stanford.edu/projects/glove/}{link}			  \\
		\rowcol	Wiki Gigaword 200 & Glove & 200 & 6B & 400K & \href{http://nlp.stanford.edu/projects/glove/}{link}			  \\
		Wiki Gigaword 100 & Glove & 100 & 6B & 400K & \href{http://nlp.stanford.edu/projects/glove/}{link}			  \\
		\rowcol	Wiki Gigaword 50 & Glove & 50 & 6B & 400K & \href{http://nlp.stanford.edu/projects/glove/}{link}			  \\
		Wiki Dependency & Skip-Gram & 300 & 1B & 174K &  
		\href{https://levyomer.wordpress.com/2014/04/25/dependency-based-word-embeddings/}{link} \\
		\rowcol Google News & Skip-Gram & 300 & 100B & 3M & \href{https://code.google.com/archive/p/word2vec/}{link}			  \\ 
		Common Crawl 840 & Glove & 300 & 840B & 2.2M & \href{http://nlp.stanford.edu/projects/glove/}{link}			  \\
		\rowcol Common Crawl 42 & Glove & 300 & 42B & 1.9M & \href{http://nlp.stanford.edu/projects/glove/}{link}			  \\
		Twitter Tweets 200 & Glove & 200 & 27B & 1.2M & \href{http://nlp.stanford.edu/projects/glove/}{link}			  \\
		\rowcol Twitter Tweets 100 & Glove & 100 & 27B & 1.2M & \href{http://nlp.stanford.edu/projects/glove/}{link}			  \\
		Twitter Tweets 50 & Glove & 50 & 27B & 1.2M & \href{http://nlp.stanford.edu/projects/glove/}{link}			  \\
		\rowcol Twitter Tweets 25 & Glove & 25 & 27B & 1.2M & \href{http://nlp.stanford.edu/projects/glove/}{link}			  \\
		Text8Corpus & Skip-Gram & 200 & 17M & 25K & \href{https://cs.fit.edu/\%7Emmahoney/compression/textdata.html}{link}			  \\
		\rowcol MoodyCorpus & Skip-Gram & 200 & 90M & 43K & \href{http://softeng.polito.it/erion/}{link}			  \\
		\bottomline
	\end{tabular} 
\label{table:embedCorp}
\end{table}
%
\noindent Wikipedia Gigaword combines Wikipedia 2014 dump with 
Gigaword 5,\footnote{\url{https://catalog.ldc.upenn.edu/LDC2011T07}}
obtaining a total of six billion tokens. Authors of \cite{conf/emnlp/PenningtonSM14}
created it to evaluate Glove performance. There are 400,000 unique words 
inside, trained from context windows of ten words to the left and ten other 
words to the right. The four versions that were derived differ in vector sizes
only having vectors of 50, 100, 200 and 300 dimensions. 
%
Wikipedia Dependency is a bundle of one billion tokens crawled from Wikipedia. 
It was trained using a slightly modified version of Skip-Gram presented in 
\cite{DBLP:conf/acl/LevyG14}. Authors have experimented with several syntactic
contexts of words. They filtered out every word with a frequency lower than one hundred,
reaching to a vocabulary size of 175,000 words and 900,000 syntactic contexts.
Authors report that the additional contexts help in producing word vectors that exhibit better 
word analogies. 
%
Google News is one of the largest and richest text sets that are available. It was trained on 
100 billion tokens and contains three million distinct words and phrases. \cite{DBLP:journals/corr/abs-1301-3781}. 
Skip-Gram was used with context windows of five words, generating vectors of 300 dimensions.
Reduced versions of the corpus were also used in \cite{DBLP:journals/corr/MikolovSCCD13} 
for validating the efficiency and training complexity of CBOW and Skip-Gram methods.
\par
Common Crawl 840 is even bigger than Google News. It was trained using Glove on 840 billion 
tokens, resulting in 2.2 million unique word vectors. It comprises texts of 
Common Crawl,\footnote{\url{http://commoncrawl.org}} a company that builds and maintains 
public datasets crawled from the Web. Common Crawl 42 is a smaller version trained on 42 
billion tokens and has a vocabulary size of 1.9 million words. Both Common Crawl 840 and 
Common Crawl 42 contain word vectors of 300 dimensions.  
%
The collection of Twitter tweets was also trained with Glove. The training bundle was made 
up of texts from two billion tweets, totaling in 27 billion tokens. A total of 1.2 million unique 
word vectors were generated. The four collections of embeddings that were produced contain
vectors of 25, 50, 100 and 200 dimensions each.  
%
Text8Corpus was used to observe the role of corpus size in the quality of  generated embeddings.
It is a smaller text collection consisting of 17 million tokens and 25,000 words only.
Text8Corpus was trained with Skip-Gram method and various parameters. 
%
The last model is MoodyCorpus,\footnote{\url{http://softeng.polito.it/erion/MoodyCorpus.zip}} 
a text bundle of song lyrics created following the work 
in \cite{ismsi17}. Its biggest part is the collection of Million Song 
Dataset.\footnote{\url{https://labrosa.ee.columbia.edu/millionsong/}}
Songs texts of different genre and epoch were added to have more diverse texts. 
Regarding text preprocessing, punctuation, numbers or other symbols, as well as 
text in brackets were cleared out and everything was lowercased. The output 
consists of 90 million tokens and 43,000 unique words. 
%
%
\subsection{Sentiment Analysis Tasks}
This section describes the sentiment analysis experiments with song lyrics and 
movie reviews that were conducted to observe patterns relating training method, 
corpus or vocabulary size and thematic relevance of texts with the performance of 
generated word vectors. 
%
The first set of experiments were conducted on sentiment polarity analysis of song lyrics.
The dataset of songs described in \cite{7536113} was utilized for one set of 
experiments. The original version contains 771 lyrics classified by three human experts.
Here, a balanced version of 314 \emph{positive} and 314 \emph{negative} songs (A628) 
is used instead. 
%
The second set of lyrics experiments was conducted using MoodyLyrics (here ML3K),  
a collection of 3,000 lyrics that was created in \cite{pub2669975}. It contains songs of 
different epochs, dating from the sixties to the current days.   
%
%
\par 
%
The second experimentation task consists in identifying the polarity of movie review
documents. Seminal work on movie reviews has been conducted by 
Pang and Lee in \cite{Pang+Lee+Vaithyanathan:02a} and \cite{Pang+Lee:04a}. 
They published sentiment polarity dataset which contains 2,000 movie reviews
labeled as positive or negative. They also constructed and released subjectivity dataset 
consisting of 5,331 subjective and 5,331 objective sentences using the same method. 
The work of Pang and Lee created the road-map for a series of similar studies which 
apply various techniques to the problem. 
%
%
Use of neural networks and word embeddings for analyzing movie reviews
appeared on more recent works like \cite{shirani2014applications}. In that 
paper, authors explore RNNs (Recurrent 
Neural Networks), and CNNs (Convolutional Neural Networks). 
%
%
Another very important work that was conducted is \cite{maas-EtAl:2011:ACL-HLT2011} 
where authors create a large dataset of 50,000 movie reviews crawled from Amazon IMDB.
That dataset has become very popular among researchers, serving as a benchmark for 
various studies such as \cite{pouransari2014deep}, \cite{DBLP:conf/naacl/Johnson015}, etc. 
Here we used a subsets of 10,000 (MR10K) reviews and the full set of 50,000 (MR50K) 
movie reviews for the experiments.
\par 
%
Every text of the four datasets (A628, ML3K, MR10K, and MR50K) was first cleaned 
and tokenized. The dataset of each experimentation set is loaded and a set of unique 
words is created. The 14 models of word embeddings are also loaded and a $15^{th}$ 
model (\emph{self\_w2v}) is trained with Skip-Gram using the corpus of the current set. Every 
line of the pretrained models is analyzed, splitting apart the word and the corresponding 
vector and building \{word: vec\} dictionary that is later used for producing the 
classification features.
%
%
To have low variance in the obtained accuracy scores, we decided to use random forests with 
many (150) estimators as classifiers on each experiment.  
Classification models were prepared using tf-idf vectorizer which has been successfully 
applied in many relevant works like \cite{DBLP:conf/ismir/HuDE09} or \cite{Zaanen:10}.
It has been shown to perform better than other schemes like boolean scoring (presence or 
absence of certain words) or term frequency alone \cite{He2008}.
Tf-idf was applied to both words (for semantic relevance) and their corresponding 
vectors (for syntactic or contextual relevance). Average 5-fold cross-validation scores are 
calculated and reported for each of the models. 
%
%
\subsection{Results and Conclusions} 
\begin{figure}[!t]
	\centering
	\begin{minipage}{.51\columnwidth}
		\centering
		\includegraphics[width=0.97\columnwidth]{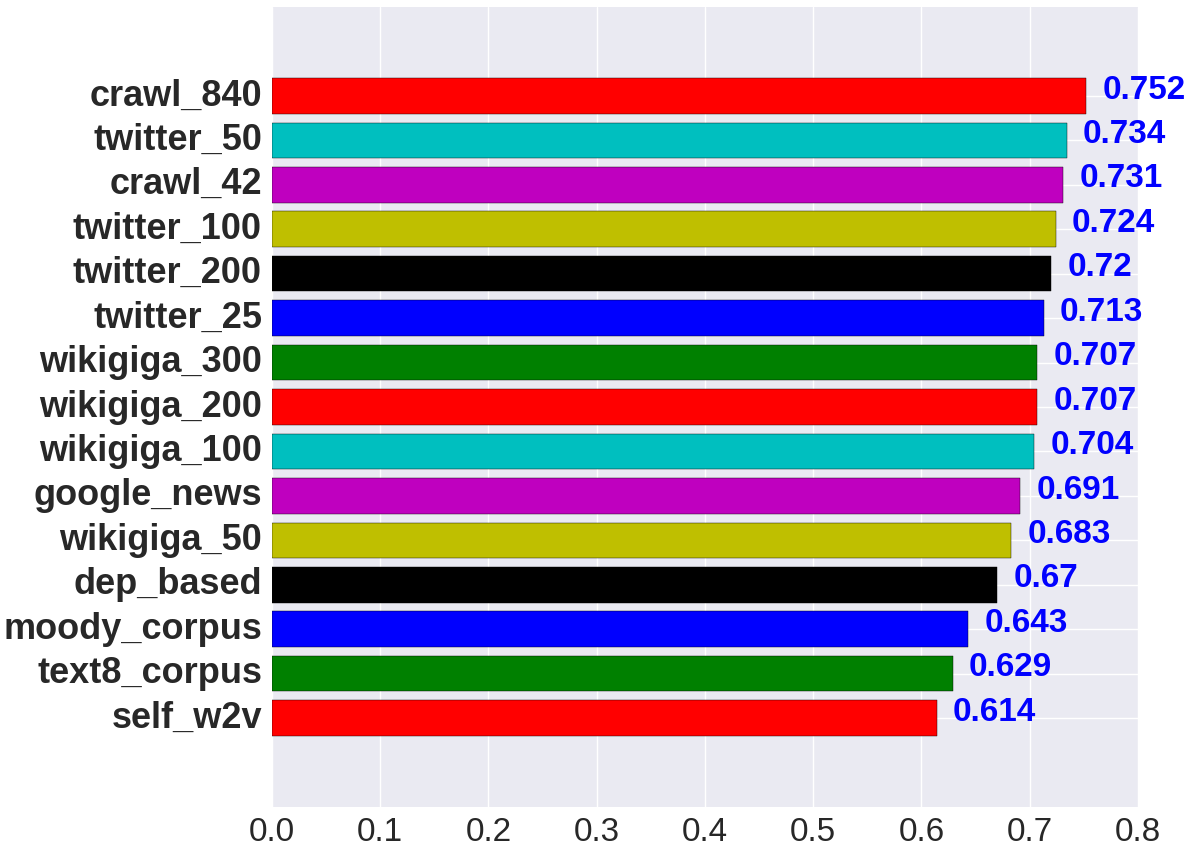}
		\caption{Lyric accuracies on A628 }
		\label{fig:embeddingLyrics1}
	\end{minipage}%
	\begin{minipage}{.51\columnwidth}
		\centering
		\includegraphics[width=0.97\columnwidth]{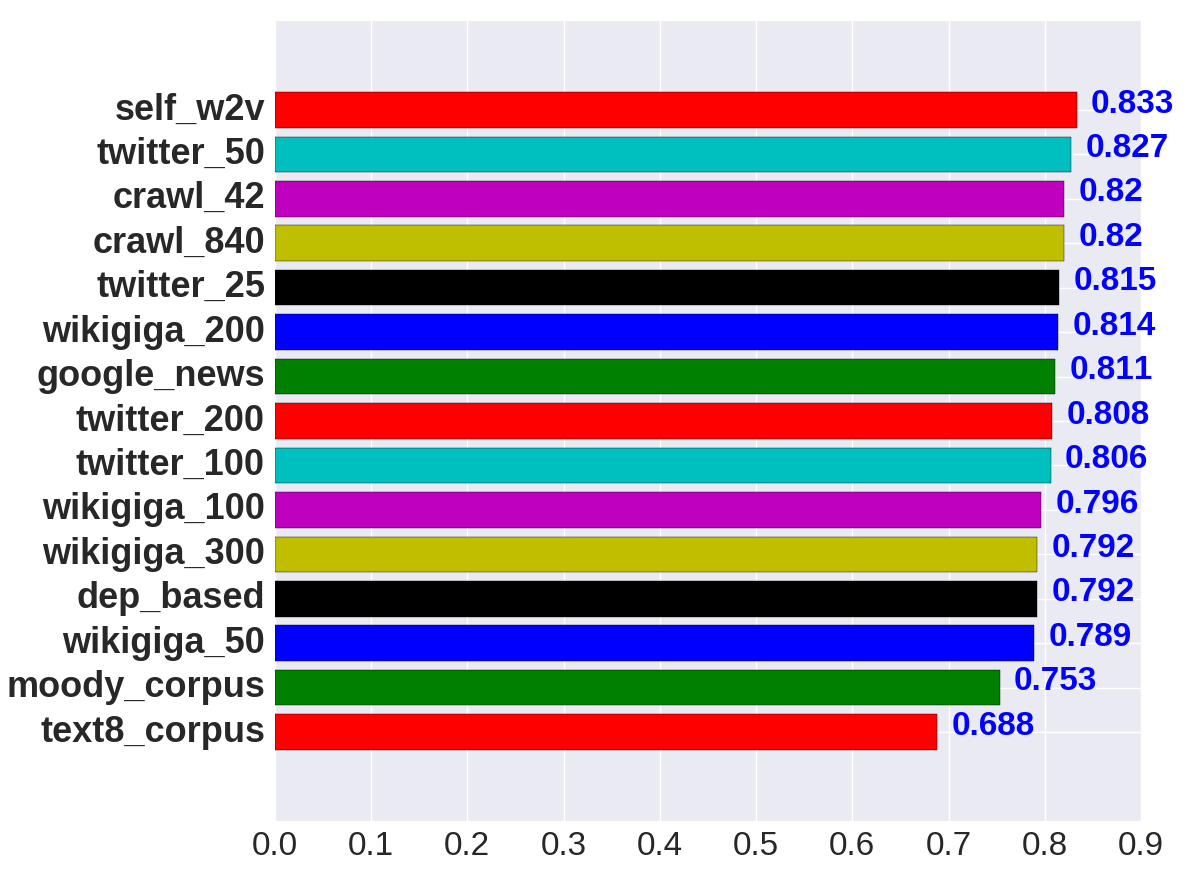}
		\caption{Lyric accuracies on ML3K}
		\label{fig:embeddingLyrics2}
	\end{minipage}
\end{figure}

%
%
Figures~\ref{fig:embeddingLyrics1} and~\ref{fig:embeddingLyrics2} show  
5-fold cross-validation accuracy results on the two song lyrics datasets.
Obviously, the top three models are \emph{crawl\_840}, \emph{twitter\_50}, 
and \emph{self\_w2v}. On A628 (smallest dataset) it is the biggest model (\emph{crawl\_840}) that 
leads, followed by \emph{twitter\_50}. \emph{Self\_w2v} is positioned at the bottom of the list,
severely penalized by the size of the A628 dataset it is trained on.   
The situation completely shifts in ML3K (larger dataset) where \emph{self\_w2v} comes at 
the top of the list, followed by \emph{twitter\_50}. We also see that other models like 
\emph{google\_news}, \emph{wikigiga} and \emph{dep\_based} are positioned in the middle of the list.
The worst models are those trained on Text8Corpus and MoodyCorpus bundles, 
with accuracy scores between  0.62 and 0.75.
%
In fact, \emph{self\_w2v} was generated from the texts of each experimentation dataset and thus depends 
on the size of that dataset. Its leap from the bottom to the top of the list with accuracy 
scores rising from 0.61 to 0.83 suggests that there is a strong correlation between corpus 
size and model accuracy. 
Top accuracy scores obtained here are similar to those reported in analogous works like 
\cite{10.1109/ISM.2009.123} where a dataset of 1032 songs is utilized.  
%
%
%
%
\par
\begin{figure}[!t]
	\centering
	\begin{minipage}{.51\columnwidth}
		\centering
		\includegraphics[width=0.97\columnwidth]{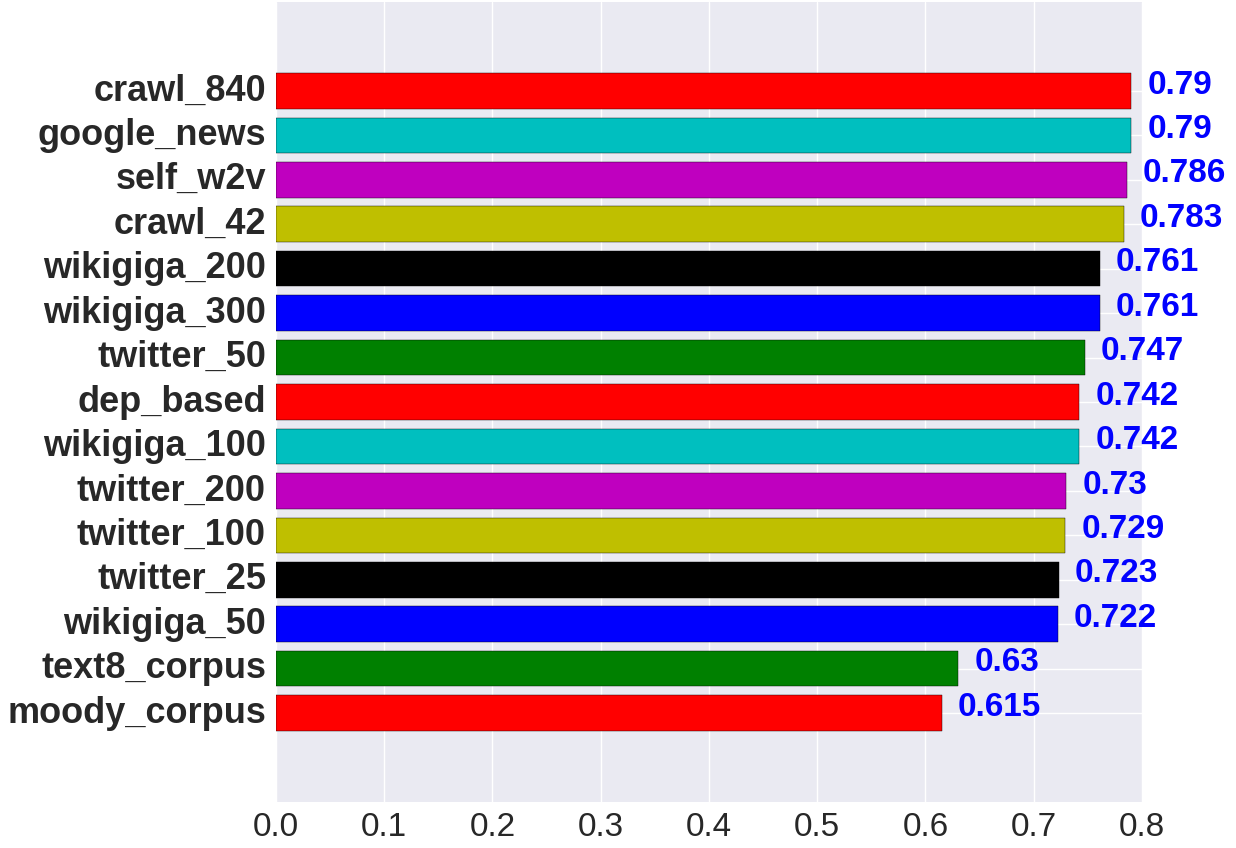}
		\caption{Review accuracies on MR10K}
		\label{fig:embeddingMovies1}
	\end{minipage}%
	\begin{minipage}{.51\columnwidth}
		\centering
		\includegraphics[width=0.97\columnwidth]{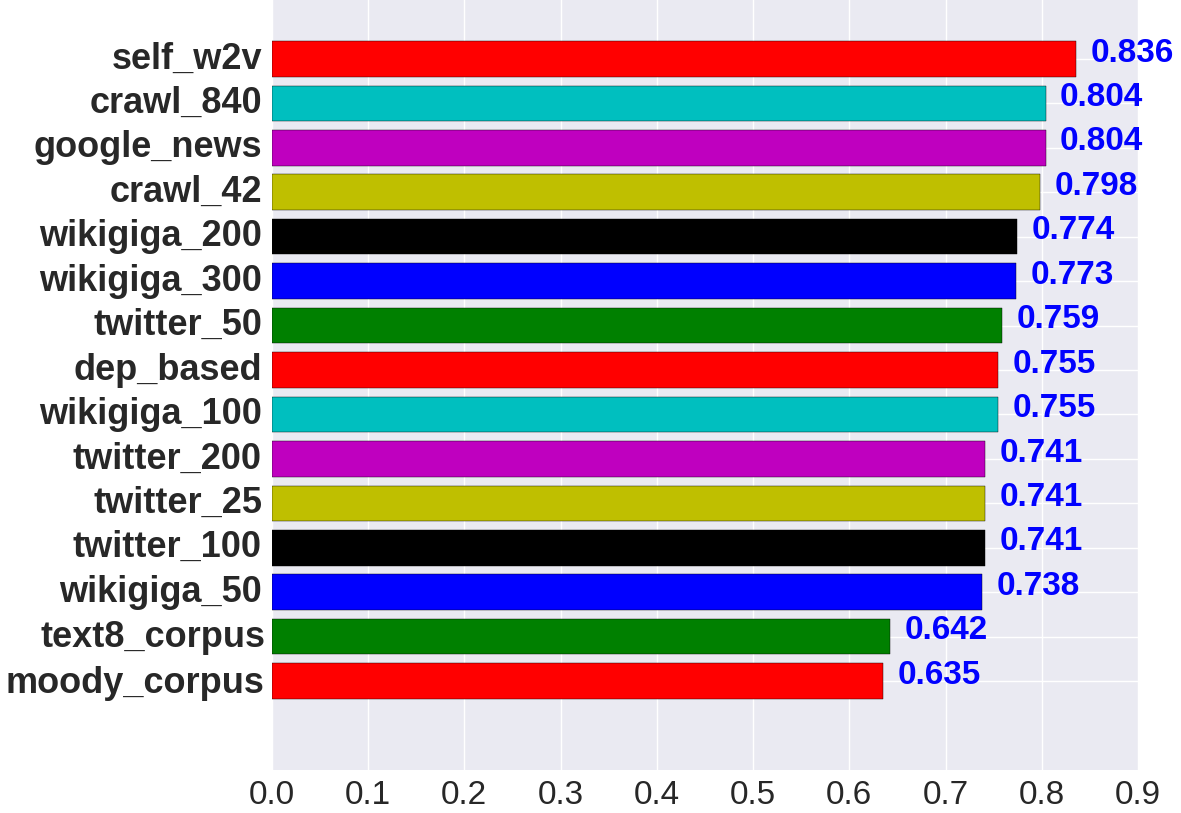}
		\caption{Review accuracies on MR50K}
		\label{fig:embeddingMovies2}
	\end{minipage}
\end{figure}

Accuracy results of the experiments on move reviews are presented in 
Figures~\ref{fig:embeddingMovies1} and~\ref{fig:embeddingMovies2}.
%
%
%
We once again see that \emph{crawl\_840} performs very well. Also, \emph{google\_news} is positioned 
among the top. Twitter models, on the other hand, are located in the middle of the list. We see that  
\emph{self\_w2v} grows considerably from the small to the bigger dataset, same as in 
the experiments with lyrics. In MR50K it has a discrete margin of more than 3\%
from the $2^{nd}$ position. Once again, \emph{wikigiga} models are positioned in the 
middle of the list and the worst models are Text8Corpus and MoodyCorpus. 
They perform badly on this task with a top accuracy of no more than 0.64 and a 
deficiency of 10\% from the closest model. 
%
There are also other studies that analyse movie reviews. Some of them have obtained 
higher scores than those reported here. Authors in \cite{maas-EtAl:2011:ACL-HLT2011} 
for example, apply a probabilistic model able to capture semantic similarity between words 
and report an accuracy of 0.88.  
Also in \cite{DBLP:conf/naacl/Johnson015}, they
combine bag-of-3-grams with a CNN of three layers and achieve an accuracy of 0.92. 
In \cite{pouransari2014deep} authors utilize a very similar experimentation setup with the 
one that was used here. They feed word vector features to a random forest classifier and achieve 
0.84 accuracy on movie reviews.  
%
%
%
%
%
%
%
%
\par 
To answer the three research questions, we had to check the statistical significance of the 
above results. Regarding RQ1 and training method, we set the following null hypothesis:
\begin{enumerate}
\item[$\mathbf{Hm_0}$] \emph{There is no significant difference between accuracy scores
of word embeddings trained with Glove and those trained with Skip-Gram.}
\end{enumerate}
\begin{table}
	\caption{Google News compared with Common Crawl} 
	\centering
	\label{table:wembMethod}
	\begin{tabular}{c | c c c l}
		\toprule
		Test & avg & t & p & ~~~~~~Verdict \\  
		\midrule
		A628 & 0.691 & 5.683 & 0.00047 & $Hm_0$ rejected   \\
		
		ML3K & 0.811 & 4.913& 0.00119 & $Hm_0$ rejected  \\
		
		MR10K & 0.79 & 1.462 & 0.182 & $Hm_0$ not rejected  \\
		
		MR50K & 0.804 & 1.547 & 0.161 & $Hm_0$ not rejected   \\
		\bottomrule
	\end{tabular}
\end{table}
For a fair comparison, we picked up results of Common Crawl 840 and Google News corpora.
These two bundles were both created from web and news texts which are not of a particular 
topic. We performed t-test analysis, comparing 
the values of \emph{google\_news} model with those of \emph{crawl\_840} in the four experiments. 
An $\alpha = 0.05$ was chosen as the level of significance. 
They have similar vocabulary sizes and differ mainly in the method they were trained with. 
Statistical results are shown in Table~\ref{table:wembMethod}. The $avg$ values are basically the 
5-fold cross-validation results reported in the figures above. As we can see, in the first two 
experiments (songs), $t$ values are much greater than $p$ values. Also, obtained $p$ values are much smaller than the pre-chosen value of $\alpha$ (0.05). As a result, we have evidence
to reject the null hypothesis and confirm that indeed \emph{crawl\_840} performs better than 
\emph{google\_news}. On the other two experiments, we have a totally different picture. Both models
got same average scores and have similar $t$ and $p$ values, with the latter being considerably greater than $\alpha$. As a result, we do not see significant difference between them on movie reviews.  
%
In conclusion, we can say that word embeddings trained with 
Glove slightly outrun those trained with Skip-Gram on polarity 
analysis of song lyrics but perform the same on movie reviews. 
\begin{table}[ht] 
	\caption{Properties of self\_w2v models}  
	\small 
	\centering    
	\setlength\tabcolsep{2.7pt}  
	\begin{tabular}
		{c c c c c c}  
		\toprule
		\textbf{Trial} & \textbf{Dataset} & \textbf{Dim} & \textbf{Size} & \textbf{Voc} & \textbf{Score}		 	\\ [0.5ex] 
		\midrule  
		1 & AM628 & 200 & 156699 & 8756	& 0.614	  \\
		2 & ML3K & 200 & 1028891 & 17890 & 0.833		  \\
		3 & MR10K & 300 & 2343641 & 53437 & 0.786	  \\ 
		4 & MR50K & 300 & 11772959 & 104203	& 0.836	  \\
		\bottomrule
	\end{tabular} 
	\label{table:SelfModelProps}
\end{table}
\par 
Regarding RQ2 and the effect of training corpus size, the results seem more convincing. 
Biggest models like \emph{crawl\_840} appeared among the best in every set of 
experiments whereas MoodyCorpus and Text8Corpus that are the smallest were  
always positoned at the bottom of the list. Moreover, \emph{self\_w2v} performed very well 
in the second experiment of each task where it was trained on the bigger datasets. The 
properties of \emph{self\_w2v} on each experiment are summarized in Table~\ref{table:SelfModelProps}. 
Despite these arguments, we still conducted a statistical examination, formulating and checking the 
following null hypothesis:
\begin{enumerate}
	\item[$\mathbf{Hs_0}$] \emph{There is no significant difference between accuracy scores
		of word embeddings trained on text corpora of different sizes.}
\end{enumerate}
\begin{table}
	\caption{crawl\_42 compared with crawl\_840}
	\centering
	\label{table:wembSize}
	\begin{tabular}{c | c c c l}
		\toprule
		Test & avg & t & p  & ~~~~Verdict \\  
		\midrule
		A628 & 0.731 & 4.97 & 0.0011 & $Hs_0$ rejected \\
		
		ML3K & 0.82 & 0.86 & 0.414 & $Hs_0$ not rejected \\
		
		MR10K & 0.783 & 2.56 & 0.033 & $Hs_0$ rejected \\
		
		MR50K & 0.798 & 2.96 & 0.027 & $Hs_0$ rejected \\
		\bottomrule
	\end{tabular}
\end{table}
The fairest comparison we could make here is the one between \emph{crawl\_42} and \emph{crawl\_840}
models. They were both trained using Glove on Common Crawl texts. The only difference
is in the size of corpora they were derived from. 
Statistical results are shown in Table~\ref{table:wembSize}. Experimental results on 
A628, MR10K, and MR50K indicate that the accuracy difference between the two models 
is significant and the null hypothesis can be rejected. Same is not true about 
results on ML3K. All in all, we can confirm that training corpus size has a strong influence 
on the performance of word embeddigns. 
%
Furthermore, it comes out that the choice between 
pretrained vectors and vectors generated from the training dataset itself also depends on 
the size of the latter. When the training dataset corpus is big enough, using it to generating 
word vectors is the best option. If it is small, it might be better to source word vectors from 
available pretrained collections.  
\par 
Regarding RQ3 and thematic relevance, we also see that pretrained models behave 
differently on the two tasks. Twitter corpora performed better on lyrics. They are large 
and rich in vocabulary with texts of an informal and sentimental language. This language 
is very similar to the one that is found on song lyrics, with \emph{love} being the predominant 
term (word cloud in \cite{ismsi17}, p. 5). 
%
Common Crawl and Google News that are trained with more informative texts performed 
best on movie review analysis. These results indicate that topic of training texts may have 
an influence on the generated word features. Nevertheless, it was not possible here to have a rigorous and fair analysis of it. Obviously, a more extensive experimental work 
that excludes the effect of other factors is required.

\chapter{Sentiment Analysis via Convolution Neural Networks}
\label{chapter6}
\ifpdf
    \graphicspath{{Chapter6/Figs/}{Chapter6/Figs/PDF/}{Chapter6/Figs/}}
\else
    \graphicspath{{Chapter6/Figs/Vector/}{Chapter6/Figs/}}
\fi
%
\begin{flushright}
	\emph{\large \dq{You can use the power of words to bury meaning or to excavate it.}} \\
	-- Rebecca Solnit, \emph{Men Explain Things to Me}, 2014 
\end{flushright}
\vskip 0.35in
\indent \indent
Dense vector representations of words are well interpreted from neural 
network layers. Convolutional and max-pooling neural networks, for example, 
are capable of capturing relations between words and sentiment categories of 
the document. They exhibit very good performance on text mining tasks
when combined together. Before building sentiment analysis models 
based on neural networks, it is essential to know how we can adapt the 
large set of network hyperparameters to the varying sizes and document 
lengths of the training data. To address this issue, we performed several experiments
where simple neural networks of stacked convolutional and max-pooling layers are 
trained with different dataset sizes and document lengths. 
\par 
This chapter is organized as follows. Section~\ref{sec:DataPropIntro} introduces 
convolutional neural networks and relevant studies that successfully apply them for 
text mining tasks. Training datasets, as well 
as data statistics and text preprocessing steps, are described in Section~\ref{sec:DataPrepAndStats}. 
Section~\ref{sec:CNNexperimentalSetup} presents the neural network structure we 
experiment with, made up of stacked convolutional and max-pooling layers, as well as the hyperparameter setup alternatives. Finally, Section~\ref{sec:CNNresultsDiscussion} 
summarizes and further discusses the obtained relations between data properties and 
performance. 
\section{Neural Network Types for Text Analysis}  
\label{sec:DataPropIntro}
%
Neural networks are excelling in various complex tasks such as speech recognition,
object detection, machine translations or sentiment analysis. Intelligent personal 
assistants, self-driven cars or chatbots passing Turing Test are examples of the 
deep learning hype. This success is mostly attributed to the ability 
of deep neural networks to generalize well when fed with tons of data.  
%
Convolutional Neural Networks (CNN) are particularly effective in mimicking the 
functioning of the visual cortex in the human brain, becoming great players on computer 
vision applications. The basic structure of CNNs was proposed by LeCun \emph{et al.} 
in \cite{lecun1998gradient} and was utilized for recognizing images of 
handwritten digits. After a decade of lethargy (the second AI winter),
they came back in the late 2000s, becoming the crucial part of several advanced 
image analysis architectures. 
%
\par
\begin{figure}[h!]
	\centering
	\includegraphics[width=0.87\textwidth]{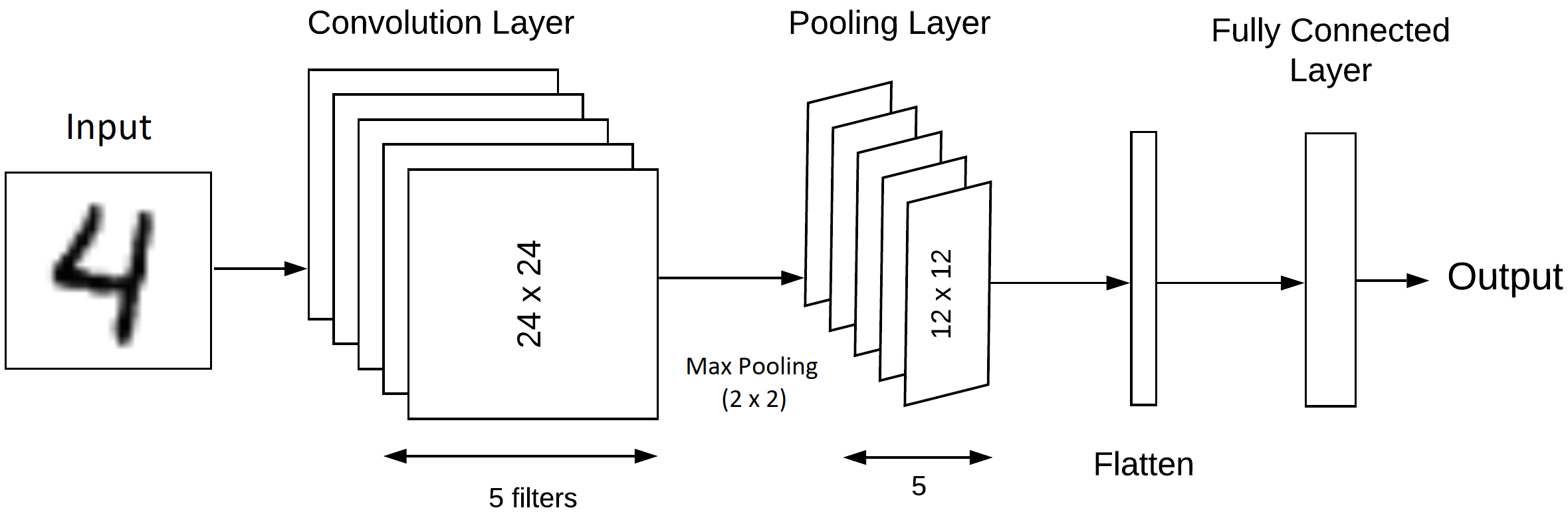}
	\caption{Simple convolution-pooling network for image recognition}
	\label{fig:CnnPool}
\end{figure}

\begin{figure}[h!]
	\centering
	\includegraphics[width=0.57\textwidth]{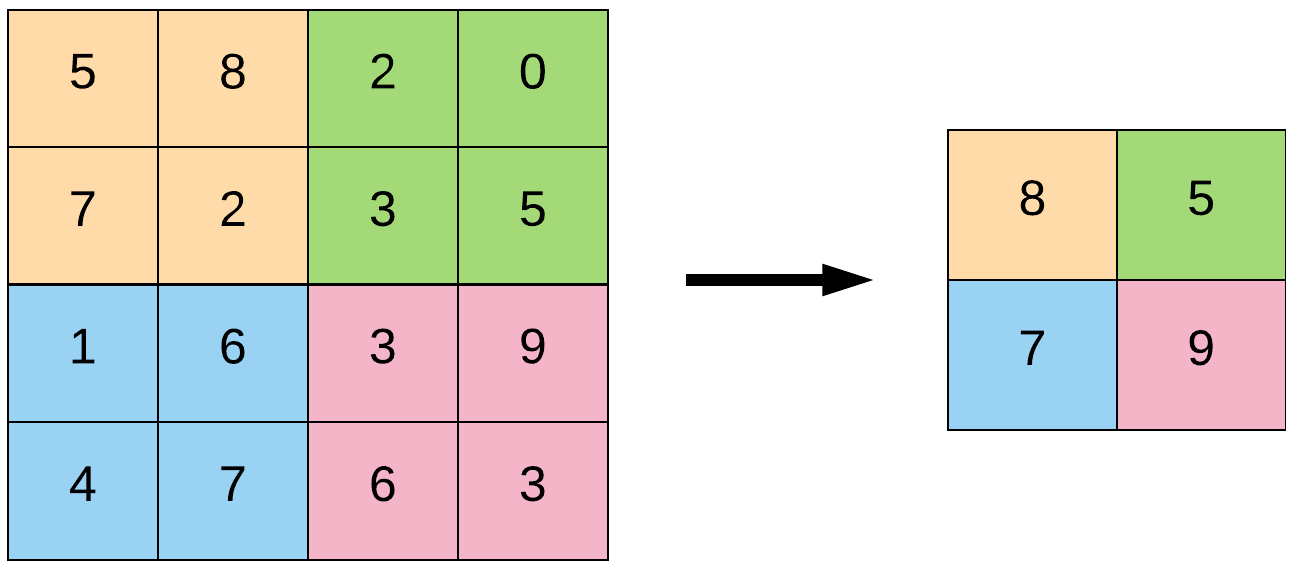}
	\caption{Illustration of a $2~x~2$ max-pooling operation}
	\label{fig:MaxPool}
\end{figure}

 
Figure~\ref{fig:CnnPool} shows a neural network based on convolutional and max-pooling
layers for classification of digit images. The convolution operation applies the filter 
over the bigger matrix of pixels moving forward at a certain stride and producing feature 
maps. This operation extracts features by preserving the spacial relations between them. 
Later on, a max-pooling operation is applied on the generated feature maps to downsample 
(reduce) their dimensionality by hopefully selecting the most relevant ones. It is illustrated in 
Figure~\ref{fig:MaxPool}. Afterwards, all feature maps are flattened (reducing dimensionality
from 3 to 2) and pushed to the dense layer that serves as a simple classifier. 
%
The neural network of Figure~\ref{fig:CnnPool} can be applied on text analysis 
as well. The only difference is in the input data and target categories. In general, if 
we have a document of $n$ words with vectors of $d$ dimensions each, we can apply on it 
a filter (usually $m$ filters) $w \in \mathbb{R}^{d \times k}$ on all windows of $k$  
consecutive words. The output is a feature map $f = [f_1, f_2, ..., f_{n - k + 1}]$ that is 
passed to the max-pooling layer. If simple max-pooling is used, than we get only one   
$f' = max(f)$. Otherwise, if we apply regional max-pooling with regions of 
size $R$, the output we get will be a set of $p$ max features $f' = [f'_1, f'_2, ..., f'_p]$, 
where $p = ceil(\frac{n - k + 1}{R})$ is the number of the resulting regions. 
\par 
Recurrent Neural Networks (RNN) represent another network family 
that has gained significant popularity in the recent years. 
They perform especially well with data that exhibit continuity 
in time, like words coming one after another as a sequence. This family of neural 
networks makes use of feedback loops that behave like \dq{memory} for 
maintaining the already seen data. Long Short-Term Memory (LSTM) networks
introduced in \cite{Hochreiter:1997:LSM:1246443.1246450} are probably the 
most popular and highly utilized version of RNNs. However, the common problem of 
RNNs is their limited memorization ability. In practice, they can look back to 
few steps and are thus not able to represent well long text sequences.    
Furthermore, RNNs are slower to train compared with CNNs which are simpler 
and faster. Examples of sentiment analysis studies utilizing RNNs are \cite{zhang2016gated} 
or \cite{Lai:2015:RCN:2886521.2886636} where they make use of gated networks. 
Also, similar studies like \cite{conf/coling/ZhouQZXBX16} employ various 
combinations of CNNs with RNNs to achieve the same goal.  
\par  
Application of CNNs for text analysis was initially impeded by problems like 
small available text datasets, high dimensionality of word features etc. 
One of the first works that successfully used them for sentiment analysis was proposed by 
Kim in \cite{DBLP:journals/corr/Kim14f}. A basic convolutional network was applied
on short sentences to extract and select word features and a simple dense layer
was used as a classifier. The author reported state-of-the-art accuracy results with 
little computation load on datasets containing from 3375 to 11855 documents.  
%
Other studies like \cite{Zhang:2015:CCN:2969239.2969312} and more recently 
\cite{DBLP:conf/eacl/SchwenkBCL17} used deeper architectures of 
9 -- 29 layers, starting from characters and building up word patterns utilized as 
classification features. They report competitive results on large datasets with hundred 
thousands of text documents. However, on smaller datasets, their networks 
are weaker than shallow counterparts which start from words as basic language elements. 
%
The debate of shallow word-level versus deeper char-level CNN-based networks was 
further disputed in \cite{DBLP:journals/corr/Johnson016a}. Authors try shallow 
word-level CNNs on same big datasets used in \cite{Zhang:2015:CCN:2969239.2969312} 
and \cite{DBLP:conf/eacl/SchwenkBCL17}. They report that shallow word-level CNNs
are not only more accurate but also compute faster than the deeper char-level CNNs.
Obviously text words do comprise a semantic value that is lost when analysis starts 
from characters. The only drawback of word-level CNNs is their high number of parameters 
required for the word representations, which counts for higher storage requirements. 
Here (and in the next chapter) we favor and experiment with simpler word-level CNNs. 
\par 
It is important to note that applying convolutional and pooling neural layers for 
analyzing sentiment polarity of texts is not easy. For example, suppose we are 
using the simple architecture of Figure~\ref{fig:CnnPool} with length (depth)
$L = 2$ made up of only one convolutional and one max-pooling layer. In this 
simplified scenario, we still have to pick many parameter values such as number 
of filters $m$, filter (kernel) length $k$, filter stride $s$, pooling region size $R$, 
dropout, $L_1$ and $L_2$ regularization norms to avoid overfitting as well as 
other parameters or functions used to train the network. This high number 
of parameters makes it hard to find the optimal performance setup. 
Moreover, performance and network parameters are strongly related with input data 
metrics such as the size of the training set, length of the document etc. For this reason, 
in \cite{worldcist18} we conducted multiple experiments that relate data properties and 
network parameters with top classification accuracy. The goal was to observe 
any patterns that could be useful for simplifying optimal network creation
by providing template parameter setups, answering the following questions:  
%
\begin{description}
\item[RQ1] \emph{What is the relation between pool region size ($R$) and training text length 
with respect to optimal accuracy score?}
\item[RQ2] \emph{What are the effects of convolutional kernel size ($k$) and network width
($W$) on accuracy score?} 
\item[RQ3] \emph{What is the relation between network depth ($L$) and training dataset size
with respect to optimal accuracy score?}
\end{description}
In the forthcoming sections, we describe the steps that were followed, 
obtained experimental results and some concluding remarks addressing the three 
research questions listed above.  
\section{Data Processing and Statistics}
\label{sec:DataPrepAndStats}
%
%
\subsection{Experimental Datasets} 
\label{sec:ExpDatasetsRoleNcnn}
%
For the experiments, public datasets of various sizes and document lengths 
were chosen. They comprise text excerpts of different thematic contents such as 
song lyrics, movie reviews, smartphone reviews etc. 
%
\begin{description} 
%
\item [Mlpn] This dataset is the collection of song lyrics from MoodyLyricsPN dataset
described in Section~\ref{sec:TagDatasets}. There are 2,500 \emph{positive} and 
2,500 \emph{negative} labeled songs in it. All lyrics were crawled from online 
music portals and are used for experimental purposes only.  
%
\item [Sent] Sentence polarity dataset was one of the first experimental text collection,
created by Pang and Lee back in 2005 \cite{Pang:2005:SSE:1219840.1219855}. 
There are 5331 \emph{positive} and 5331 \emph{negative} texts extracted from IMDB 
archive of movies. The reviews are short and consist of one sentence and about 10 -- 20 
words only.  
%
\item [Imdb] This is the IMDB dataset of movie reviews described in  \cite{Maas:2011:LWV:2002472.2002491}. It contains 50,000 movie reviews
that were manually labeled as \emph{positive} or \emph{negative} and has been used as 
ground truth in many sentiment analysis studies. The dataset is also available in various
libraries, preprocessed and ready for use.  
%
\item [Phon] This dataset contains user reviews of unlocked smartphones sold in 
Amazon. Users usually provide text comments, a 1--5 star rating or both for products
they purchase. The dataset contains entries with both star rating and text description. 
All reviews with a 3-star rating were cleared out to better separate 1-star and 2-star 
reviews which were considered as \emph{negative}, from 4-star and 5-star reviews
that were considered as \emph{positive}. A final number of  232,546 reviews was 
reached. 
%
\item [Yelp] This is the biggest dataset that was used. It was created 
by Zhang \emph{et al.} in \cite{DBLP:journals/corr/ZhangZL15} and contains 
598,000 Yelp user reviews about businesses such as restaurants, hotels, etc. 
All reviews were balanced and labeled with the corresponding emotional polarity. 
\end{description}
%
It is important to note that the first three datasets were created using systematic 
methods involving human judgment for the labeling process. They can be thus used 
as ground truth for cross-interpretation of results. The same thing is not certainly true about
the last two which are labeled considering user star ratings only. In fact, there is no proven 
guarantee that few stars do necessarily denote \emph{negative} language and many stars a 
\emph{positive} one. All affect datasets created by Zhang \emph{et al.} in \cite{Zhang:2015:CCN:2969239.2969312} (here we use Yelp only) are generated in that 
way. They are being used in several studies (e.g., \cite{DBLP:conf/acl/JohnsonZ17} or \cite{DBLP:journals/corr/Johnson016a}) which do not acknowledge that limitation. 
While it is highly desirable to have big labeled datasets for experimentation, creating them 
based on simplistic heuristics like star ratings only is probably not enough and appropriate 
data quality examinations are required. In fact, lack of big and professionally labeled text 
datasets was one of the reasons why we chose to focus on shallow word-level network 
architectures.   
\begin{table}[ht] 
	\caption{Document length statistics for each dataset}  
		\small 
	\centering    
	\setlength\tabcolsep{2.7pt}  
	\begin{tabular}
		{l | c c c c c}  
		\toprule
		\textbf{~~~~~~~Dataset} & \textbf{Docs} & \textbf{MinLen} & \textbf{AvgLen} & 
		\textbf{MaxLen} & \textbf{UsedLen}		 	\\ [0.1ex]
		\midrule   
		Song Lyrics & 5K & 23 & 227 & 2733 & 450	  \\ [0.3ex]
		Sentence Polarity & 10K & 1 & 17 & 46 & 30	  \\  [0.3ex]
		Movie  Reviews & 50K & 5 & 204 & 2174 & 400		  \\ [0.3ex]
		Phone Reviews & 232K & 3 & 47 & 4607 & 100	  \\  [0.3ex]
		Yelp Reviews & 598K & 1 & 122 & 963	& 270	  \\ 
		\bottomrule
	\end{tabular} 
\label{table:MP8documentStatistics}
\end{table}
\subsection{Preprocessing and Statistics}
\label{sec:DataPrepSteps}
%
The usual text preprocessing steps were performed on each document of the five 
datasets. Firstly, any remaining html tags (many texts were crawled from websites) were 
removed. Smiley symboles like :D, :-), :), :(, :-(, :P, were kept in, as they are very 
effective for emotion identification. Stopwords are usually discarded as they contain 
little or no semantic value. We cleared out the subset  
\{\dq{the}, \dq{these}, \dq{those}, \dq{this},  \dq{of}, \dq{at}, \dq{that}, \dq{a}, 
\dq{for}, \dq{an}, \dq{as}, \dq{by}\}. Presence of short form residues such as 
\dq{ll}, \dq{d}, \dq{s}, \dq{t}, \dq{m} 
and negation forms like \dq{couldn}, \dq{don}, \dq{hadn} or \dq{didn}
can completely shift the emotional polarity of a phrase or sentence. They were thus 
kept in. Finally, any remaining \dq{junky} characters were removed and 
everything was lowercased.   
\par 
Having every document in its final form, we further observed the length 
distributions summarized in Table~\ref{table:MP8documentStatistics}.
As we can see, song text lengths fall between 23 and 2733 words, averaging 
at 227. Smartphone and movie review lengths are highly dispersed.
The former range from 3 to 4607 with an average of 47. The later fall between  
5 to 2174 averaging at 204. Regarding Yelp reviews, we see that they are less 
dispersed spanning between 1 and 963 words. The dataset of sentence polarities 
is the most uniform, with document lengths from 1 to 46. A more detailed observation
of length distributions revealed that most documents are quite short. For example, 
in review datasets, there are very few documents longer than 500 words. For this 
reason, as well as to decrease computation load of the experiments, we clipped 
the few long documents and padded the shorter ones with zeros to reach uniform 
lengths. Actually, no more than 10\% of documents were clipped. The last column
of Table~\ref{table:MP8documentStatistics} reports experimentation lengths
of each dataset. 
\section{Experimental Setup}
\label{sec:CNNexperimentalSetup}
%
\subsection{Representation of Words}
\label{sec:WordRepGoogNews}
\begin{figure} 
	\centering
	\includegraphics[width=0.62\textwidth]{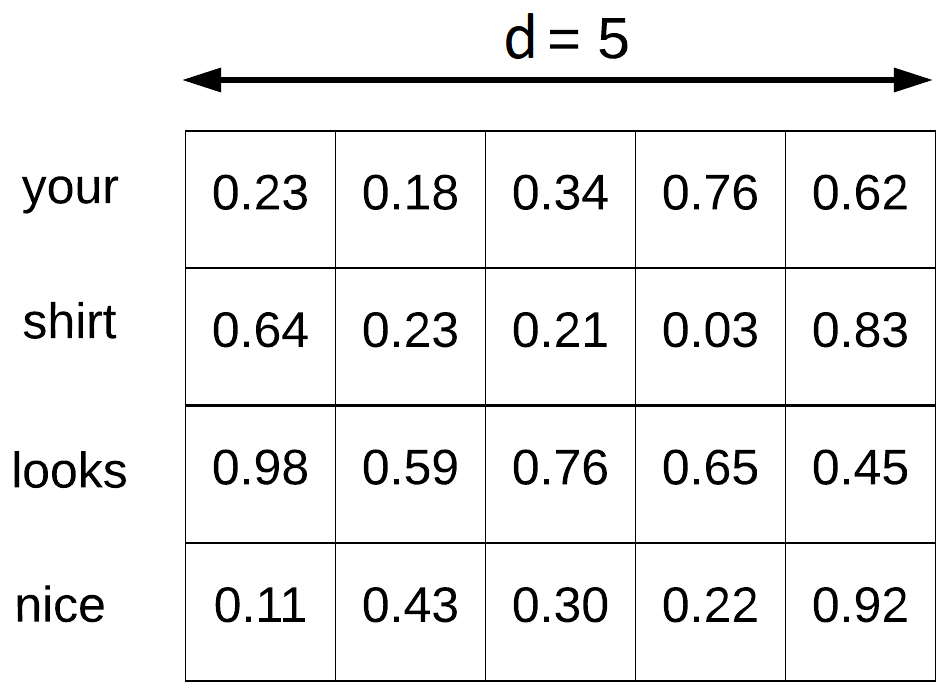} 
	\caption{Matrix representation of a short sentence}
	\label{fig:TextMatrix} 
\end{figure}

%
As explained in Section~\ref{sec:WordRepModels}, word embeddings or distributed  
word representations in general, are gradually replacing the bag-of-words model in 
many applications. They are dense and well suited for the common neural network
structures that are used today. The ability to retain syntactic and especially semantic
word relations makes them appropriate for representing documents on text mining tasks.  
For example, we can assume that for every possible word in our dataset documents, 
there is a word vector of $d$ dimensions generated with one of the methods described 
in Section~\ref{sec:PopularEmbMethods}. Then we can represent each of those documents 
as a matrix of numbers. The document representation becomes very similar to 
a matrix of pixels that represents an image. If we have $d = 5$ (in practice 100 -- 300), 
the representation of a short sentence like \dq{your shirt looks nice} will be as shown in 
Figure~\ref{fig:TextMatrix}. 
%
One possibility for obtaining word embeddings is to initialize them with random values 
and then utilize the neural network for tuning their values based on the appearance of the words in the
training dataset. This method can be applied when the supervised training dataset is enough
large. As suggested in \cite{pub2668229}, when relatively small labeled text sets are 
available, sourcing pretrained word vectors that were generated from big text corpora gives better 
results. The first three datasets of Table~\ref{table:MP8documentStatistics} are relatively
small. For this reason, we decided to utilize the word 
embeddings of 300 dimensions that are available in 
GoogleNews\footnote{\url{https://code.google.com/p/word2vec/}} collection. They were created 
from a text bundle of news documents containing 100 billion tokens and three million
unique words and phrases. Relevant studies like 
\cite{DBLP:journals/corr/LauB16} or \cite{DBLP:journals/corr/Kim14f} report excellent
results when using them on similar tasks.
\subsection{Data-driven Experimentation Networks}
\begin{figure}
	\centering
	\includegraphics[width=0.70\textwidth]{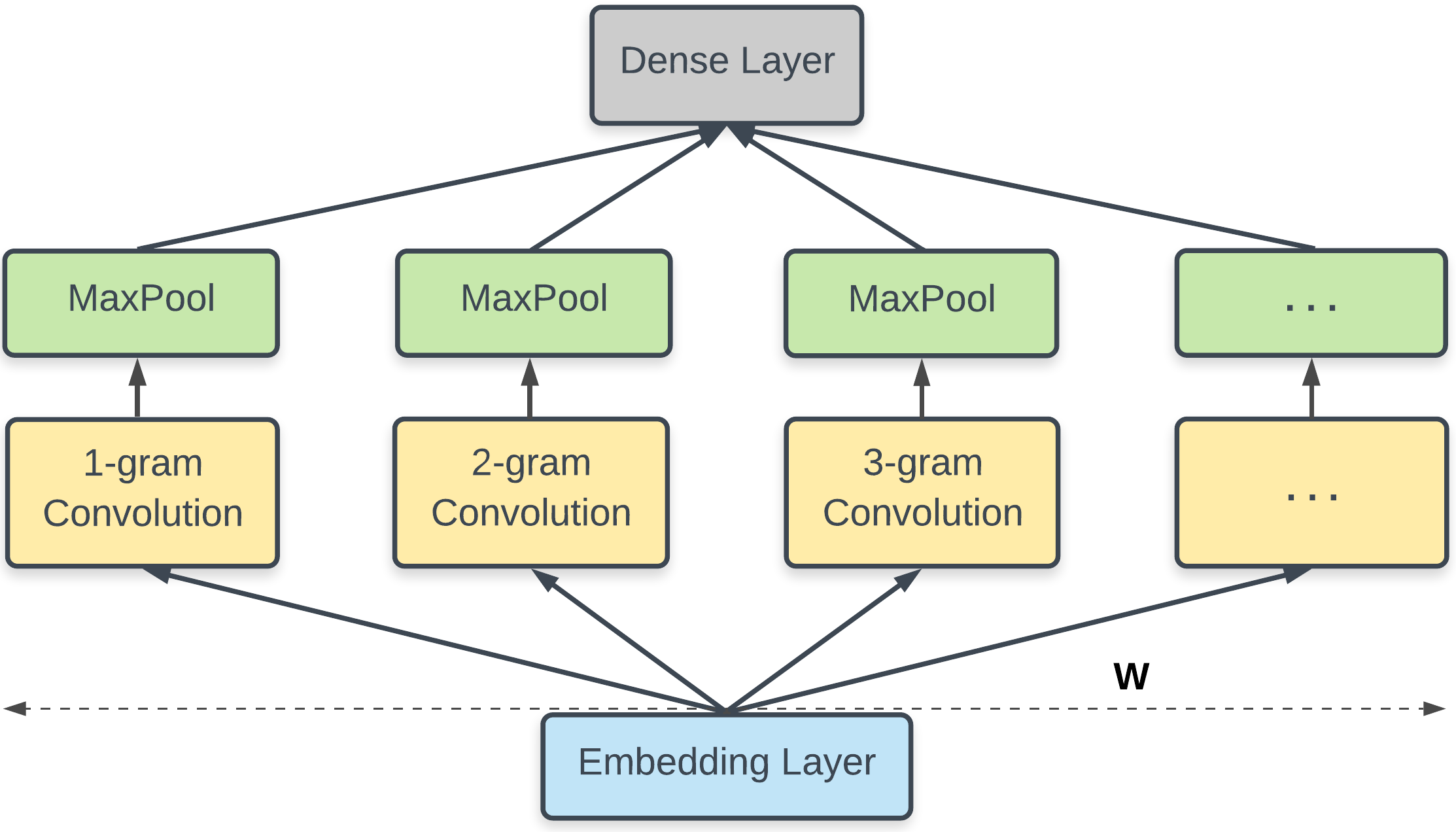}
	\caption{Basic neural network structure}
	\label{fig:BasicNetStruct}
\end{figure}

%
The basic network structure chosen for our experiments is presented in 
Figure~\ref{fig:BasicNetStruct}. The embedding layer is actually not trainable. It just uses
the corresponding word vector sourced from GoogleNews collection, for every word appearing
in the documents. Next come the convolutional layers that work as feature extractors. 
There is one stack with $W$ of them that are used in parallel, with filter size $k = 1$
for extracting word features, $k = 2$ for $2$-gram features, $k = 3$ for $3$-grams and 
so on. In fact, the added value of $2$-gram and $3$-gram features in performance have 
been pointed out in various studies like \cite{Wang:2012:BBS:2390665.2390688}. 
Each convolutional layer applies 70 filters to capture relations between feature maps 
and the \emph{positive} and \emph{negative} document categories. 
%
%
Regarding activation function of convolutions, 
$relu(x)=max(0, x)$, $tanh(x) = \frac{e^{2x} - 1}{ e^{2x}+ 1}$ and 
$softsign(x)=\frac{x}{1 + |x|}$ were explored. The best in most of trials was $relu$ 
which is the one we used. 
%
Convolutional layers are directly followed by max-pooling layers that subsample data, 
selecting the most salient features. For retaining local information of word combinations, regional
max-pooling with varying region size is used. Studies like 
\cite{DBLP:journals/corr/ZhangW15b} and \cite{DBLP:conf/eacl/SchwenkBCL17}
show that it outperforms other pooling methods such as k-max or average pooling.
Generated feature maps of length 
$p = ceil(\frac{n - k + 1}{R})$
are recombined together (flattened) and fed to a dense layer that serves as classifier.
Length (depth) of the network $L$ is the total number of convolutional and max-pooling layer 
stacks. As we can see, the network of Figure~\ref{fig:BasicNetStruct} has 
length $ L = 2 $.
%
For the classification, a single dense layer of 80 nodes was used in each experiment. 
Overfitting was mitigated using 0.1 $L_2$ regularization and 0.35 dropout norms. 
Finally, to compute loss and optimize network training, binary cross-entropy and Adam 
optimizer were employed. 
\par
\begin{figure}
	\centering
	\includegraphics[width=0.67\textwidth]{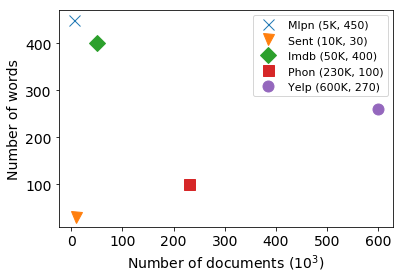}
	\caption{Size-length distribution of training datasets}
	\label{fig:size-words-plan}
\end{figure}

 
In Figure~\ref{fig:size-words-plan} we can see a planar projection of the training data 
length and size for each of the five datasets that were utilised. To find the optimal network 
structure and observe the role of dataset attributes in performance, alternative versions of 
the scheme in Figure~\ref{fig:BasicNetStruct} (one stack of convolution and max-pooling layers) 
were considered. The parameters of these network structures are driven by training data 
characteristics. For example, different values of the region 
size $R$ were used for adapting the network to long and short documents of the datasets. 
Usually bigger datasets (e.g., the last two of Table~\ref{table:MP8documentStatistics}) are better 
interpreted using deeper neural networks with more training parameters. As a result, a 
deeper network structure is obtained by duplicating the convolutional and 
max-pooling stacks in the same order. The rest of the networks that were tried are 
similar, changing only in max-pooling region sizes. 
During each experiment, a 70/10/20 percent data split for training, development, and testing 
was used respectively. The following section presents the obtained accuracy scores.  
%
%
%
\section{Results and Discussion} 
\label{sec:CNNresultsDiscussion}
%
Top neural networks and their accuracy score on each dataset are summarized in 
Table~\ref{table:mp8FiveNetworkScores}. Simple network structures with just one or two 
consecutive convolutional and max-pooling stacks produce very good results. Top score on phone 
reviews (Phon) seems excellent. Contrary, the one obtained on song lyrics (Mlpn) seems somehow 
disappointing. Unfortunately, we still do not have a comparison basis for these two datasets. On 
sentence polarity dataset (Sent), a peak score of 79.89\% was obtained. The many studies 
that have experimented with this dataset report accuracy scores from 76\% to 82\%. The 
very best result we found in the literature is 83.1\%. It was reported by Zhao \emph{et al.} in \cite{DBLP:journals/corr/ZhaoLP15}. In that study, they proposed a self-adaptive sentence 
model that forms a hierarchy of representations from words to sentences and then to entire 
document. Their model is implemented using gating networks.   
On movie reviews (Imdb), an accuracy of 90.68\% was reached. Top score on literature is
92.23\% and was reported by Johnson and Zhang in \cite{DBLP:journals/corr/Johnson014}. 
Same authors report a peak score of  97.36\% on Yelp reviews (top score reached here is 94.95) 
in \cite{DBLP:conf/acl/JohnsonZ17}. In both studies, they utilized deep and highly complex
neural network models.   
\begin{table}[ht] 
	\caption{Accuracies of top five network structures}  
	\centering    
	\setlength\tabcolsep{4.4pt}  
	\begin{tabular}
		{l | c c c c c}  
		\toprule
		\textbf{~~~~~~~~~~~Network} & \textbf{Mlpn} & \textbf{Sent} & \textbf{Imdb} & 
		\textbf{Phon} & \textbf{Yelp}	\\ [0.1ex]
		\midrule 
		Conv-Pool (R=4, L=2) & 72.24 & \textbf{79.89} & 87.98 & 95.31	& 92.32	  \\ [0.3ex]
		Conv-Pool (R=25, L=2) & \textbf{75.63} & 74.46 & 90.12 & 95.15 & 93.51	  \\ [0.3ex]
		Conv-Pool (R=16, L=4) & 73.34 & 75.08 & 89.87 & \textbf{96.57} & 94.86	  \\ [0.3ex]
		Conv-Pool (R=25, L=4) & 75.44 & 74.22 & \textbf{90.68} & 95.64 & 93.84  \\ [0.3ex]
		Conv-Pool (R=27, L=6) & 71.88 & 74.04 & 89.11 & 95.21 & \textbf{94.95}  \\ 
		\bottomrule
	\end{tabular} 
\label{table:mp8FiveNetworkScores}
\end{table}                                                                                                                         
\par 
It is important to note that the goal of the experiments presented in this chapter was not to 
obtain record-breaking results. The top-performing results we compared with were obtained 
from complex neural networks with millions of parameters. Contrary, the last model of 
Table~\ref{table:mp8FiveNetworkScores} (the most complex one) has fewer than 200 thousand
parameters. 
%
From that table, we can notice interesting patterns that can answer the three research 
questions posed at the end of Section~\ref{sec:DataPropIntro}. Regarding RQ1, we see that 
the three datasets of longer documents (Imdb, Mlpn, and Yelp), perform better on networks 
with bigger max-pooling regions. Their top scores are reached with aggregate downsampling 
coefficients $R = 5~x~5 = 25,~R = 25$ and $R = 3~x~3~x~3 = 27$ respectively. Contrary, 
Sent and Phon datasets that contain shorter texts reach their peak scores on networks with 
smaller values of $R$. In fact, $R$ is the parameter that dictates length $p$ of the final 
feature maps ($p = ceil(\frac{n - k + 1}{R})$).  
According to the results, highest accuracy scores are always reached with $p$ values 
that are within $7 - 15$. 
%
%
\par 
Regarding the role of filter size (RQ2), convolutions of $k = 1, k = 2 $ and $k = 3$ 
were essential for optimal scoring. Omitting one of them reduced accuracy. 
On the other hand, wider networks with convolutions of filters $k = 4$ 
or $k = 5$ did not perform any better. For this reason we stopped at $W = 3$.
%
Regarding RQ3 and network depth, we see that vertical expansion with $L = 4$ improves 
results for the two bigger datasets Phon and Imdb. They play better with the 
structures of four convolutional-pooling stacks. Moreover, Yelp which is the biggest 
dataset reaches optimal performance on a deeper network of six stacks of convolutional
and max-pooling layers. Aiming towards simplification and low training times, we did 
not try deeper networks ($L > 6$) for further improvements. Sent and Mlpn that are 
the smallest datasets, reach their peak scores on the simpler networks of two stacks only
($L = 2$). This is something intuitive since deeper networks are more data hungry and tend to 
overfit small datasets. Obviously, other layer combinations and network architectures 
should be explored to obtain top-notch results. In the next chapter, we direct our efforts 
precisely on this issue.  

\chapter{A Neural Network Architecture for Sentiment Analysis Prototyping}
\label{chapter7}
\ifpdf
    \graphicspath{{Chapter7/Figs/}{Chapter7/Figs/PDF/}{Chapter7/Figs/}}
\else
    \graphicspath{{Chapter7/Figs/Vector/}{Chapter7/Figs/}}
\fi
%
\begin{flushright}
	\emph{\large \dq{If you torture the data long enough, it will confess.}} \\
	-- Ronald Coase, Nobel Prize in economics 
\end{flushright}
\vskip 0.35in
\indent \indent
As described in the previous chapter, using neural networks for feature extraction, 
feature selection, and sentiment polarity prediction of texts produces very good results. 
However, finding the best network architecture and hyperparameter setup can be tedious 
because of the many design alternatives and hyperparameter values that need to be tuned. 
This is a general problem that has been faced by many researchers that use neural 
networks for various tasks. Image recognition and computer vision  
research communities have reacted by creating prepackaged deep neural  
architectures that are mostly based on convolutional layers. 
Those architectures are able to correctly classify objects from thousands of categories 
when trained with millions of object images. Considering that text datasets are quite 
smaller, it makes sense to tackle the complexity problem encapsulating network design 
parameters and hyperparameters in architectures of few layers. 
\par
This chapter presents and proposes NgramCNN, a shallow neural architecture for simplifying 
sentiment analysis model prototyping. Section~\ref{sec:NgramCnnIntro} presents 
similar studies successfully implementing same concepts for analyzing images. 
There are also studies that propose simple neural networks made up of convolution
and recursive network combinations. Basic architectural components such as 
word representation layer, as well as convolutional and max-pooling layers are described 
in details in Section~\ref{sec:NgramCnnArchs}. Section~\ref{sec:setupBaselines} 
shows text preprocessing steps and experiments conducted to evaluate the performance
of NgramCNN models. Finally, Section~\ref{sec:NgramCnnResults} further discusses
the obtained results.  
%
\section{Popular Neural Network Architectures} 
\label{sec:NgramCnnIntro}
%
Deep learning is today the buzzword technology for many tasks and domains like 
image recognition, machine translation, speech recognition or sentiment analysis. 
Prepackaged neural network architectures are very easy to use with little domain 
knowledge, producing top results. Furthermore, performance scales fairly well 
with increasing data availability and computation speed that have been 
growing steadily in the recent years. 
%
The basic structure of Convolutional Neural Networks (CNN) has been utilized
to form dozens of advanced architectures that are producing record-breaking 
results in the yearly ImageNet challenge \cite{Russakovsky:2015:ILS:2846547.2846559}. 
Images of thousands of categories are correctly identified via neural architectures 
like \emph{VGG-19}, 
\emph{AlexNet}, 
\emph{Inception} 
and more. These advanced image classification models have been included in libraries
that are freely available in Python or other programming languages. Furthermore, they
are continuously updated with various architectural optimizations introduced in the 
newer versions. 
%
A simple idea would be to utilize the above off-the-shelf complex networks on sentiment 
analysis tasks as well. However, this is hardly possible, given the high complexity of 
computer vision neural networks. They do usually have millions of trainable parameters
and are thus trained with datasets containing millions of images. Sentiment analysis 
datasets are usually smaller. 
\par 
As a result, similar but simpler architectures are proposed in literature studies for text 
analysis tasks as well. In \cite{kueflermerging} for example, author proposes a complex 
architecture called Language Inception Model, inspired by \emph{Inception} of Google. 
The basic structures of his architecture are convolution layers of different filter lengths 
that operate in parallel over text phrases to capture $n$-gram features. To preserve the value of 
word orders, he also applies a Long Short-Term Memory (LSTM) network in parallel with the 
Inception-like structure. The generated features from the two parallel structures are 
merged together and pushed to the fully connected layer that serves as the classifier. The author 
uses IMDB large movie review dataset and network structures like unigram, bigram 
or trigram RNNs to evaluate his architecture. He reports that the simpler bigram RNN 
performs slightly better than his bigger Inception-like architecture. 
%
Also in \cite{DBLP:conf/acl/JohnsonZ17}, authors propose a pyramid-shaped,
word-level architecture of convolutional layers that increases network depth while 
decreasing computation time per layer. The latter effect is achieved by keeping a fixed 
number of feature maps and applying $2$-stride downsampling from one layer to the 
next. The max-pooling layers appear only after certain blocks of consecutive convolutions.  
They report leading results in five of the large datasets created in
\cite{Zhang:2015:CCN:2969239.2969312}.  
\par 
In \cite{DBLP:conf/coling/SantosG14} we find a similar neural network architecture called 
CharSCNN that is designed to exploit information of characters, words and sentences 
at the same time. It is created to work particularly well with short and noisy texts 
such as tweets. Authors claim that character embeddings are able to capture important 
morphological and shape information of words. For this reason, each character is 
encoded in a fixed-length character embedding vector and an embedding matrix is 
created. To generate word-level embeddings, authors use Skip-Gram method
with windows size 9. Character-level and word-level embeddings are produced from
an English Wikipedia collection of 1.75 billion tokens and joined together. A convolution
layer is then applied to produce local features and a max-pooling layer finally creates the 
feature vectors of the entire sentence. CharSCNN is tested on Stanford Sentiment 
Treebank corpus of sentences and Stanford Twitter Sentiment corpus of tweets, achieving
state-of-the-art results in both.  
\par 
All above architectures differ in complexity, effectiveness, and number of parameters.
However, their fundamental logic is the same: using deep and complex neural network
constructs for extracting and selecting features in combination with one or few simple
feed-forward layers for classification.  In \cite{Razavian:2014:CFO:2679599.2679731}, 
authors support the same idea: generic features extracted from complex CNNs are very 
powerful and versatile. Their learned representations result high effective even in 
analogous tasks. 
%
Following the same logic, in this chapter, we probe architectures of CNNs 
that can package the complexity in feature extraction and selection layers and then use a 
single dense layer as a classifier. The goal is to find simple network configurations
that are fast to train and can be easily used as templates, yet providing competing results 
on sentiment analysis of medium-sized text datasets. We build on the results of the 
previous chapter, using the optimal convolution filter size and max-pooling region size 
parameters that were found. Same representation of text words was used as well. 
Performance of different architectures with hierarchical convolutional and max-pooling 
layer combinations are examined. The following sections describe their design, conducted 
experiments, and the corresponding results.

\section{Neural Architecture Design Alternatives}
\label{sec:NgramCnnArchs} 
\subsection{NgramCNN Basic Architecture}   
\begin{figure}[ht]
	\centering
	\includegraphics[width=0.70\textwidth]{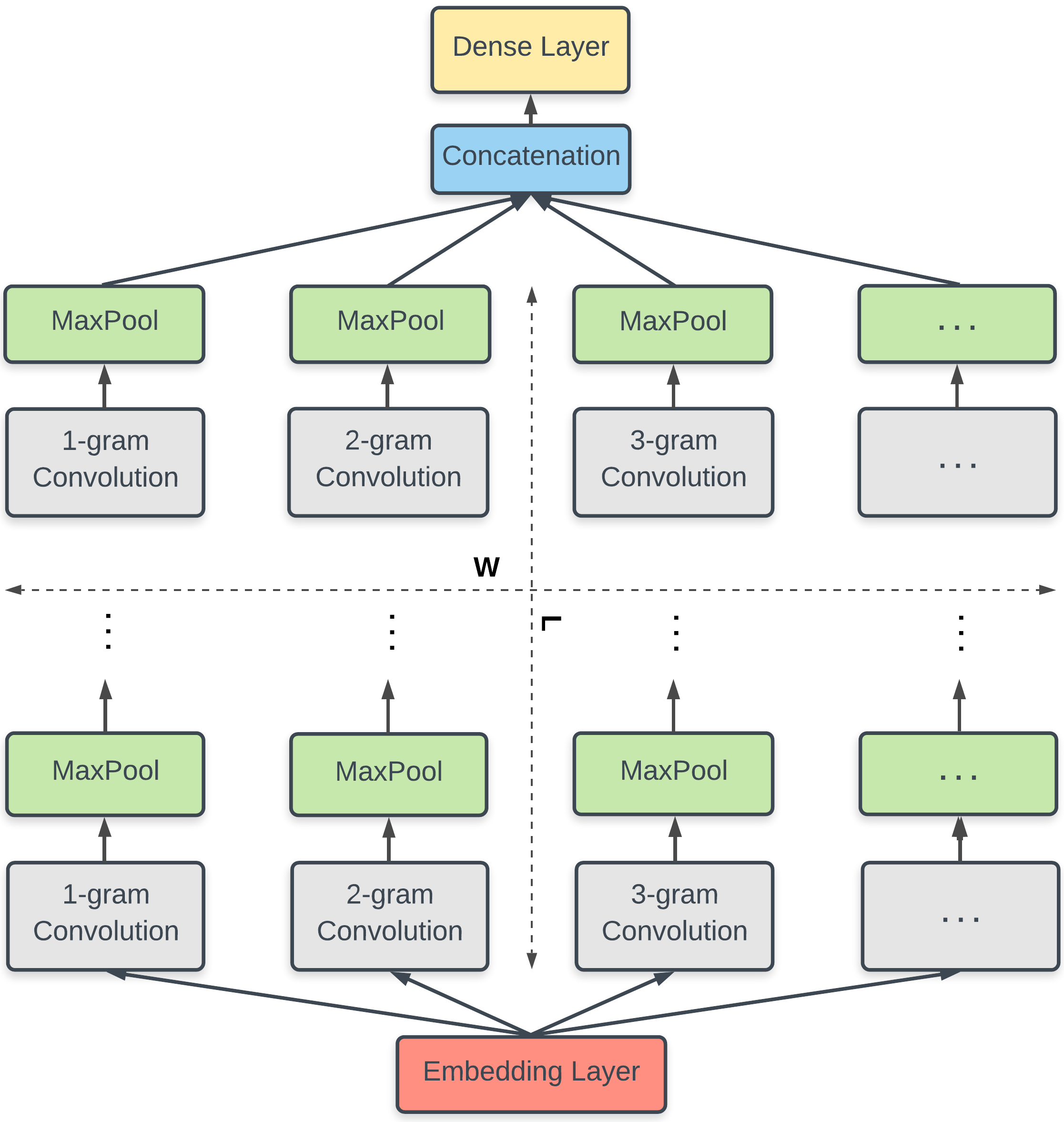}
	\caption{NgramCNN architecture scheme}
	\label{fig:NgramCnn1}   
\end{figure}

As shown in Figure~\ref{fig:NgramCnn1}, the basic architecture we 
designed (NgramCNN) is an extension and generalization of the neural 
networks explored in the previous section. It starts from the vector representations
of words which can be static (sourced from pretrained bundles) or trained on experimenting
dataset. Afterwards, it applies $W$ parallel convolutions of growing kernel lengths 
($k=1, \ldots, W$). These operations extract word and phrase combination 
features out of the text documents. Parameter $W$ (width) can be different, depending
on the task and data. Max-pooling layers that follow, select the maximal value from 
each feature map region of length $R$, downsampling the feature sets. The reduced 
features do usually contain the most salient samples. As shown in the previous chapter, 
the values of $R$ in the different max-pooling layers can be adjusted based on number 
of words ($n$) in the training documents. A series of similar parallel convolutional and 
max-pooling layers go on, forming a structure of $L$ stacks. Same as with $R$, the 
value of $L$ can be adapted to specific use cases and the available training data. Finally, 
output features of the last max-pooling layer are concatenated and pushed to the 
feed-forward classification layer. More than one classification layer can be used, 
though.   
\subsection{NgramCNN Pyramid Architecture}    
\begin{figure}[ht]
	\centering
	\includegraphics[width=0.70\textwidth]{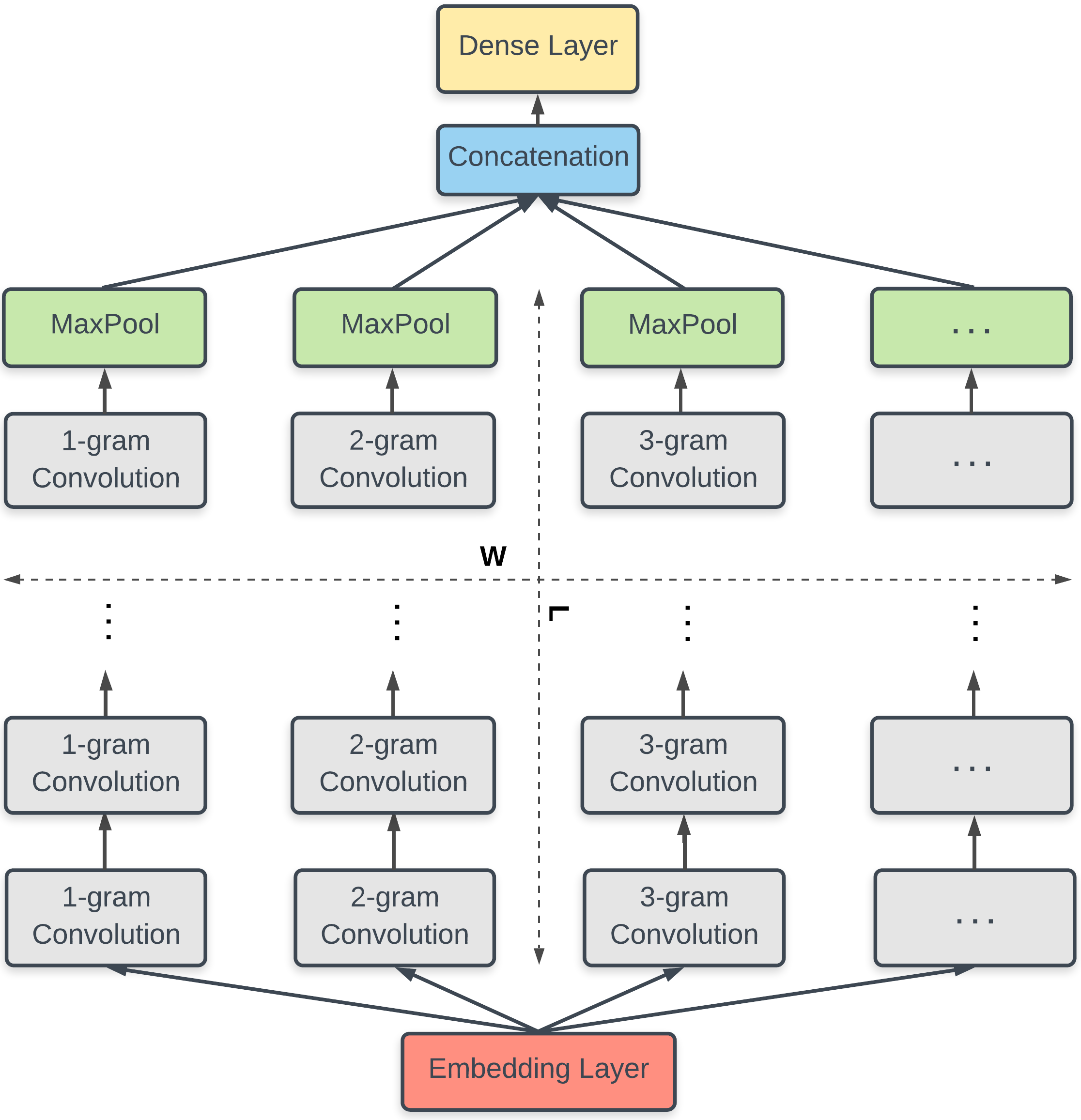}
	\caption{NgramCNN with downsampling convolutions}
	\label{fig:NgramCnn2}   
\end{figure}

A slightly different architecture is the one presented in Figure~\ref{fig:NgramCnn2}.  
Embedding layer, the first stack of convolutions, last stack of max-pooling layers and 
the classification layer are exactly the same. The difference is only in the downsampling 
stacks. Here we use striding convolutions with stride $s > 1$ instead of regional 
max-pooling with region length $R$ for the intermediate stacks. Max-pooling is still 
used in the final feature selection stack. If a fixed value of $s$ (e.g., $s = 2$) is used in 
all downsampling convolution layers, this architecture becomes pyramid-shaped and 
very similar with the one in \cite{DBLP:conf/acl/JohnsonZ17}. The feature set length 
is equally reduced in each stack of convolutions till the final max-pooling stack.  
%
When compared with the basic version of NgramCNN, there are a couple of differences 
to note between striding convolution layers and max-pooling layers. One obvious advantage
of striding convolutions is their ability to preserve positional information in feature sets. 
Max-pooling on the other hand, forgets everything about the spatial structure of the data, 
selecting features that are probably the most valuable for classification. There is still a problem 
with convolutions regarding training time. They are slower than max-pooling as they have 
parameters for updating weights. Training algorithms (e.g., \emph{Adam}) need to store 
information for the striding convolution but not for max-pooling. A study addressing this 
issue is \cite{DBLP:journals/corr/SpringenbergDBR14} where authors conduct several 
object-recognition experiments. They conclude that convolutions with increased 
stride can completely replace max-pooling layers, simplifying neural network structures 
without performance loss. Nevertheless, their results apply to image analysis only and 
do not count for other types of tasks like sentiment analysis, topic recognition etc.  
\subsection{NgramCNN Fluctuating Architecture}    
\begin{figure}
	\centering
	\includegraphics[width=0.70\textwidth]{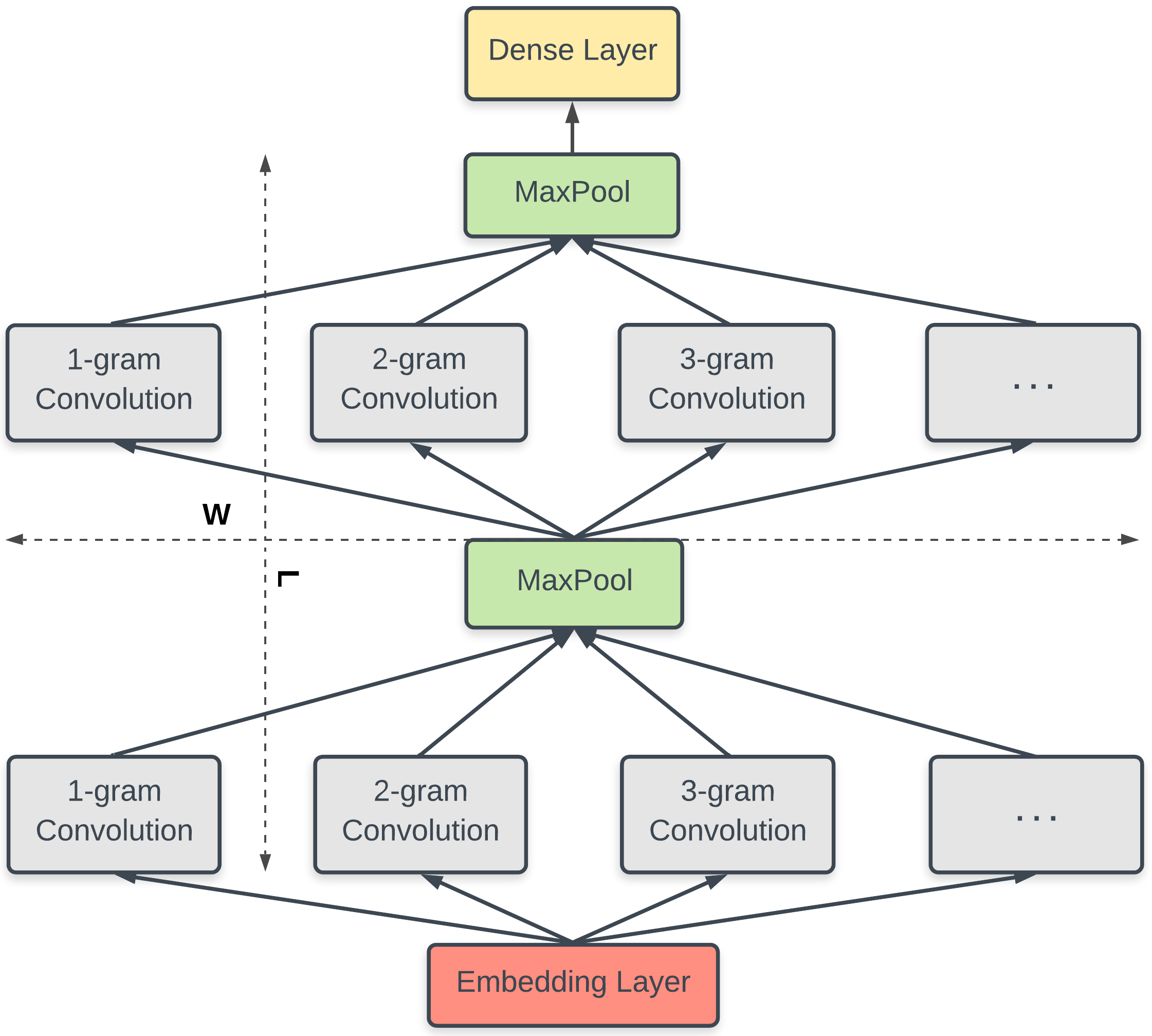}
	\caption{NgramCNN with fluctuating features}
	\label{fig:NgramCnn3}   
\end{figure}

Our third alternative architecture is shown in Figure~\ref{fig:NgramCnn3}. It is 
significantly different from the first two. The only common parts are embedding 
(first) and classification (last) layers. Feature selection and extracting layers are 
organized in a fluctuating form, with features expanding and contracting after each
convolutional and max-pooling stack of layers. 
The first stack of convolutions is same as in the other two architectures. $W$ layers
of convolutions with increasing filter sizes are performed in parallel. All generated feature 
maps are concatenated and pushed to a single max-pooling layer of region size $R$ which 
significantly reduces (contracts) their size. The second round of convolutions is applied to
the new features, expanding their size. The process goes on with another max-pooling 
layer (stack) for subsampling, repeating feature contraction and expansion in a total 
of $L$ stacks. Once again, $W, L,$ and $R$ parameters can be chosen to adapt to 
attributes of available training data.     
\section{Experimental Settings and Baselines}
\label{sec:setupBaselines}
In this section, we describe various network parameter decisions we made 
in the conducted experiments where we compare the above architectures. 
These experiments extend over those presented in \cite{icgda2018} (evaluation 
of basic NgramCNN) to consider the two alternative architectures as well.  
Other neural network models presented in similar text mining studies are 
also described and used as comparison baselines.  
\subsection{Common Network Settings}
\label{sec:NgCnnSettings}
%
A common setup of basic parameters was chosen for all three network 
architectures presented in the previous section. Based on the results in that chapter, 
we fixed $W$ to $3$, excluding convolution layers of longer 
filters from every network stack. Regarding network length (or depth), 
structures of four-layer stacks were best in most of the previous experiments.
As a result, a fixed $L = 4$ was chosen to keep the networks simple and 
quick to train.  
%
Regarding datasets, we reuse four of the five of those described in 
Section~\ref{sec:ExpDatasetsRoleNcnn}, excluding Mlpn only. This latter dataset 
has a small size, is overfitted by most of the models and has not been used in similar 
studies to compare with. Same data preprocessing steps as those of 
Section~\ref{sec:DataPrepSteps} were applied.     
%
Region sizes $R$ of max-pooling layers were also chosen based on 
experimental results of the previous chapter. Since hierarchical pooling 
is used, the product of region sizes in the two stacks must equal the optimal
value reported in Section~\ref{sec:CNNresultsDiscussion} for each dataset. 
Hence we chose $R = 2, ~ R = 5, ~ R = 4$ and $R = 5$ for Sent, Imdb, Phon and
Yelp datasets respectively. In the case of NgramCNN pyramid architecture, 
the above values correspond to convolution stride ($s$) of the second layer
as well.  
%
Regarding the representation of text documents, pretrained word embeddings 
sourced from GoogleNews bundle were used once again, same as in 
Section~\ref{sec:WordRepGoogNews}.    
\subsection{Baseline Models}   
\label{sec:NgramCnnBaselines}
Using the settings and parameters described above, we contrast the three NgramCNN 
architectures with each other as well as against similar neural network models 
proposed in relevant studies. One of the comparison models (SingleLSTM) is made up 
of a single LSTM layer positioned above the embedding layer. It is followed by the 
feed-forward classification layer. Another baseline model was presented in 
\cite{Lai:2015:RCN:2886521.2886636} (BLSTM-POOL). Here they use a bidirectional 
recurrent network (left and right LSTM) to capture word contexts. A max-pooling layer 
that follows is used for automatic selection of top features for classification. Authors report 
very good results on topic identification task and lower but still good results on sentiment analysis 
of movie reviews. A more complex baseline model was introduced in 
\cite{conf/coling/ZhouQZXBX16} (BLSTM-2DCNN). It builds upon the 
bidirectional structure of \cite{Lai:2015:RCN:2886521.2886636}, adding two-dimensional 
convolution and pooling layers applied on word-feature windows. The model is exercised
on various datasets and tasks and achieves top scores in sentiment analysis of short sentences.
\par 
We also compared against a simple model made up of few convolution layers in a single
stack (SingleCNN). This model is quite similar to that of Kim in 
\cite{DBLP:journals/corr/Kim14f}. Here, convolution layers of filter size 3, 4, and 5 
operate on word embeddings and the feature maps they produce are concatenated together.
Max-pooling and a dropout layer for regularization follow the convolutions and a dense layer
is used for classification. Considering the diversity of the datasets and tasks we experiment
on, it is worth comparing against simpler and more traditional classification models that 
work with bag-of-words text representation and check their performance. As a result, 
we also implemented Support Vector Machine and Logistic Regression classifiers
using tf-idf feature vectorizer. They were optimized with grid-searched regularization 
parameters.  
%
\section{Results and Discussion}
\label{sec:NgramCnnResults}
This section presents and discusses classification performance scores obtained from
the datasets. The three variants of NgramCNN are compared with each other and the 
rest of baseline models. The sensitivity of results with respect to various network and 
training parameters is also discussed. 
\subsection{Sentiment Polarity Classification Results}
%
Each experiment was run on 70/10/20 percent splits for training, development, and testing
respectively. Table~\ref{table:NgCnnResults} summarizes classification accuracy scores
of the models on each of the four datasets. On sentiment polarity analysis (smallest dataset
with shortest documents), NgramCNN models perform relatively badly. Optimized Logistic 
Regression and Support Vector Machine, together with SingleCNN network achieve very 
good results whereas LSTM-based models (SingleLSTM and BLSTM-POOL) perform 
worse. Best score of 82.32\% was reached from BLSTM-2DCNN presented in 
\cite{conf/coling/ZhouQZXBX16} which integrates both recurrent and convolution 
networks. This result is slightly lower from the very top score of 83.1\% 
reported in \cite{DBLP:journals/corr/ZhaoLP15}.  
%
On movie reviews dataset, we see a different picture. The first two versions of NgramCNN
score above 91\%. Top accuracy score on this dataset is 92.23\%, as reported in 
\cite{DBLP:journals/corr/Johnson014}. Fluctuating architecture is well behind, together 
with the two linear models and SingleCNN. LSTM-based models, on the other hand, 
achieve very poor results that are below 86\%. Obviously, recurrent neural networks do 
not work well on text analysis tasks of long documents. 
\par 
On smartphone reviews, we see similar results, with NgramCNN basic architecture peaking
at 95.92\%. Unfortunately, we did not find a literature score report on this dataset to compare 
with. The Pyramid architecture is very close whereas the Fluctuating version is again 
well behind. We also see that LSTM models perform relatively well (BLSTM-2DCNN),
at least on short documents. Linear models, on the other hand, perform weakly.
%
Yelp business reviews is the last dataset we experimented on. Once again, best results are
achieved from NgramCNN Basic which is again followed by the Pyramid version. 
Top literature score on this dataset is  97.36\% (as reported in \cite{DBLP:conf/acl/JohnsonZ17})
which is 2.5\% higher than that of NgramCNN Basic. SingleCNN is also a good player,
with an accuracy of 93.86\%. The rest of the models, including Fluctuating architecture, are 
behind of more than 2\%.   
%
\begin{table}[ht] 
	\caption{Accuracy scores of NgramCNN variants and baselines}  
	\centering    
	\setlength\tabcolsep{4.7pt}  
	\begin{tabular}
		{l | c c c c}  
		\toprule
\textbf{~~~~~~Network} & \textbf{Sent} & \textbf{Imdb} & \textbf{Phon} & \textbf{Yelp}	\\ [0.1ex] 
\midrule 
NgCNN Basic & 79.87 & 91.14 & \textbf{95.92}	& \textbf{94.95}	  \\ [0.3ex]
NgCNN Pyramid & 79.52 & \textbf{91.21} & 95.7 & 94.83		  \\ [0.3ex]
NgCNN Fluctuate & 77.41 & 89.32 & 93.45 & 92.27		  \\ [0.3ex]
Optimized LR & 81.63 & 89.48 & 92.46 & 91.75  \\ [0.3ex]
Optimized SVM & 82.06 & 88.53 & 92.67 & 92.36  \\ [0.3ex]
SingleLSTM & 80.33 & 84.93 & 93.71 & 90.22	  \\  [0.3ex]
SingleCNN & 81.79 & 89.84 & 94.25 & 93.86	  \\  [0.3ex]
BLSTM-POOL & 80.96 & 85.54 & 94.33 & 91.19  \\ [0.3ex]
BLSTM-2DCNN & \textbf{82.32} & 85.70 & 95.52 & 91.48  \\
\bottomrule
	\end{tabular} 
\label{table:NgCnnResults}
\end{table}
\subsection{Further Observations}
%
Based on the results of Table~\ref{table:NgCnnResults}, we see that basic version of 
NgramCNN is the best of the three design alternatives. Pyramid version that uses striding 
convolutions in the intermediate stacks, is quite similar
in design and slightly worse in performance. To further assess the role of striding 
convolutions and max-pooling layers, we explored another version which uses no 
max-pooling layers at all. This version performed even worse with accuracy scores 
lower than 2\%. Apparently, regional max-pooling does a good job as a feature 
selector and is also faster to train. Fluctuating version, on the other hand, scored
badly on every dataset. It is obviously not a good network design choice. 
%
The most intuitive explanation for this is the way it mixes up the features of the 
different convolution filters. Treating them separately as in the first two architectures 
yields better classification results. 
%
It also comes out that models based on recurrent neural networks perform well on 
short documents but considerably worse on longer ones. This has to do with their
internal design. Their memorization abilities are limited and they cannot  
preserve long-term word dependencies. 
%
Furthermore, these kinds of networks usually take longer to train. Due to their simpler
design, models based on CNNs were much faster, with training times comparable 
to those of the two linear models. 
%
These later performed comparably well on the smaller datasets in which they might be 
acceptable choices in cases when performance is not crucial. 
\par 
Regarding network hyperparameters, the most important design decision such as
depth and width of network or size of max-pooling regions were chosen based on results
of the previous chapter, as explained in Section~\ref{sec:NgCnnSettings}. It is important to note
that NgramCNN architecture is highly flexible. Different design choices such as more 
convolutional and max-pooling stacks or longer filters can be explored for even higher 
performance if other datasets are available. 
Remaining parameters were chosen based on grid search results. The number of epochs until 
convergence was irregular and specific on each dataset. Sentences converged in three epochs,
whereas smartphone and movie reviews required seven and four epochs respectively. Yelp 
dataset that was the biggest required nine epochs to converge. There was not much sensitivity
with respect to batch size. A batch of 60 was usually optimal in most experiments. To conclude, 
\emph{Softplus} and \emph{Sigmoid} were equally good as activation functions for the 
output layer. 
\subsection{Concluding Remarks}
%
In this chapter, we presented a couple of neural network architectures for sentiment 
analysis that are based on convolutional layers applied on top of pretrained word 
embedding representations of texts. They follow the same fundamental design of the 
top performing architectures on image recognition tasks: complexity in feature extraction 
and selection combined with simplicity in classification. Based on obtained results, this 
design appears successful in text analysis as well. 
%
The basic design of convolution and max-pooling 
stacks of layers resulted the best performing one. Even though it is not a record-breaking 
architecture in terms of classification accuracy, it is yet very fast and highly adaptable to 
various kinds of datasets. Many versions that differ in width, depth and pooling region
sizes can be easily constructed and potentially produce state-of-the-art results if properly 
tuned. We thus believe that it can be very useful as a prototyping or baseline sentiment 
analysis model, especially for practitioners.  
%
One limitation of our experiments is the use of static word vectors for text 
representations. If bigger text datasets are available, tuning word vectors on training texts
would require more computation time but probably produce slightly better features. Another 
improvement of NgramCNN could be to employ sentiment analysis on different splits (phrases) 
of text documents and predict their emotional polarity independently. The overall polarity of 
the document could be assessed utilizing aggregation schemes like Dempster-Shafer 
Inference or Abductive Reasoning \cite{citeulike:13327098}.

\chapter{Conclusions and Prospects}
\label{chapter8}
\ifpdf
    \graphicspath{{Chapter8/Figs/}{Chapter8/Figs/PDF/}{Chapter8/Figs/}}
\else
    \graphicspath{{Chapter8/Figs/Vector/}{Chapter8/Figs/}}
\fi
%
%
\indent \indent
%
This thesis addressed different aspects regarding sentiment analysis of songs and various 
types of texts. The main concerns were the need for viable methods to obtain 
bigger labeled datasets, the adequacy of distributed word representations for sentiment
analysis of texts and the complexity reduction of neural network models for 
easier and faster prototyping. Three main goals were set out: \emph{(i)} exploring 
the viability of crowdsourcing methods like crawled social tags for constructing
datasets of song emotions, \emph{(ii)} probing the role of factors like training
corpus size or generation method in the quality of word embeddings that are 
produced, \emph{(iii)} designing a neural network architecture made up of 
convolution and max-pooling layers for encapsulating complexity and simplifying 
sentiment analysis model creation. 
%
In short, the main contributions of this thesis are: 
\begin{itemize}
\item The validity proof of crowdsourcing methods as alternatives for creating labeled 
data collection by comparing emotional spaces of social tags with those of psychologists, 
presented in Chapter~\ref{chapter3}.
\item The creation of two datasets of song emotions using \emph{Last.fm} social tags
and the exploration of lexicons as possible means for text data programming,
also introduced in Chapter~\ref{chapter3}. 
\item A systematic literature survey about hybrid and context-based recommender 
systems, the results of which are summarized in Chapter~\ref{chapter4}. 
\item The design of an in-car music mood recommender system in collaboration 
with Telecom Italia JOL Mobilab, described in Chapter~\ref{chapter4}.
\item The empirical identification of interesting relations between text size, training method 
or text topic and the generated word vectors (Chapter~\ref{chapter5}).
\item A set of patterns relating neural network hyperparameters with the size of datasets 
and length of documents, with respect to optimization of text sentiment prediction
revealed in Chapter~\ref{chapter6}. 
\item A neural network architecture made up of convolutional and max-pooling layers 
that encapsulates most of hyperparameter tuning complexity, simplifying the creation 
of models for sentiment analysis of texts (Chapter~\ref{chapter7}).
\end{itemize}
%
Literature observations, as well as conducted experiments, show that social tags
and other forms of crowdsourcing are viable for collecting labels 
of data if correctly applied. Furthermore, the excellent compliance between user tag 
emotion spaces and those of psychological emotion models, makes us conclude 
that massive crowd tags are a trustworthy source of intelligence. In this context,
the creation of MoodyLyrics4Q and MoodyLyricsPN was based on \emph{Last.fm}
affect tags. A variant of Russell's model with tags of four categories 
was used for representing emotion classes. MoodyLyrics4Q is the first music dataset that 
is both large and polarized, follows a popular emotion model, and is released for public 
use. 
\par 
The systematic survey on hybrid and context-based recommender systems revealed current 
tendencies such as the growing interest in the field, the popularity of K-nearest neighbors
and K-means algorithms, the preference of weighted filtering technique combinations
over the more complex ones, frequent applications on movies, and the use of accuracy 
for evaluating the quality of recommendations. Regarding context-based recommenders, 
those implementing input data pre-filtering are the most simple and common. 
The recommender of songs described in Section~\ref{sec:MoodCarRS} was designed 
to work in the context of car driving, using the intelligence of collected user tags about songs.  
The goal was to enhance comfort and safety of driving experience by adjusting music 
emotions to contextual factors like driving patterns and emotional state of the driver, 
time of day, etc. 
\par
The experimental work with word embeddings revealed relations between training 
data and methods with the quality of the generated word vectors. Among the most 
important insights, we can mention the superiority of Glove over Skip-Gram on 
analysis of song lyrics and the direct relation between corpus size and vocabulary richness
with the performance of word vectors. Another important conclusion drawn from 
those experiments is the fact that for classification based on small training datasets, 
it is better to source pretrained word vectors created from big text corpora.  
\par 
Numerous experiments with convolutional and max-pooling neural networks, trained with 
datasets of different sizes and variable-length text documents, revealed certain patterns 
relating those data properties with network hyperparameters, the number of stacked layers or 
pooling size and the optimal classification performance. Among the most important 
tendencies that were observed, those relating document length with the size of generated 
feature maps ($7 -15$) and the number of layers with dataset size were particularly valuable. 
The experiments that were conducted covered text mining tasks such as sentiment polarity 
prediction of song lyrics, movie reviews, and product reviews.  
\par 
The results of those experiments were used to address neural network 
complexity problem by designing an architecture of convolutional and regional 
max-pooling layers. 
Its role is to encapsulate feature extraction and selection complexity by having 
most of the hyperparameters preset and adjusting the few remaining ones in accordance
with training data properties. Using NgramCNN architecture as a starting point, 
one can easily create models that differ in width, depth and pooling region
sizes and can possibly generate competitive results.
This way, model creation on tasks such as sentiment analysis
or text mining in general, becomes easier for practitioners. The architecture 
can be applied to various dataset sizes and document lengths with little parameter 
tunning and provides state-of-the-art performance in text polarity prediction. 
\par 
The democratization of deep learning and the trend towards data-driven intelligent 
solutions requires even more and bigger public datasets. The most popular benchmarking
datasets that are currently used for sentiment analysis such as IMDB Movie 
Reviews\footnote{\url{http://ai.stanford.edu/~amaas/data/sentiment}} or Cornell Sentence
Polarity\footnote{\url{http://www.cs.cornell.edu/people/pabo/movie-review-data}}
are limited in text polarity (\emph{negative} vs. \emph{positive}) recognition. Practical systems do 
often require datasets with hundred thousands of documents 
containing \emph{neutral} text category as well. It is thus important to crawl 
opinion texts from websites and build bigger and better experimentation datasets.
The high cost of labeling large amounts of documents can be significantly 
reduced utilizing crowdsourcing or data programming. The latter was
not very successful with song lyrics (Section~\ref{sec:LableGenerationLexicons})
when employing affect terms of lexicons for emotion category discrimination. 
Nevertheless, more advanced heuristics could solve this problem, especially for
shorter documents such as product reviews. 
\par 
With enough data in possession, other promising ideas involving deep neural networks 
can be explored. For example, automatic features of convolutional or other types
of neural networks are not always optimal. Ensemble methods of different levels
might produce improved features and results when applied in the context of deep neural 
networks. As a result, one possibility could be to aggregate multiple types of local
and distributed word features (feature-level fusing). Different feature types 
do usually represent complementary data characteristics. Each of them reflects a 
different view of the original data. Sentiment analysis systems are thus expected to 
benefit from feature aggregations.  
\par 
Another problem is the information loss that happens in the consecutive convolution
and pooling operations. Recurrent neural networks are immune to this problem but 
do not work well with longer texts due to their inability to memorize long word 
sequences. A possibility could be to use recurrent-convolutional networks 
(proved to work optimally in \cite{7942788}, \cite{conf/coling/ZhouQZXBX16} and more) 
on short phrases of the longer document and then perform decision-level fusion for 
the overall prediction. Various aggregation schemes like Dempster-Shafer Inference 
or Abductive Reasoning can be explored in the decision step \cite{citeulike:13327098}.  
\begin{spacing}{0.9}


\bibliographystyle{unsrt} 
\cleardoublepage
\bibliography{References/references} 



\end{spacing}


\begin{appendices} 

\chapter{Systematic Literature Review Tables}  
\label{sec:Appendix1}
%
%
\begin{table}[h!]  
	\caption{Digital libraries inquired for primary studies} 
	\small
	\centering     
	\begin{tabular}
		{l l}   
		\topline
		\headcol \textbf{Source} & \textbf{URL}  \\ 
		\midline
		SpringerLink & http://link.springer.com   \\    
		\rowcol Science Direct & http://www.sciencedirect.com   \\ 
		IEEExplore & http://ieeexplore.ieee.org         \\ 
		\rowcol ACM Digital Library & http://dl.acm.org         \\ 
		Scopus & http://www.scopus.com                  \\ 
		\bottomline
	\end{tabular}  
\label{table:DigitalLibraries}
\end{table} 
%
%
\begin{table}[h!]
	\caption{Inclusion and exclusion criteria for filtering papers}
	\small 
	\centering      
	\begin{tabular}
		{l}
		\topline
		\headcol \textbf{Inclusion criteria}  		\\ [0.5ex] %
		\midline
		Papers addressing hybrid recommender systems, algorithms, strategies, etc. 			\\ [0.8ex]
		\rowcol Papers that do not directly address hybrid RSs, but describe RS engines that generate \\
		\rowcol recommendations combining different data mining techniques. 		\\ [0.8ex] 		
		Papers from conferences and journals       					 	 					\\ [0.8ex]
		\rowcol Papers published from 2005 to date             		 					\\ [0.8ex]
		Papers written in English only                 		 					\\ [0.8ex]
		\hline
		\headcol \textbf{Exclusion criteria}  		\\
		\hline 
		Papers that do not address RSs in any way 					 				\\ [0.8ex]
		\rowcol Papers addressing RSs but not proposing or implementing any hybrid combination		\\		
		\rowcol of different strategies or intelligent techniques.		 			 		\\ [0.8ex]	   
		Papers reporting only abstracts or illustration slides that miss detailed information \\ [0.8ex]
		\rowcol Grey literature                                 		   					\\ 
		\bottomline
	\end{tabular}   
\label{table:InclusionExclusion} 
\end{table}
\newpage
%
%
\newcolumntype{L}[1]{>{\raggedright\let\newline\\\arraybackslash\hspace{0pt}}p{#1}}
\begingroup
\fontsize{6pt}{7pt}\selectfont
\begin{longtable}{l L{2cm} l L{3.8cm} L{0.7cm} L{3.8cm}}
	\caption{Final list of selected papers} \\
	\topline
	\headcol P & Authors & Year & Title & Source & Publication details \\
	\midline
	\endfirsthead
	\multicolumn{6}{l}
	{\tablename\ \thetable\ -- \textit{Continued from previous page}} \\
	\hline
	\headcol P & Authors & Year & Title & Source & Publication details \\
	\midline
	\endhead
	\hline \multicolumn{6}{l}{\textit{Continued on next page}} \\
	\endfoot
	\hline
	\endlastfoot
	\hypertarget{P1}{P1} & Wang, J.; \newline De Vries, P. A.; Reinders, J. T. M.; & 2006 & Unifying User-based and Item-based Collaborative Filtering Approaches by Similarity Fusion  & ACM & 29th Annual International ACM SIGIR Conference on Research \& Development on Information Retrieval, Seattle 2006 \\ [1ex]
	
	\rowcol \hypertarget{P2}{P2} & Gunawardana, A.; Meek, C.; & 2008 & Tied Boltzmann Machines for Cold Start Recommendations & ACM 
	& 2nd ACM Conference on Recommender Systems, Lousanne, Switzerland, 23rd-25th October 2008 \\ [1ex]
	
	P3 & Gunawardana, A.; Meek, C.; & 2009 & A Unified Approach to Building Hybrid Recommender Systems & ACM 
	& 3rd ACM Conference on Recommender Systems, New York, October 23-25, 2009 \\ [1ex]
	
	\rowcol P4 & Park, S. T.; Chu, W.; & 2009 & Pairwise Preference Regression for Cold-start Recommendation & ACM & 3rd ACM Conference on Recommender Systems, 
	New York, October 23-25, 2009  \\ [1ex]
	
	\hypertarget{P5}{P5} & Ghazanfar, M. A.; Prugel-Bennett, A.; & 2010 & An Improved Switching Hybrid Recommender System Using Naive Bayes Classifier and 
	Collaborative Filtering & ACM & Proceedings of the International MultiConference of Engineers and Computer Scientists 2010, Vol I, Hong Kong, 
	March 17-19, 2010 \\ [1ex]
	
	\rowcol \hypertarget{P6}{P6} & Zhuhadar, L.; Nasraoui, O.; & 2010 & An Improved Switching Hybrid Recommender System Using Naive Bayes Classifier and 
	Collaborative Filtering & ACM & Proceedings of the International MultiConference of Engineers and Computer Scientists 2010, Vol I, Hong Kong, 
	March 17-19, 2010 \\[1ex]
	
	\hypertarget{P7}{P7} & Hwang, C. S.; & 2010 & Genetic Algorithms for Feature Weighting in Multi-criteria Recommender Systems & ACM & Journal of Convergence 
	Information Technology, Vol. 5, N. 8, October 2010 \\[1ex]
	
	\rowcol \hypertarget{P8}{P8} & Liu, L.; Mehandjiev, N.; Xu, D. L.; & 2011 & Multi-Criteria Service Recommendation Based on User Criteria Preferences
	& ACM & 5th ACM Conference on Recommender Systems, Chicago, Oct 23rd-27th 2011 \\[1ex]
	
	P9 & Bostandjiev, S.; O’Donovan, J.; Höllerer, T.; & 2012 & TasteWeights: A Visual Interactive Hybrid Recommender System & ACM & 6th ACM Conference on Recommender Systems, Dublin, Sep. 9th-13th, 2012 \\[1ex]
	
	\rowcol \hypertarget{P10}{P10} & Stanescu, A.; Nagar, S.; Caragea, D.; & 2013 & A Hybrid Recommender System: User Profiling from Keywords and Ratings
	& ACM & A Hybrid Recommender System: User Profiling from Keywords and Ratings \\ [1ex]
	
	\hypertarget{P11}{P11} & Hornung, T.; Ziegler, C. N.; Franz, S.; & 2013 & Evaluating Hybrid Music Recommender Systems & ACM & 2013 IEEE/WIC/ACM International 
	Conferences on Web Intelligence (WI) and Intelligent Agent Technology (IAT) \\ [1ex]
	
	\rowcol \hypertarget{P12}{P12} & Said, A.; Fields, B.; Jain, B. J.; & 2013 & User-Centric Evaluation of a K-Furthest Neighbor Collaborative Filtering 
	Recommender Algorithm & ACM & The 16th ACM Conference on Computer Supported Cooperative Work and Social Computing, Texas, Feb. 2013 \\ [1ex]
	
	\hypertarget{P13}{P13} & Hu, L.; Cao, J.; Xu, G.; Cao, L.; Gu, Z.; Zhu, C.; & 2013 & Personalized Recommendation via Cross-Domain Triadic 
	Factorization & Scopus & 22nd ACM International WWW Conference, May 2013, Brasil \\ [1ex]
	
	\rowcol \hypertarget{P14}{P14} & Christensen, I.; Schiaffino, S.; & 2014 & A Hybrid Approach for Group Profiling in Recommender Systems 
	& ACM & Journal of Universal Computer Science, vol. 20, no. 4, 2014 \\ [1ex]
	
	P15 & Garden, M.; Dudek, G.; & 2005 & Semantic feedback for hybrid recommendations in Recommendz & IEEE & IEEE 2005 International 
	Conference on e-Technology, e-Commerce and e-Service \\ [1ex]
	
	\rowcol P16 & Bezerra, B. L. D.; Carvalho, F. T.; Filho, V. M.; & 2006 & C2 :: A Collaborative Recommendation System Based on Modal 
	Symbolic User Profile & IEEE & Proceedings of the 2006 IEEE/WIC/ACM International Conference on Web Intelligence \\ [1ex]
	
	P17 & Ren, L.; He, L.; Gu, J.; Xia, W.; Wu, F.; & 2008 & A Hybrid Recommender Approach Based on Widrow-Hoff Learning & IEEE & IEEE 2008 Second International Conference on Future Generation Communication and Networking \\ [1ex]
	
	\rowcol P18 & Godoy, D.; Amandi, A.; & 2008 & Hybrid Content and Tag-based Profiles for Recommendation in Collaborative Tagging 
	Systems & IEEE & IEEE 2008 Latin American Web Conference \\ [1ex]
	
	P19 & Aimeur, E.; Brassard, G.; Fernandez, J. M.; Onana, F. S. M.; Rakowski, Z.; & 2008 & Experimental Demonstration of 
	a Hybrid Privacy-Preserving Recommender System & IEEE & The Third International Conference on Availability, Reliability and Security, 
	IEEE 2008 \\ [1ex]
	
	\rowcol \hypertarget{P20}{P20} & Yoshii, K.; Goto, M.; Komatani, K.; Ogata, T.; Okuno, H. G.; & 2008 & An Efficient Hybrid Music Recommender System Using 
	an Incrementally Trainable Probabilistic Generative Model & IEEE & IEEE TRANSACTIONS ON AUDIO, SPEECH, AND LANGUAGE PROCESSING, VOL. 
	16, NO. 2, FEBRUARY 2008 \\ [1ex]
	
	\hypertarget{P21}{P21} & Maneeroj, S.; Takasu, A.; & 2009 & Hybrid Recommender System Using Latent Features & IEEE & IEEE 2009 International Conference 
	on Advanced Information Networking and Applications \\ [1ex]
	
	\rowcol P22 & Meller, T.; Wang, E.; Lin, F.; Yang, C.; & 2009 & New Classification Algorithms for Developing Online Program
	Recommendation Systems & IEEE & IEEE 2009 International Conference on Mobile, Hybrid, and On-line Learning \\ [1ex]
	
	\hypertarget{P23}{P23} & Shambour, Q.; Lu, J.; & 2010 & A Framework of Hybrid Recommendation System for Government-to-Business Personalized 
	e-Services & IEEE & IEEE 2010 Seventh International Conference on Information Technology \\ [1ex]
	
	\rowcol \hypertarget{P24}{P24} & Deng, Y.; Wu, Z.; Tang, C.; Si, H.; Xiong, H.; Chen, Z.; & 2010 & A Hybrid Movie Recommender Based on Ontology and 
	Neural Networks & IEEE & A Hybrid Movie Recommender Based on Ontology and Neural Networks \\ [1ex]
	
	\hypertarget{P25}{P25} & Yang, S. Y.; Hsu, C. L.; & 2010 & A New Ontology-Supported and Hybrid  Recommending Information System for Scholars & Scopus & 13th International Conference on Network-Based Information Systems \\ [1ex]
	
	\rowcol \hypertarget{P26}{P26} & Basiri, J.; Shakery, A.; Moshiri, B.; Hayat, M.; & 2010 & Alleviating the Cold-Start Problem of Recommender Systems 
	Using a New Hybrid Approach & IEEE & IEEE 2010 5th International Symposium on Telecommunications (IST'2010) \\ [1ex]
	
	\hypertarget{P27}{P27} & Valdez, M. G.; Alanis, A.; Parra, B.; & 2010 & Fuzzy Inference for Learning Object Recommendation & IEEE & IEEE 2010 
	International Conference on Fuzzy Systems \\ [1ex]
	
	\rowcol \hypertarget{P28}{P28} & Choi, S. H.; Jeong, Y. S.; Jeong, M. K.; & 2010 & A Hybrid Recommendation Method with Reduced Data for Large-Scale 
	Application & IEEE & IEEE Transactions on systems, man and cybernetics - Part C: Applicatios and Reviews, VOL. 40, NO. 5, 
	September 2010 \\ [1ex]
	
	\hypertarget{P29}{P29} & Ghazanfar, M. A.;  Prugel-Bennett, A.; & 2010 & Building Switching Hybrid Recommender System Using Machine Learning 
	Classifiers and Collaborative Filtering & IEEE & IEEE IAENG International Journal of Computer Science, 37:3, IJCS\_37\_3\_09 \\ [1ex]
	
	\rowcol P30 & Castro-Herrera, C.; & 2010 & A Hybrid Recommender System for Finding Relevant Users in Open Source Forums & Scopus & IEEE 3rd International Conference on Managing Requirements Knowledge, 
	Sept. 2010 \\ [1ex]
	
	P31 & Tath, I.; Biturk, A.; & 2011 & A Tag-based Hybrid Music Recommendation System Using Semantic Relations and Multi-domain 
	Information & IEEE & 11th IEEE International Conference on Data Mining Workshops, Dec. 2011 \\ [1ex]
	
	\rowcol P32 & Kohi, A.; Ebrahimi, S. J.; Jalali, M.; & 2011 & Improving the Accuracy and Efficiency of Tag Recommendation System by
	Applying Hybrid Methods & IEEE & IEEE 1st International eConference on Computer and Knowledge Engineering (ICCKE), October 13-14, 
	2011 \\ [1ex]
	
	P33 & Kohi, A.; Ebrahimi, S. J.; Jalali, M.; & 2011 & Improving the Accuracy and Efficiency of Tag Recommendation System by
	Applying Hybrid Methods & IEEE & IEEE 1st International eConference on Computer and Knowledge Engineering (ICCKE), October 13-14, 
	2011 \\ [1ex]
	
	\rowcol \hypertarget{P34}{P34} & Fenza, G.; Fischetti, E.; Furno, D.; Loia, V.; & 2011 & A hybrid context aware system for tourist guidance based on collaborative filtering & Scopus & 2011 IEEE International Conference on Fuzzy Systems, 
	June 27-30, 2011, Taipei, Taiwan \\ [1ex]
	
	P35 & Shambour, Q.; Lu, J.; & 2011 & A Hybrid Multi-Criteria Semantic-enhanced Collaborative Filtering Approach for 
	Personalized Recommendations & IEEE & 2011 IEEE/WIC/ACM International Conferences on Web Intelligence and Intelligent Agent 
	Technology \\ [1ex]
	
	\rowcol \hypertarget{P36}{P36} & Li, X.; Murata, T.; & 2012 & Multidimensional Clustering Based Collaborative Filtering Approach for Diversified 
	Recommendation & IEEE & The 7th International Conference on Computer Science \& Education July 14-17, 2012. Melbourne, 
	Australia \\ [1ex]
	
	\hypertarget{P37}{P37} & Shahriyary, S.; Aghabab, M. P.; & 2013 & Recommender systems on web service selection problems using a new hybrid 
	approach & IEEE & IEEE 4th International Conference on Computer and Knowledge Engineering, 2014 \\ [1ex]
	
	\rowcol \hypertarget{P38}{P38} & Yu, C. C.; Yamaguchi, T.; Takama, Y.; & 2013 & A Hybrid Recommender System based Non-common Items in Social Media & IEEE & IEEE International Joint Conference on Awareness Science and Technology and Ubi-Media Computing, 2013 \\ [1ex]
	
	P39 & Buncle, J.; Anane, R.; Nakayama, M.; & 2013 & A Recommendation Cascade for e-learning & IEEE & 2013 IEEE 27th International 
	Conference on Advanced Information Networking and Applications \\ [1ex]
	
	\rowcol \hypertarget{P40}{P40} & Bedi, P.; Vashisth, P.; Khurana, P.; & 2013 & Modeling User Preferences in a Hybrid Recommender System using Type-2 Fuzzy Sets & Scopus & IEEE International Conference on Fuzzy Systems, July 2013 \\ [1ex]
	
	\hypertarget{P41}{P41} & Andrade, M. T.; Almeida, F.; & 2013 & Novel Hybrid Approach to Content Recommendation based on Predicted 
	Profiles & IEEE & 2013 IEEE 10th International Conference on Ubiquitous Intelligence \& Computing \\ [1ex]
	
	\rowcol P42 & Yao, L.; Sheng, Q. Z.; Segev, A.; Yu, J.; & 2013 & Recommending Web Services via Combining Collaborative Filtering 
	with Content-based Features & IEEE & 2013 IEEE 20th International Conference on Web Services \\ [1ex]
	
	P43 & Luo, Y.; Xu, B.; Cai, H.; Bu, F.; & 2014 & A Hybrid User Profile Model for Personalized Recommender System with Linked 
	Open Data & IEEE & IEEE 2014 Second International Conference on Enterprise Systems \\ [1ex]
	
	\rowcol \hypertarget{P44}{P44} & Sharif, M. A.; Raghavan, V. V.; & 2014 & A Clustering Based Scalable Hybrid Approach for Web Page 
	& IEEE & 2014 IEEE International Conference on Big Data \\ [1ex]
	
	P45 & Xu, S.; Watada, J.; & 2014 & A Method for Hybrid Personalized Recommender based on Clustering of Fuzzy User Profiles & IEEE & IEEE International Conference on Fuzzy Systems (FUZZ-IEEE) July 6-11, 2014, Beijing, China \\ [1ex]
	
	\rowcol \hypertarget{P46}{P46} & Lee, K.; Lee, K.; & 2014 & Using Dynamically Promoted Experts for Music Recommendation & IEEE & IEEE Transactions on 
	Multimedia, VOL. 16, NO. 5, August 2014 \\ [1ex]
	
	\hypertarget{P47}{P47} & Chughtai, M. W.; Selamat, A.; Ghani, I.; Jung, J. J.; & 2014 & E-Learning Recommender Systems Based on Goal-Based 
	Hybrid Filtering & IEEE & International Journal of Distributed Sensor Networks Volume 2014pages \\ [1ex]
	
	\rowcol P48 & Li, Y.; Lu, L.; Xufeng, L. & 2005 & A hybrid collaborative filtering method for multiple-interests and 
	multiple-content recommendation in E-Commerce & Science Direct & Expert Systems with Applications 28 (2005) 67–77 \\ [1ex]
	
	\hypertarget{P49}{P49} & Kunaver, M.; Pozrl, T.; Pogacnik, M.; Tasic, J.; & 2007 & Optimisation of combined collaborative recommender 
	systems & Science Direct & International Journal of Electronics and Communications (AEU), 2007, 433-443 \\ [1ex]
	
	\rowcol P50 & Albadvi, A.; Shahbazi, M.; & 2009 & A hybrid recommendation technique based on product category attributes & Scopus & Expert Systems with Applications 36 (2009) 11480–11488 \\ [1ex]
	
	\hypertarget{P51}{P51} & Capos, L. M.; Fernandez-Luna, J. M.; Huete, J. F.; Rueda-Morales, M. A.; & 2010 & Combining content-based and 
	collaborative recommendations: A hybrid approach based on Bayesian networks & Science Direct & International Journal of Approximate 
	Reasoning 51 (2010) 785–799 \\ [1ex]
	
	\rowcol \hypertarget{P52}{P52} & Barragans-Martínez, A. B.; Costa-Montenegro, E.; Burguillo, J. C.; Rey-Lopez, M.; Mikic-Fonte, F. A.; Peleteiro, A.; 
	& 2010 & A hybrid content-based and item-based collaborative filtering approach to recommend TV programs enhanced with 
	singular value decomposition & Science Direct & International Journal of Information Sciences 180 (2010) 4290–4311 \\ [1ex]
	
	\hypertarget{P53}{P53} & Wen, H.; Fang, L.; Guan, L.; & 2012 & A hybrid approach for personalized recommendation of news on the Web
	& Science Direct & International Journal of Expert Systems with Applications 39 (2012) 5806–5814 \\ [1ex]
	
	\rowcol P54 & Porcel, C.; Tejeda-Lorente, A.; Martinez, M. A.; Herrera-Viedma, E.; & 2012 & A hybrid recommender system for 
	the selective dissemination of research resources in a Technology Transfer Office & Science Direct & International Journal of Information 
	Sciences 184 (2012) 1–19 \\ [1ex]
	
	\hypertarget{P55}{P55} & Noguera, J. M.; Barranco, M. J.; Segura, R. J.; Martinez, L.; & 2012 & A mobile 3D-GIS hybrid recommender system for tourism & Science Direct & International Journal of Information Sciences 215 (2012) 37–52 \\ [1ex]
	
	\rowcol P56 & Salehi, M.; Pourzaferani, M.; Razavi, S. A.; & 2013 & Hybrid attribute-based recommender system for learning 
	material using genetic algorithm and a multidimensional information model & Science Direct & Egyptian Informatics Journal (2013) 
	14, 67–78 \\ [1ex]
	
	P57 & Zang, Z.; Lin, H.; Liu, K.; Wu, D.; Zhang, G.; Lu, J.; & 2013 & A hybrid fuzzy-based personalized recommender 
	system for telecom products/services & Science Direct & International Journal of Information Sciences 235 (2013) 117–129 \\ [1ex]
	
	\rowcol \hypertarget{P58}{P58} & Kardan, A. A.; Ebrahimi, M.; & 2013 & A novel approach to hybrid recommendation systems based on association rules 
	mining for content recommendation in asynchronous discussion groups & Science Direct & International Journal of Information Sciences 219 
	(2013) 93–110 \\ [1ex]
	
	\hypertarget{P59}{P59} & Lucas, J. P.; Luz, N.; Moreno, M. N.; Anacleto, R.; Figueiredo, A. A.; Martins, C.; & 2013 & A hybrid recommendation 
	approach for a tourism system & Science Direct & International Journal of Expert Systems with Applications 40 (2013) 3532–3550 \\ [1ex]
	
	\rowcol \hypertarget{P60}{P60} & Son, L. H.; & 2014 & HU-FCF: A hybrid user-based fuzzy collaborative filtering method in Recommender Systems & Science Direct & International Journal of Expert Systems with Applications 41 (2014) 6861–6870 \\ [1ex]
	
	\hypertarget{P61}{P61} & Son, L. H.; & 2014 & HU-FCF++: A novel hybrid method for the new user cold-start problem in recommender systems & Scopus & Engineering Applications of Artificial Intelligence 41(2015)207–222 \\ [1ex]
	
	\rowcol P62 & Lekakos, G.; Caravelas, P.; & 2006 & A hybrid approach for movie recommendation & Springer & Multimed Tools Appl (2008) 36:55–70 DOI 10.1007/s11042-006-0082-7, Springer \\ [1ex]
	
	\hypertarget{P63}{P63} & Lekakos, G.; Giaglis, G. M.; & 2007 & A hybrid approach for improving predictive accuracy of collaborative 
	filtering algorithms & Springer & User Model User-Adap Inter (2007) 17:5–40 DOI 10.1007/s11257-006-9019-0, Springer \\ [1ex]
	
	\rowcol P64 & Degemmis, M.; Lops, P.; Semeraro, G.; & 2007 & A content-collaborative recommender that exploits WordNet-based 
	user profiles for neighborhood formation & Springer & User Model User-Adap Inter (2007) 17:217–255, DOI 10.1007/s11257-006-9023-4, 
	Springer \\ [1ex]
	
	P65 & Cho, J.; Kang, E.; & 2010 & Personalized Curriculum Recommender System Based on Hybrid Filtering & Springer & ICWL 2010, 
	LNCS 6483, pp. 62–71, 2010, Springer \\ [1ex]
	
	\rowcol P66 & Aksel, F.; Biturk, A.; & 2010 & Enhancing Accuracy of Hybrid Recommender Systems through Adapting the Domain Trends & Scopus & Workshop on the Practical Use of Recommender Systems, Algorithms and Technologies held in conjunction with RecSys 2010. Sept. 30, 2010, Barcelona \\ [1ex]
	
	\hypertarget{P67}{P67} & Lampropoulos, A. S.; Lampropoulos, P. S.; Tsihrintzis, G. A.; & 2011 & A Cascade-Hybrid Music Recommender 
	System for mobile services based on musical genre classification and personality diagnosis & Springer & Multimed Tools Appl 
	(2012) 59:241–258 DOI 10.1007/s11042-011-0742-0, Springer \\ [1ex]
	
	\rowcol \hypertarget{P68}{68} & Chen, W.; Niu, Z.; Zhao, X.; Li, Y.; & 2012 & A hybrid recommendation algorithm adapted in e-learning 
	environments & Springer & World Wide Web (2014) 17:271–284 DOI 10.1007/s11280-012-0187-z \\ [1ex]
	
	P69 & Sanchez, F.; Barrileo, M.; Uribe, S.; Alvarez, F.; Tena, A.; Mendez, J. M.; & 2012 & Social and Content Hybrid 
	Image Recommender System for Mobile Social Networks & Springer & Mobile Netw Appl (2012) 17:782–795 DOI 10.1007/s11036-012-0399-6, 
	Springer \\ [1ex]
	
	\rowcol \hypertarget{P70}{P70} & Zheng, X.; Ding, W.; Xu, J.; Chen, D.; & 2013 & Personalized recommendation based on review topics & Scopus & SOCA (2014) 8:15–31 DOI 10.1007/s11761-013-0140-8 \\ [1ex]
	
	P71 & Cao, J.; Wu, Z.; Wang, Y.; Zhuang, Y.; & 2013 & Hybrid Collaborative Filtering algorithm for bidirectionalWeb service recommendation & Springer & Knowl Inf Syst (2013) 36:607–627 DOI 10.1007/s10115-012-0562-1 \\ [1ex]
	
	\rowcol P72 & Burke, R.; Vahedian, F.; Mobasher, B.; & 2014 & Hybrid Recommendation in Heterogeneous Networks 
	& Springer & UMAP 2014, LNCS 8538, pp. 49–60, 2014, Springer \\ [1ex]
	
	P73 & Nikulin, V.; & 2014 & Hybrid Recommender System for Prediction of the Yelp Users Preferences
	& Springer & ICDM 2014, LNAI 8557, pp. 85–99, 2014, Springer \\ [1ex]
	
	\rowcol P74 & Sarne, G. M. L.; & 2014 & A novel hybrid approach improving effectiveness of recommender systems
	& Springer & J Intell Inf Syst DOI 10.1007/s10844-014-0338-z \\ [1ex]
	
	\hypertarget{P75}{P75} & Zhao, X.; Niu, Z.; Chen, W.; Shi, C.; Niu, K.; Liu, D.; & 2014 & A hybrid approach of topic model and matrix 
	factorization based on two-step recommendation framework & Springer & J Intell Inf Syst DOI 10.1007/s10844-014-0334-3, 
	Springer \\ [1ex]
	
	\rowcol \hypertarget{P76}{P76} & Nilashi, M.; Ibrahim, O. B.; Ithnin, N.; Zakaria, R.; & 2014 & A multi-criteria recommendation system using 
	dimensionality reduction and Neuro-Fuzzy techniques & Springer & Soft Comput DOI 10.1007/s00500-014-1475-6, Springer-Verlag 
	Berlin Heidelberg 2014 \\ [1ex]
	\bottomline
	\label{table:FinalIncluded}
\end{longtable}
\endgroup
%
%
\begin{table}[h!]
	\caption{Questions for quality assessment} 
     \footnotesize
	\centering     
	\begin{tabular}
		{p{7.8cm} l		 r}   
		\toprule
		\textbf{Quality Question} & \textbf{Score} & \textbf{Weight}  \\ [0.5ex] 
		\midrule
		QQ1. Does it clearly describe the addressed
		problems ? & yes/partly/no (1/0.5/0) & 1 \\     [0.8ex]         
		QQ2. Does it review related work 
		on the problems? & yes/partly/no (1/0.5/0) & 0.5      \\	[0.8ex]
		QQ3. Does it recommend future
		research work? & yes/partly/no (1/0.5/0) & 0.5 \\	[0.8ex]
		QQ4. Does it describe each module 
		of the system? & yes/partly/no (1/0.5/0) & 1.5  \\	[0.8ex]
		QQ5. Does it empirically  
		evaluate the system? & yes/partly/no (1/0.5/0) & 1.5 \\	[0.8ex]
		QQ6. Does it present a clear formulation 
		of findings? & yes/partly/no (1/0.5/0) & 1  \\		[0.8ex]
		\bottomrule
	\end{tabular} 
\label{table:PaperQuality}
\end{table}
\begin{table}[h!] 
	\caption{Form for data extraction} 
	\small 
	\centering      
	\begin{tabular}
		{l l l}  
		\topline
		\headcol \textbf{Extracted Data} & \textbf{Explanation} & \textbf{RQ}  				\\ [0.5ex]  
		\midline
		ID & A unique identifier of the form Pxx we set to each paper & -					\\
		\rowcol Title & - & RQ1																		\\
		Authors & - & -																		\\
		\rowcol Publication year & - & RQ1															\\
		Conference year & - & -																\\
		\rowcol Volume & Volume of the journal & -													\\
		Location & Location of the conference & -													\\
		\rowcol  Source & Digital library from which was retrieved & -								\\
		Publisher & - & -																	\\
		\rowcol Examiner & Name of person who performed data extraction & -							\\
		Participants & Study participants like students, academics, etc. 					\\
		\rowcol Goals & Work objectives & -															\\
		Application domain & Domain in which the study is applied & RQ5								\\
		\rowcol Approach & Hybrid recommendation approach applied & RQ3							\\
		Contribution & Contribution of the research work & -								\\
		\rowcol Dataset & Public dataset used to train and evaluate the algorithm & RQ6			\\
		DM techniques & Data mining techniques used & RQ3									\\
		\rowcol Evaluation methodology & Methodology used to evaluate the RS & RQ6					\\
		Evaluated characteristic & RS characteristics evaluated & RQ6 								\\
		\rowcol Future work & Suggested future works & RQ7											\\
		Hybrid class & Class of hybrid RS & RQ4												\\
		\rowcol Research problem & - & RQ2															\\
		Score & Overall weighted quality score & - 											\\
		\rowcol Other Information & - & -															\\
		\bottomline 
	\end{tabular}  
\label{table:DataExtraction}
\end{table}

\end{appendices}

\printthesisindex 

\end{document}